\documentclass[10pt,journal,compsoc]{IEEEtran}

\ifCLASSOPTIONcompsoc
  \usepackage[nocompress]{cite}
\else
  \usepackage{cite}
\fi

\usepackage{url}
\usepackage{caption}
\usepackage{graphicx}
\usepackage{amsmath}
\usepackage{booktabs}
\usepackage{algorithm}
\usepackage{algorithmic}
\usepackage{amsfonts,amsmath,amssymb,amsthm}
\usepackage{color}
\usepackage{multirow}
\usepackage{enumitem}
\usepackage{nicefrac}
\usepackage{bbm}
\usepackage{bm}
\usepackage[normalem]{ulem}

\usepackage{caption}

\newcommand{\abs}[1]{\lvert#1\rvert}

\newcommand{\norm}[1]{\left\| #1 \right\|_F}
\newtheorem{lemma}{Lemma}
\newtheorem*{lemma*}{Lemma}
\newtheorem{thm}{Theorem}
\newtheorem*{thm*}{Theorem}
\newtheorem{prop}{Proposition}
\newtheorem*{prop*}{Proposition}

\newtheorem{remark}{Remark}
\newtheorem{assumption}{Assumption}

\newtheorem{definition}{Definition}[section]

\def\g{\gamma}

\def\l{\lambda}
\def\0{{\bf 0}}
\def\1{{\bf 1}}
\def\dX{{\tilde{X}}}

\def\t{{^\top}}

\def\T{{\mathcal T}}

\def\C{{\mathcal C}}
\def\M{{\mathcal M}}

\def\R{{\mathbb{R}}}
\def\S{{\mathbb S}}

\def\O{{\mathcal O}}

\def\tr{\text{\normalfont tr}}

\def\({\big(}
\def\){\big)}

\newcommand{\nnz}[1]{\text{\normalfont nnz}(  #1 ) }

\usepackage{subfigure}
\graphicspath{{figures/}}

\newcommand{\NM}[2]{\| #1 \|_{#2}}
\def\hl{\textbf}

\usepackage{array}
\newcolumntype{L}[1]{>{\raggedright\let\newline\\\arraybackslash\hspace{0pt}}m{#1}}
\newcolumntype{C}[1]{>{\centering\let\newline  \\\arraybackslash\hspace{0pt}}m{#1}}
\newcolumntype{R}[1]{>{\raggedleft\let\newline \\\arraybackslash\hspace{0pt}}m{#1}}
\begin{document}
%
\title{Efficient Low-Rank Semidefinite Programming with Robust Loss Functions}
%
%
%
%

\author{Quanming Yao,~\IEEEmembership{Member~IEEE},
        Hansi Yang,
        En-Liang Hu~\IEEEmembership{Member~IEEE}, \\
        and~James T. Kwok,~\IEEEmembership{Fellow IEEE}
\IEEEcompsocitemizethanks{\IEEEcompsocthanksitem 
	Q. Yao is with 4Paradigm Inc
	and Department of Electronic Engineering, Tsinghua University.
	E-mail: qyaoaa@connect.ust.hk
	
	\IEEEcompsocthanksitem
	H. Yang is with Department of Electronic Engineering, Tsinghua University.
	E-mail: yhs17@mails.tsinghua.edu.cn
	
	\IEEEcompsocthanksitem
	E.-L. Hu is with Department of Mathematics, Yunnan Normal University.
	Email: ynel.hu@gmail.com
	
	\IEEEcompsocthanksitem J.T. Kwok is with Department of Computer Science and Engineering,
	Hong Kong University of Science and Technology. 
	Email: jamesk@cse.ust.hk
	
	\IEEEcompsocthanksitem 
	The work is performed when H. Yang was an intern in 4Paradigm;
	and correspondence is to Q. Yao.
}
\thanks{}}

%
%

\markboth{}{}
%



\IEEEtitleabstractindextext{%
\begin{abstract}
In real-world applications, it is 
important for
machine learning algorithms to be robust against data outliers or corruptions.
In this paper, we focus on 
improving the robustness of a large class of learning algorithms that are formulated as low-rank semi-definite
programming (SDP) problems.
Traditional formulations use
the square loss,
which is notorious for being sensitive to outliers.
We propose to replace this with 
 more robust noise models, including the
 $\ell_1$-loss 
and other nonconvex losses.
However, the resultant optimization problem becomes difficult as the objective is
no
longer convex or smooth. 
To alleviate this problem, we design an efficient algorithm based on majorization-minimization.
The crux is on constructing a good optimization surrogate, and we show that this surrogate can
be efficiently obtained by the alternating direction method of multipliers (ADMM).
By properly monitoring ADMM's convergence, 
the proposed algorithm is empirically efficient and also
theoretically guaranteed to converge to a critical point.
Extensive experiments are performed on four
machine learning applications using both synthetic and real-world data sets.
Results show that the proposed algorithm
is not only fast but also has better performance than the state-of-the-arts.
\end{abstract}

}

\maketitle

\IEEEdisplaynontitleabstractindextext

\IEEEpeerreviewmaketitle

\IEEEraisesectionheading{\section{Introduction}\label{sec:introduction}}

\IEEEPARstart{M}{any} machine learning problems involve the search for
matrix-valued solutions. The corresponding optimization problems are often 
formulated as linear
semidefinite programs
(SDP) 
\cite{helmberg1996interior,review:sdp96,boyd2004convex,lemon2016low}
of the form:
\begin{align}\label{eq:lsdp}
\min\nolimits_{Z \in \mathbb{S}_{+}} \tr(Z A)
\quad
\text{s.t.}
\;
\tr(Z Q_{\tau}) = t_{\tau},
\;
\forall \tau = 1, \dots, m,
\end{align}
in which the objective is linear and the target matrix $Z\in \R^{n\times n}$ is positive semi-definite (PSD).
Here, $\S_+$ is the cone of PSD matrices,
$Q_{\tau}$ and $A$ are some matrices and
$t_{\tau}$ is a scalar.
Prominent examples include
matrix completion~\cite{soren12,jaggi,laurent2014positive,bishop2014deterministic,bhargava2017active},
ensemble learning~\cite{singh2010ensemble},
clustering~\cite{Kulis-2007,pirinen2019exact},
sparse PCA~\cite{d:spca07,yuan2013truncated,zou2018selective}, 
maximum variance unfolding (MVU)~\cite{weinberger04,song08},
and non-parametric kernel learning (NPKL)~\cite{zhenguo08,zhuang11}.
The instantiations of $Q_\tau, t_\tau$ and $A$
are application-dependent (with details in Section~\ref{sec:app}).
For example,
in matrix completion,
$Q_{\tau}$ is constructed from the positions of observed entries,
$t_{\tau}$ is the corresponding observed value, 
and $A$ is the identity matrix.
In NPKL,
$Q_{\tau}$
is constructed from sample indices in the must-link/cannot-link constraint,
$t_{\tau}$ indicates whether it is a must-link or cannot-link,
and $A$ is a Laplacian matrix for the data manifold.

A standard SDP solver is the interior-point method (IPM)
\cite{helmberg1996interior}.  
In each 
iteration, a sub-problem 
based on the Lagrangian and log-det barrier function
has to be solved
numerically, and
each such sub-problem iteration takes $\O(n^3)$ time.
To reduce the computational cost,
Yang et al.
\cite{toh2015sdpnal+}
introduced 
an efficient  Newton-based algorithm 
SDPNAL+
to solve the augmented Lagrangian of (\ref{eq:lsdp}).
While $\O(n^3)$ time is still required
in each iteration of the sub-problem,
the total number of iterations can be much smaller,
and enables SDPNAL+ to be faster than IPM.

An alternative approach is 
to completely avoid enforcing the variate to be PSD. 
Instead, 
$Z$ 
is replaced
by a low-rank decomposition $XX\t$, where $X \in \R^{n \times r}$ and $\text{rank}(Z)\le r$ \cite{burer2003,burer2005local,Kulis-2007,bach:sdp,lemon2016low}.  Many learning problems, such as matrix completion, MVU and NPKL, also prefer a low-rank $Z$.
Problem~(\ref{eq:lsdp}) is then transformed to the 
nonlinear optimization problem:
\begin{align}\label{eq:nonlinear}
\!\!\!\min\nolimits_X \tr(XX\t A)
\; \text{s.t.}
\;
\tr(XX\t Q_{\tau}) = t_{\tau},
\;
\forall \tau = 1, \dots, m.
\end{align}
It can be shown theoretically that the factorized problem is equivalent to the original problem when the rank of the solution is deficient \cite{bach:sdp,burer2005local,srinadh15,zheng2015convergent,boumal2016non,boumal2020deterministic}.
Burer and Monteiro \cite{burer2003} introduced
SDPLR, which 
uses the augmented Lagrangian method together
with limited memory BFGS to solve 
(\ref{eq:nonlinear}).
Due to the loss of convexity, more inner and outer loop iterations may be needed.  

To avoid handling the $m$ difficult constraints
($\tr(XX\t Q_{\tau})$ $=$ $t_{\tau}$)
in (\ref{eq:nonlinear}),
a common trick is to 
allow them to be
slightly violated~\cite{nocedal2006numerical}
This
leads to
the optimization problem:
\begin{align}
\label{eq:new}
\min_{X}
\sum\nolimits_{\tau=1}^m \frac{1}{2}( \tr(X\t Q_{\tau}X)-t_{\tau} )^2
+ 
\frac{\gamma}{2} \tr (X\t AX),
\end{align}
where the first term measures constraint violations,
and $\gamma$ is a hyper-parameter controlling the corresponding penalty.
To prevent over-fitting, we further add the regularizer $\NM{X}{F}^2$ to 
(\ref{eq:new}), leading to:
\begin{align}\label{eq:sdpfac}
\!\!\!\!\!
\min_{X} \!
\sum\nolimits_{\tau=1}^m \!
\frac{1}{2}( \tr(X^{\top} \! Q_{\tau} X) \! - \! t_{\tau} )^2 
\! + \! 
\frac{\gamma}{2} \tr (X^{\top} \!\! A X)
\! + \! \frac{\lambda}{2} \NM{X}{F}^2,\!\!
\end{align} 
where $\lambda > 0$ is a tradeoff parameter.
This regularizer has also been popularly
used in matrix factorization  applications
\cite{mnih2008probabilistic,zheng2012practical,hastie2015matrix,lin2017robust}.
One then only needs to solve the smooth unconstrained optimization problem 
(\ref{eq:sdpfac})
w.r.t. $X$.
Gradient descent 
\cite{srinadh15,zheng2015convergent,bhojanapalli2016dropping} 
has been developed
as the state-of-the-art solver for this type of problems.
It has convergence guarantees
with linear/sub-linear convergence rates for certain low-rank formulations~\cite{srinadh15,pumir2018smoothed}.

Recall that the square loss is used
in \eqref{eq:new} and 
(\ref{eq:sdpfac})
to measure constraint violations.
It is well-known that the square loss is sensitive to outliers~\cite{huber1992robust,de2003framework,candes2011robust,zheng2012practical,lin2017robust}.
This can be problematic as,
for example, 
in MVU,
the samples can be corrupted 
\cite{dekel2010learning};
in kernel learning,
the similarity constraints
may come from spammers
\cite{raykar2010learning};
in matrix completion,
there can be attacks in the observed entries~\cite{burke2015robust,yao2018scalable}.
These corruptions and noise
can significantly deteriorate 
the performance \cite{raykar2010learning,lin2017robust}.
To make the models more 
robust, a common approach is to 
replace the square loss by more robust noise models.
These include the $\ell_1$-loss~\cite{huber1992robust,lu2013online}
and, more recently, concave losses
such as the minimax concave penalty (MCP)~\cite{zhang2010nearly}
and log-sum penalty (LSP) \cite{candes2008enhancing}.
These concave loss functions are similar in shape to Tukey’s biweight function in robust statistics~\cite{huber1992robust}, which flattens more for larger values.
Recently, they have also been successfully used in
matrix factorization
\cite{de2003framework,eriksson2010efficient,candes2011robust,zheng2012practical,jiang2015robust,lin2017robust,yao2017efficient,yao2018scalable}.
However, so far
they have not been used in SDPs.

Motivated by the needs for both optimization efficiency and robustness in
learning the matrix variate, 
we propose 
in this paper
the use of robust loss functions
with the matrix factorization in \eqref{eq:sdpfac}.
However,
the resulting optimization problem
is neither convex (due to factorization) nor smooth (due to the robust loss).
Hence,
none of the above-mentioned solvers 
can be used.
To handle this difficult problem, we propose 
a new optimization algorithm based on Majorization-Minimization (MM) \cite{lange2000optimization,hunter2004tutorial}.
The crux of MM is on constructing a good surrogate that is easier to optimize. We
show that this surrogate can be efficiently optimized by  the
alternating direction method of multipliers (ADMM) \cite{boyd2011distributed,he20121}.
While MM 
only guarantees convergence to limit points,
we show that the proposed algorithm ensures convergence to a critical point even
when the ADMM is only solved inexactly.
Efficiency and robustness
of the proposed algorithm 
are demonstrated 
on five machine learning applications, namely, PSD matrix completion,
nonparametric kernel learning, maximum variance unfolding, sparse PCA, and 
symmetric non-negative matrix factorization.
Results
show that it
is not only faster, but also has better performance over the state-of-the-arts.

A preliminary version of this paper has been published in the IJCAI-2019 conference \cite{hu2019robust}.
Compared with the conference version,
the major changes here are:
\begin{itemize}[leftmargin=*]
\item 
In 
\cite{hu2019robust},
we assumed that the 
optimization of the
convex surrogate is solved
exactly.
Here, 
we allow the subproblem to be solved only inexactly,
making the whole algorithm more efficient in practice (Section~\ref{sec:admm}).
Besides,
we show that
when the inexactness is properly controlled,
convergence to critical points is still theoretically guaranteed
(Section~\ref{sec:conv}).

\item To further promote robustness,
we consider using a nonconvex loss 
(Section~\ref{sec:nloss})
to replace the $\ell_1$-loss
used in the conference version.
We show that the proposed algorithm
can still be applied with some modifications,
and convergence is also guaranteed.

\item Two more applications
namely, PSD matrix completion 
(Section~\ref{sec:matcomp})
and
symmetric nonnegative matrix factorization 
(Section~\ref{sec:snmf}),
are presented.

\item Extensive experiments with more applications,
baseline algorithms,
convergence studies, and ablation study
are performed in 
Section~\ref{sec:exp}.
\end{itemize}

\noindent
\textbf{Notations.}
We use uppercase letters for matrices, and lowercase letters for scalars.
The transpose of a vector or matrix is denoted by the superscript $(\cdot)^{\top}$. 
The identity matrix is denoted $I$.
For a matrix $A=[a_{ij}]$,
$\norm{A}= (\sum\nolimits_{ij}a_{ij}^2)^{1/2}$ is its Frobenius norm;
$\NM{A}{*} = \sum\nolimits_i \sigma_i(A)$
is its nuclear norm,
where $\sigma_i(A)$ is the $i$th singular value of $A$;
and $\tr(A)$ is its trace (when the matrix is square). 
A matrix is positive semi-definite (PSD) if its eigenvalues are
non-negative.
Besides,
$\odot$ denotes the element-wise product between two matrices: $[ A \odot B ]_{ij}
= A_{ij} B_{ij}$; and
$\abs{\mathcal{S}}$ is  the size of a set $\mathcal{S}$.

\section{Related Works}
\label{sec:relw}

\subsection{Majorization-Minimization (MM)}
\label{ssec:mm}

Majorization-minimization (MM) \cite{lange2000optimization,hunter2004tutorial} is a general technique to make difficult optimization problems easier.
Consider a
function $f(X)$, 
which is hard to optimize. Let the iterate at
the $k$th MM iteration be $X_k$. The next iterate is generated as
\begin{align}
X_{k+1} = X_k + \arg\min\nolimits_{\dX} 
h_k( \dX; X_k ),
\label{eq:mmnext}
\end{align}
where $h_k$ is a
surrogate that is being optimized instead of $f$. A good
surrogate should have the following properties \cite{lange2000optimization}: 
\begin{itemize}[leftmargin = 25px]
	\item[(P1).] $f(\dX + X_k) \le h_k(\dX; X_k)$ for any $\tilde{X}$; 
	\item[(P2).]  $0\in\arg\min_{\tilde{X}} (h_k(\dX; X_k) - f(\dX + X_k))$ and $f(X_k) = h_k(\bm{0}; X_k)$; and 
	\item[(P3).] $h_k$ is convex on $\tilde{X}$. 
\end{itemize}
Condition (P3) allows the minimization of $h_k$ in (\ref{eq:surr}) to be easily solved.
Moreover, 
from (P1) and (P2),
we have
\begin{equation} \label{eq:surr}
f(X_{k + 1})
\le \min\nolimits_{\tilde{X}} h(\tilde{X}; X_k)
\le h( \bm{0}; X_k) = f(X_k).
\end{equation} 
Thus, 
the objectives obtained
in successive iterations are non-increasing. However,
MM does not
guarantee convergence of the sequence $\{X_k\}$ \cite{hunter2004tutorial,lange2000optimization,mairal2013optimization,lin2017robust}.

\subsection{Alternating Direction Method of Multipliers (ADMM)}
\label{sec:rel:admm}

Recently, the alternating direction method of multipliers (ADMM)
\cite{boyd2011distributed,he20121} has been popularly used in machine learning and
data mining.
Consider optimization problems of the form
\begin{equation} \label{eq:admm}
\min\nolimits_{X,Y} \; \phi(X)+\psi(Y) \; : \; AX+By= c,  
\end{equation} 
where $\phi, \psi$ are convex functions, and $A,B$ (resp. $c$) are
constant matrices (resp. vector). ADMM considers the augmented Lagrangian
$
\mathcal{L}(X,Y,\nu) 
= \phi(X)+\psi(Y)+\nu^{\top} (AX+By-c)+\frac{\rho}{2} \NM{AX+By-c}{2}^2,
$
where $\nu$ is the dual variable, and $\rho>0$ is a penalty parameter.
At the $t$th iteration, the values of $X$ and $Y$ (denoted $X_t$ and $Y_t$)
are updated by minimizing $\mathcal{L}(X,Y,\nu_{t})$ w.r.t.  $X$ and $Y$ 
in an alternating manner:
\begin{align}
X_{t + 1} = & \arg \min\nolimits_{X} \mathcal{L}(X, Y_{t},\nu_{t}),
\label{eq:admm:sub1}
\\
Y_{t + 1} = & \arg \min\nolimits_{Y} \mathcal{L}(X_{t + 1},Y,\nu_{t}).
\label{eq:admm:sub2}
\end{align}
Then, $\nu$ is updated as $\nu_{t + 1} = \nu_{t} + \rho(AX_{t + 1} + B Y_{t + 1} - c)$.

\subsection{Robust Matrix Factorization (RMF)}
\label{sec:rmf}

In matrix factorization (MF), the data matrix $O \in \R^{m \times n}$ is approximated by
$U V^{\top}$,
where $U \in \R^{m \times r}$, $V \in \R^{n \times r}$ and $r \ll \min( m, n )$ is
the rank.
In general, some entries of $O$ may be missing
(as in applications such as structure from motion \cite{basri2007photometric} and
recommender systems \cite{mnih2008probabilistic}).
The MF problem is thus formulated as:
\begin{equation} \label{eq:mf}
\min_{U,V}
\frac{1}{2} \NM{\Omega  \odot  ( O  -  U V^{\top} ) }{F}^2
+ \frac{\lambda}{2}( \NM{U}{F}^2 + \NM{V}{F}^2 ),
\end{equation} 
where 
$\Omega \in \{ 0, 1 \}^{m \times n}$ 
contain indices to the observed entries in $O$ (with $\Omega_{ij} = 1$ if $O_{ij}$ is observed, and 0 otherwise),
and 
$\lambda \ge 0$ is a regularization parameter.
The square loss 
in \eqref{eq:mf} 
is sensitive to outliers.  
In \cite{de2003framework},
it is replaced by the $\ell_1$-loss,
leading to robust matrix factorization (RMF):
\begin{equation}
\min_{U,V} 
\NM{\Omega \odot ( O - U V^{\top} ) }{1}
+ \frac{\lambda}{2} ( \NM{U}{F}^2 + \NM{V}{F}^2 ).
\label{eq:rmf}
\end{equation}
In recent years, many RMF solvers have been developed, e.g., \cite{eriksson2010efficient,kim2015efficient,lin2017robust,zheng2012practical}.
However, as the objective in \eqref{eq:rmf} is neither convex nor smooth,
these solvers lack scalability,
robustness
and/or convergence guarantees.
Recently,
the RMF-MM algorithm \cite{lin2017robust}
solves \eqref{eq:rmf} using MM.  Let the $k$th iterate be $(U_k, V_k)$.  RMF-MM  tries to find increments $(\tilde{U}, \tilde{V})$ that should be
added to
$(U_k, V_k)$
in order to  obtain
the target $(U, V)$, i.e.,
$U = U_k + \tilde{U}$
and 
$V = V_k + \tilde{V}$.
Substituting into \eqref{eq:rmf},
the objective 
can be rewritten as
\begin{align*}
F^k(  \tilde{U}, \tilde{V} ) 
\equiv & \NM{\Omega \odot ( O \! - \! ( U_k + \tilde{U} ) ( V_k + \tilde{V} )^{\top} )}{1}
\\
& + \frac{\lambda}{2} \NM{U_k + \tilde{U}}{F}^2
+ \frac{\lambda}{2} \NM{V_k + \tilde{V}}{F}^2.
\end{align*}
The following Proposition
constructs a surrogate $H^k$ 
satisfying properties (P1)-(P3) 
in Section~\ref{ssec:mm}
for being a good MM surrogate.
Unlike $F^k$, 
$H^k$ is jointly convex in $(\tilde{U}, \tilde{V})$.

\begin{prop}[\cite{lin2017robust}]
\label{pr:rmf:surr}
Let $\nnz{ \Omega_{(i,:)} }$ (resp. $\nnz{\Omega_{(:,j)}}$)
be the number of nonzero elements in the $i$th row (resp. $j$th column) of $\Omega$, 
$\Lambda^r
= \text{Diag}
(\sqrt{\smash[b]{ \nnz{ \Omega_{(1,:)} } }}, \dots, \sqrt{\smash[b]{ \nnz{\Omega_{(m,:)} }} })$,
and $\Lambda^c  = \text{Diag}(  \sqrt{\smash[b]{  \nnz{ \Omega_{(:,1)}} } }, \dots, \sqrt{\smash[b]{
		\nnz{ \Omega_{(:, n)} }}})$.
Then,
$F^k ( \tilde{U}, \tilde{V} ) \le H^k( \tilde{U}, \tilde{V} )$,
where
\begin{align}
\!\!\!\!
H^k(\tilde{U}, \tilde{V}) 
& \! \equiv \!
\NM{\Omega \! \odot \! ( O \! - \! U_k V_k^{\top} \! - \! \tilde{U} V_k^{\top} \! - \! U_k \tilde{V}^{\top} )}{1} 
\! + \! \frac{1}{2} \NM{\Lambda^r \tilde{U}}{F}^2
\notag
\\
+ & \frac{\lambda}{2} \NM{U_k + \tilde{U}}{F}^2 
+ \frac{\lambda}{2} \NM{V_k + \tilde{V}}{F}^2 + \frac{1}{2} \NM{\Lambda^c \tilde{V}}{F}^2.
\label{eq:surr2}
\end{align}
Equality holds iff $( \tilde{U}, \tilde{V} ) = (0, 0)$.
\end{prop}

Because of the coupling of 
$\tilde{U}, V_k$ 
(resp.  $U_k, \tilde{V}$)
in $\tilde{U} V_k^{\top}$ (resp. $U_k \tilde{V}^{\top}$) in  
\eqref{eq:surr2},
$H^k$
is still difficult to optimize.
Thus, 
RMF-MM uses ADMM to optimize \eqref{eq:surr2}.
RMF-MM is guaranteed to generate critical points of \eqref{eq:rmf}.

\section{SDP Learning with $\ell_1$-Loss}

Here, 
we replace the square loss in \eqref{eq:sdpfac} by the more
robust $\ell_1$-loss. 
This leads to the following robust version of (\ref{eq:sdpfac}):
\begin{align}
\min_{X}
R(X) 
\equiv 
\sum\nolimits_{\tau=1}^m 
& \abs{\tr(X\t Q_{\tau}X) - t_{\tau}}
\notag
\\
& + \frac{\gamma}{2} \tr( X^{\top} A X )
+ \frac{\lambda}{2}\|X\|_F^2.
\label{eq:rsdp}
\end{align}
With 
the 
$\ell_1$-loss,
the objective in
(\ref{eq:rsdp})
is neither convex nor smooth.
Hence, existing algorithms for solving (\ref{eq:sdpfac})
(such as L-BFGS \cite{burer2003},
gradient descent \cite{srinadh15,zheng2015convergent},
and coordinate descent \cite{enliang11})
can no longer be used.

\subsection{Optimization Algorithm}

Recall from Section~\ref{ssec:mm} that MM is a general technique to make difficult
optimization problems easier to optimize.
Recently, MM has also been used successfully in the RMF solvers of
RMF-MM \cite{lin2017robust} and RMFNL \cite{yao2018scalable}.
In this Section,
we design an efficient solver for \eqref{eq:rsdp} based on MM.
While 
RMF-MM and RMFNL 
construct the surrogate
by first factorizing
the target matrix $Z$ 
as $X Y^{\top}$ and then bounding
$X$ and $Y$ 
separately,
our construction of the  
surrogate for (\ref{eq:rsdp}) 
is significantly different.

\subsubsection{Constructing a Convex Surrogate}
\label{sec:cvxsurr}

Let 
$X_k$
be 
the iterate at the $k$th MM iteration.
Recall 
from \eqref{eq:mmnext}
that the next iterate is constructed as
$X_k+
\tilde{X}$  for some
$\tilde{X} \in 
\R^{n \times r}$.
The following Lemma
bounds $R$
for any $\tilde{X}$,
where
$R$ is the objective defined in \eqref{eq:rsdp}.
\begin{lemma} \label{pr:upper1}
Let $C = A+\frac{\l}{\g}I$.
For any $\tilde{X} \in \R^{n \times r}$, 
\begin{align*}
& 
R(
X_k
+
\dX
) 
\le \sum\nolimits_{\tau=1}^m
\abs{\tr(2\dX\t Q_{\tau}X_k + X_k^{\top} Q_{\tau}X_k) - t_{\tau}} 
\\
& \! + \!
\sum\nolimits_{\tau=1}^m \abs{\tr(\dX\t Q_\tau \dX)} 
\! + \! \frac{\gamma}{2} \tr(\dX^\t C\dX + (X_k + 2\dX)^{\top} C X_k).
\end{align*} 
\end{lemma}
As $| \tr(\tilde{X}^{\top} Q_\tau\dX) |$ is convex only when $Q_\tau \in \S_+$ \cite{boyd2004convex},
the upper bound above is not convex in general.
The following provides a looser bound on 
$| \tr(\tilde{X}^{\top} Q_\tau\dX) |$ 
which is convex w.r.t. $\dX$. 
We 
first
introduce some notations.
Given a symmetric square matrix $M$,
let 
its eigenvalues 
be 
$\g_i$'s 
and 
the corresponding eigenvectors
be $v_i$'s.
Let $M_+ = \sum\nolimits_i\max(\g_i,0)v_iv_i\t$ be the matrix constructed by using
only the positive eigenvalues, and
similarly $M_- = \sum\nolimits_i\min(\g_i,0)v_iv_i\t$
is constructed from only the negative eigenvalues.
Obviously, $M=M_+ + M_-$.

\begin{lemma}\label{lemma:sur_abstr}
$| \tr(\dX\t Q_\tau\dX) | \le \tr( \dX\t \bar{Q}_\tau\dX )$, where
$\bar{Q}_\tau = \frac{1}{2} (Q_\tau+Q_\tau^{\top})_+ - \frac{1}{2}(Q_\tau+Q_\tau^{\top})_{-}$ is PSD.
\end{lemma}

Combining Lemmas~\ref{pr:upper1} and \ref{lemma:sur_abstr},
a surrogate 
$H_k$
is constructed as follows.
\begin{prop} \label{pr:cvxsurr}
$R(\dX + X_k)\le H_k( \dX; X_k )$,
where
\begin{eqnarray}
H_k(\dX; X_k) 
\!\!\!\! & \equiv & \!\!\!\! \tr( \dX\t ( B \dX + \g C X_k ) )
\notag
\\
& \!\!\! + & \!\!\!\!\!\!
2\sum\nolimits_{\tau=1}^m \abs{\tr( \dX \t Q_{\tau} X_k ) + (b_k)_\tau} + c_k,
\label{eq:surrh}
\end{eqnarray}
with $B = \sum\nolimits_{\tau=1}^m \bar{Q}_\tau + \frac{1}{2}(\l I +\g A_+)$, 
$C=A +\frac{\l}{\g}I$,
$(b_k)_\tau = \frac{1}{2}(\tr( X_k \t Q_{\tau} X_k ) - t_{\tau})$, 
$c_k=\frac{\gamma}{2}\tr(X_k\t (A+\frac{\l}{\g}I)X_k)$.
Equality holds iff $\tilde{X} = \mathbf{0}$.
\end{prop}
It is easy to see  that
$H_k(\dX;X)$ is convex w.r.t. $\dX$
and $R(X_k) = H_k( \bm{0}; X_k)$.
Besides,
from Proposition~\ref{pr:cvxsurr},
we also have 
$R(\dX + X_k) \le H_k(\dX; X_k)$ for any $\tilde{X}$,
and $\mathbf{0} = \arg\min_{\tilde{X}} (H_k(\dX; X_k) - R(\dX + X_k))$. 
Thus,
$H_k$
satisfies
the three desired properties for a MM surrogate in Section~\ref{ssec:mm}.

\begin{table*}[ht]
\centering
\caption{Comparison of the proposed SDP-RL algorithm with existing algorithms on
matrix completion problems (details are in Section~\ref{sec:matcomp}).
The last row shows SDP-RL on sparse data,
where ``nnz'' is the number of nonzero elements.
For algorithms with subproblems, $T$ is the number of iterations to solve the subproblem. }
\vspace{-8px}
\setlength\tabcolsep{10pt}
\begin{tabular}{cc | c  c | c c}
	\hline
	                        &                                      &         \multicolumn{2}{c|}{model}          &                   \multicolumn{2}{c}{complexity}                    \\
	                        &                                      & factorized               & loss             & space                         & 
									time (per-iteration)                                \\ \hline
	            \multicolumn{2}{c|}{FW~\cite{soren12}}             & $\times$                 & square loss      & $\O(n^2)$                     & $\O(n^2)$                           \\ \hline
	       \multicolumn{2}{c|}{L-BFGS~\cite{nocedal2006numerical}}         & $\surd$                  & square loss      & $\O(nr)$                      & $\O(nr^2)$                          \\ \hline
	        \multicolumn{2}{c|}{nmAPG~\cite{nconv_apg15}}          & $\surd$                  & square loss      & $\O(nr)$                      & $\O(nr^2)$                          \\ \hline
	\multicolumn{2}{c|}{ADMM($\ell_1$)~\cite{boyd2011distributed}} & $\times$                 & $\ell_1$ loss    & $\O(n^2)$                     & $\O(n^2rT)$                         \\ \hline
	         \multicolumn{2}{c|}{SDPLR~\cite{burer2003}}           & $\times$                 & $\ell_1$ loss    & $\O(nr)$                      & $\O(n^2rT)$                         \\ \hline
	      \multicolumn{2}{c|}{SDPNAL+~\cite{toh2015sdpnal+}}       & $\times$                 & $\ell_1$ loss    & $\O(n^2)$                     & $\O(n^3T)$                          \\ \hline\hline
	\multirow{2}{*}{SDP-RL} &              dense data              & \multirow{2}{*}{$\surd$} & $\ell_1$ loss or & $\O(n^2)$                    & $\O(n^2 r T)$                       \\ \cline{2-2}\cline{5-6}
	                        &             sparse data              &                          & nonconvex loss   & $\O(\text{nnz}(\Omega) + nr)$ & $\O((\text{nnz}(\Omega)r + nr^2)T)$ \\ \hline
\end{tabular}
\label{tab:compare}
\end{table*}

\subsubsection{Solving the Surrogate Inexactly by ADMM}
\label{sec:admm}

From \eqref{eq:mmnext},
$X_k$ 
is updated 
at the $k$th MM iteration
as
$X_{k + 1} = X_k + \dX^*$,
where
\begin{equation} 
\label{eq:subprob}
\dX^* = \arg\min\nolimits_{\dX} H_k(\dX; X_k).
\end{equation} 
First, \eqref{eq:subprob}  can be easily rewritten as
\begin{eqnarray}
\!\!\!\!\! \!\!\!\!\! & \min\nolimits_{\dX, \{ e_{\tau}\}} & \!\!\!\!\! \tr(\dX\t (B\dX+\g CX_{k})) 
+ 2\sum\nolimits_{\tau=1}^m \abs{e_\tau}
+ c_k
\label{eq:surradmm}
\\
\!\!\!\!\! \!\!\!\!\! & \text{s.t.} & \!\!\!\!\! 
e_\tau =\tr( \dX \t Q_{\tau} X_{k} ) + (b_k)_\tau,
\;
\tau =1, \dots, m.\nonumber
\end{eqnarray}
As in Section~\ref{sec:rel:admm},
let $\tilde{\nu}_{\tau}$ be
the 
dual variable 
associated 
with the $\tau$th constraint
in \eqref{eq:surradmm}.
The dual of 
(\ref{eq:surradmm})
is given by the following Proposition.

\begin{prop}\label{pr:dual}
The dual problem of 
(\ref{eq:surradmm})
is 
\begin{equation} \label{eq:dual}
\max\nolimits_{ \{ \tilde{\nu}_{\tau} \} \in \mathcal{C} } \mathcal{D}_k(\{
\tilde{\nu}_{\tau} \}),
\end{equation} 
where
$\mathcal{D}_k (\{ \tilde{\nu}_{\tau} \})$
$=$ 
$c_k \!+ \!$ 
$\frac{\gamma}{2}$ 
$\sum\nolimits_{\tau=1}^m$ 
$\tilde{\nu}_\tau ( \tr ( (CX_k)^{\top} B^{-1}$ 
$Q_\tau X_k )$ 
$\!-\!$ 
$\frac{2}{\gamma} (b_k)_\tau )$
$\!-\!$ 
$\frac{1}{4}$ 
$\sum\nolimits_{\tau_1 = 1}^m$ 
$\sum\nolimits_{\tau_2 = 1}^m$ 
$\tilde{\nu}_{\tau_1}\!$  
$\tilde{\nu}_{\tau_2}\!$ 
$( \tr ( (  Q_{\tau_1} X_k)^{\top} \!\! B^{-1}$  
$(Q_{\tau_2}  X_k) )$ 
$\!-\!$ 
$\frac{\gamma^2}{4}$ 
$\tr ( (CX_k)^{\top}$
$\!\!B^{-1}$
$(CX_k) )$
and
$\mathcal{C} = \cup_{\tau = 1}^m \{ \tilde{\nu}_{\tau}  \,|\, |\tilde{\nu}_{\tau}| \le 2 \}$.
\end{prop}

By using the Slater condition~\cite{boyd2004convex},
the
following Lemma
shows that
strong duality 
for \eqref{eq:surradmm}
and (\ref{eq:dual})
holds.
\begin{lemma} \label{lem:dual}
Strong duality 
for \eqref{eq:surradmm}
and (\ref{eq:dual})
holds.
\end{lemma}

In the following, we again use ADMM to solve
\eqref{eq:surradmm}.
At the $k$th ADMM iteration,
it can be easily shown that 
the 
updates in \eqref{eq:admm:sub1} and \eqref{eq:admm:sub2}
have the following
closed-form solutions:
\begin{align}
\dX_{t + 1} & =  \dX_{t} - \tilde{B}_k^{-1} (2B \dX_t +  \tilde{C}_k X_{k} ),
\label{eq:admmx} 
\\
(e_{\tau})_{t + 1} & =  \max(0, \tilde{e}^-_{\tau}) + \min(0, \tilde{e}^+_{\tau}),
\label{eq:admme}
\end{align}
where
$\tilde{B}_k$ 
$=$ 
$2B + \rho\sum\nolimits_{\tau=1}^m  Q_\tau X_k X_k^{\top} Q_\tau^{\top}$,
$\tilde{C}_k$ 
$=$ 
$\gamma C$ 
$+$ 
$\sum\nolimits_{\tau=1}^m$ 
$( \rho (\tr ( \dX_t \t Q_{\tau} X_{k} ) - e_{\tau} + (b_k)_\tau ) {-} \tilde{\nu}_{\tau} ) Q_{\tau}$,
and
$\tilde{e}^{\pm}_{\tau}$
$=$ 
$\tr ( \dX_{t + 1}^{\t} Q_{\tau} X_{k} ) + (b_k)_\tau + \frac{\tilde{\nu}_{\tau} \pm 2}{2\rho}$.
Each of the ADMM dual variables $\{ \tilde{\nu}_{\tau} \}_{\tau = 1, \dots, m}$
is 
then
updated as 
\begin{align}
\!\!\!
(\tilde{\nu}_{\tau})_{t + 1} 
& 
\! = \! 
(\tilde{\nu}_{\tau})_{t}
\! + \! \rho
\big(
(e_\tau)_{t + 1} 
\! - \! \tr( \dX_{t + 1} \t Q_{\tau} X_{k} ) 
\! + \! (b_k)_\tau
\big).
\label{eq:admmnu}
\end{align}
Because of 
strong duality (Lemma~\ref{lem:dual}),
the duality gap 
is zero
at optimality.
Recall that 
\eqref{eq:subprob}  and
\eqref{eq:surradmm} have the same objective value,
one can thus use the duality gap
\begin{equation} \label{eq:gap}
\delta_k(\dX_t, \{ (\tilde{\nu}_{\tau})_t \}) 
= H_k(\dX_t; X_k) - \mathcal{D}_k(\{ (\tilde{\nu}_{\tau})_t \}),
\end{equation}
at the $t$th ADMM iteration
to monitor convergence. 
In other words, the ADMM iterations can be stopped and an approximate solution to 
\eqref{eq:subprob} is found
when
$\delta_k(\dX_t, \{ (\tilde{\nu}_{\tau})_t \}) $
is smaller than a pre-defined threshold $\epsilon_k$.
The whole procedure for approximately solving subproblem \eqref{eq:subprob} 
is shown in Algorithm~\ref{alg:admm}.

\begin{algorithm}[ht]
	\caption{Solving subproblem \eqref{eq:subprob} by ADMM.}
	\label{alg:admm}
	\begin{algorithmic}[1]
		\REQUIRE pre-defined tolerance $\epsilon_k$;
		\STATE {\bfseries Initialization:} $t = 1,\tilde{X}_{1} = 0$;
		\WHILE{$\delta_k(\dX_t, \{ (\tilde{\nu}_{\tau})_t \}) \ge \epsilon_k$}
		\STATE obtain $\dX_{t}$ from \eqref{eq:admmx};
		\FOR{$\tau = 1,\dots, m$}
		\STATE obtain $(e_{\tau})_{t + 1}$ from \eqref{eq:admme};
		\STATE update $(\tilde{\nu}_{\tau})_{t + 1}$ from \eqref{eq:admmnu};
		\ENDFOR
		\STATE compute duality gap $\delta_k(\dX_t, \{ (\tilde{\nu}_{\tau})_t \})$;
		\STATE update $t = t + 1$;
		\ENDWHILE	
		\RETURN $\dX_{t}$. 	
	\end{algorithmic}	    
\end{algorithm}

\subsubsection{Complete Algorithm}
\label{sec:comp:alg}

The whole procedure for solving \eqref{eq:rsdp}, which will be called
SDP-RL (SDP with Robust Loss),
is shown in Algorithm \ref{alg:rsdpmm}.
Note that 
SDPLR~\cite{burer2003} and
SDPNAL+~\cite{toh2015sdpnal+} can also solve
optimization problems of the form:
\begin{align}
\!\!\!\!\!
\min\nolimits_{Z\in \S_+} 
\!\tr (Z A) 
\;\;
\text{s.t.}
\; 
| \tr(ZQ_{\tau}) 
\! - \!
t_{\tau} | 
\! \le \! 
\Delta, \; \tau =  1, \dots, m,
\label{eq:rsdpeq}
\end{align}
which is equivalent to the following optimization problem
\begin{align*}
\min\nolimits_{Z\in \S_+} \tr (Z A) 
+
\lambda \sum\nolimits_{\tau = 1}^m | \tr(ZQ_{\tau}) - t_{\tau} |,
\end{align*} 
with $\ell_1$-loss, 
when the regularization parameter $\lambda$ is a properly set.
Table~\ref{tab:compare} compares the proposed
SDP-RL 
with other algorithms using the $\ell_1$ and square losses on matrix completion problems (Section~\ref{sec:matcomp}).
As can be seen,
SDP-RL is both robust and fast.

\begin{algorithm}[ht]
\caption{SDP-RL: SDP with robust loss.}
\label{alg:rsdpmm}
\begin{algorithmic}[1]
	\STATE {\bfseries Initialization:}  $X_1 = 0$.
	\FOR {$k=1,\ldots, K$}	    
	\STATE obtain $\dX_t$ from Algorithm~\ref{alg:admm} with tolerance $\epsilon_k$;
	\STATE update $X_{k+1} = \dX_t + X_k$;		
	\ENDFOR	
	\RETURN $X_{K+1}$. 	
\end{algorithmic}	    
\end{algorithm}


\subsection{Convergence Analysis}
\label{sec:conv}

We make the following Assumption on the objective in \eqref{eq:rsdp}.

\begin{assumption}\label{ass:obj}
$\!\!\!\!\! \underset{\|X\|_F \rightarrow \infty}{\lim} \!\!\! R(X) = \infty$
and $\underset{X}{\inf} \; R(X) > - \infty$.
\end{assumption}

Related algorithms such as
RMF-MM \cite{lin2017robust} and RMFNL \cite{yao2018scalable} 
solve the sub-problem exactly
when their ADMM iterations 
terminate.
Here, we relax this condition and allow 
solving the sub-problem inexactly.
Hence, the proofs in \cite{lin2017robust,yao2018scalable} cannot be directly applied.

We assume
the following condition 
on the sequence of thresholds
$\{\epsilon_k\}$ in Algorithm~\ref{alg:rsdpmm}.

\begin{assumption}\label{ass:err}
$\epsilon_k \ge 0$ for $k = 1, \dots, \infty$ and
$\sum\nolimits_{k = 1}^{\infty} \epsilon_k$ is a finite positive constant.
\end{assumption}

\begin{remark} \label{rmk:eps}
A popular choice satisfying 
Assumption~\ref{ass:err} is 
$\epsilon_k = c_0/k^{b_0}$,
where $c_0 > 0$ and $b_0 > 1$ are constants \cite{apg_inexact,yao2016efficient}.
\end{remark}

Usually, MM only guarantees
that the objective value 
is non-increasing
\cite{lange2000optimization,hunter2004tutorial,lin2017robust}. In contrast,
the following Theorem
shows that 
the sequence of iterates obtained is bounded, and  
its limit points are also critical points.

\begin{thm} \label{thm:conv1}
With Assumptions~\ref{ass:obj} and \ref{ass:err}, 
for the sequence $\{ X_{k} \}$ generated from Algorithm~\ref{alg:rsdpmm},
we have
(i) $\{ X_k \}$ is bounded; and 
(ii) any limit point of $\{ X_k \}$ is a critical point of $R$.
\end{thm}


\section{SDP Learning using Nonconvex Loss}
\label{sec:nloss}

\begin{figure}[ht]
	\centering
	\includegraphics[width=0.60\columnwidth]{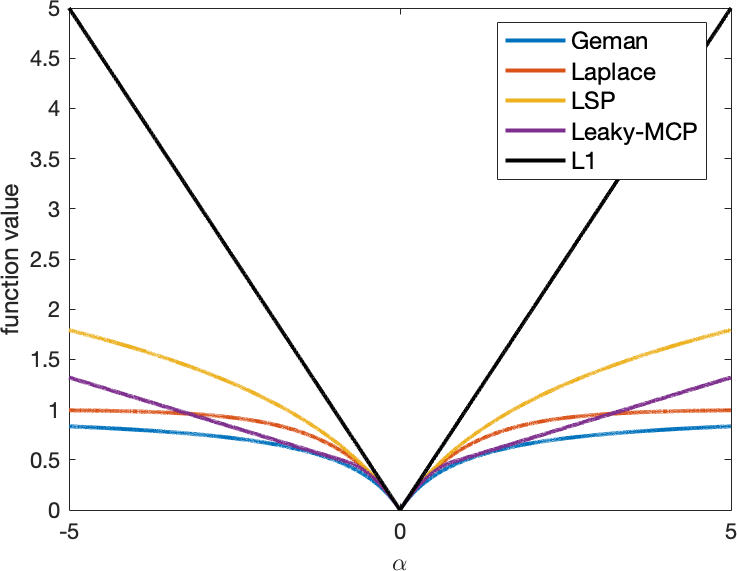}
	\vspace{-5px}
	\caption{More robust loss function $\phi$ in Table~\ref{tab:nloss} and the $\ell_1$ loss.}
	\label{fig:noncvx}
\end{figure}

The $\ell_1$-loss 
always linearly penalizes the difference between the prediction and noisy
observation.
In very noisy circumstances, 
a loss function $\phi$  flatter than the $\ell_1$ loss 
can be more robust~\cite{zhang2010analysis,gong2013general,yao2018scalable}.
Some common examples include the
	Geman penalty \cite{geman1995nonlinear},
	Laplace penalty \cite{trzasko2009highly},
	log-sum penalty (LSP) \cite{candes2008enhancing}, and 
	leaky-minimax concave penalty (MCP) \cite{zhang2010nearly}
(Figure~\ref{fig:noncvx}). 
They have been used in applications such as 
robust matrix factorization for affine rigid structure-from-motion~\cite{yao2018scalable},
where outliers arise from feature mismatch; and
sparse coding to learn more discriminative dictionaries~\cite{lu2013online,jiang2015robust},
in which large deviations come from damaged, deteriorating, or missing parts of an image.

In this section, we make the following Assumption on $\phi$.
\begin{assumption} \label{ass:loss}
$\phi(\alpha)$ is a concave and increasing function on $\alpha \ge 0$ with $\phi(0) = 0$.
\end{assumption}

\begin{table}[ht]
\caption{Example nonconvex $\phi$ functions.}
\centering
\vspace{-8px}
\label{tab:nloss}
\setlength\tabcolsep{3pt}
\begin{tabular}{C{98px} | c}
	\hline
	& $\phi(\alpha)$  \\ \hline
	Geman penalty \cite{geman1995nonlinear} & $\frac{\abs{\alpha}}{\theta+\abs{\alpha}}$ \\ \hline
	Laplace penalty \cite{trzasko2009highly} & $1-\exp(-\frac{\abs{\alpha}}{\theta})$ \\ \hline
	log-sum penalty (LSP) \cite{candes2008enhancing} & $\log(1+\abs{x})$ \\ \hline
	leaky-minimax concave penalty (MCP) \cite{zhang2010nearly} & $
	\begin{aligned}
	\begin{cases}
		-\frac{1}{2}\alpha^2 + \theta \abs{\alpha}, & 0 \le \alpha \le \theta-\eta \\
	\eta \abs{\alpha} + \frac{1}{2}(\theta-\eta)^2, &  \alpha > \theta - \eta
	\end{cases}
	\end{aligned}$          \\ \hline
\end{tabular}
\end{table}

Table~\ref{tab:nloss} shows
the corresponding $\phi$ functions for some popular nonconvex penalties.
With a nonconvex loss $\phi$,
problem~(\ref{eq:rsdp}) becomes
\begin{align}
\label{eq:rsdpnc}
\min_{X}
\dot{R}(X) 
\equiv 
& \sum\nolimits_{\tau=1}^m \phi\big( \abs{\tr(X\t Q_{\tau}X) - t_{\tau}} \big)
\\
& + \frac{\gamma}{2} \tr( X^{\top}\! A X )
+ \frac{\lambda}{2}\NM{X}{F}^2.
\notag
\end{align}
Because of the nonconvexity of $\phi$,
optimization 
of \eqref{eq:rsdpnc}
is even more difficult.
Again, we will alleviate this problem with the use of MM.

\subsection{Convex Surrogate and Its Optimization}

For any $\tilde{X} \in \R^{n \times r}$, 
the following Lemma
bounds 
$\dot{R}(X_k+ \tilde{X})$,
where
$\dot{R}$
is the objective 
in \eqref{eq:rsdpnc}.

\begin{lemma} \label{pr:upper2}
For any $\tilde{X} \in \R^{n \times r}$, 
\begin{align}
\dot{R}(X_k+ \tilde{X}) 
\le
&
\frac{\gamma}{2} \tr( (2 X_k+\tilde{X})^{\top} C \tilde{X} ) + \dot{c}_k 
\\ 
&
+ 
\sum\nolimits_{\tau=1}^{m} (q_k)_{\tau}
|  \tr( \tilde{X}^{\top} Q_{\tau} \tilde{X} ) | 
\notag
\\
&
+   
\sum\nolimits_{\tau=1}^{m} 
(q_k)_{\tau}
|  \tr( (2 \tilde{X} \!+ \!X_k)^{\top} \!Q_{\tau}\! X_k ) \! - \! t_{\tau} |,
\notag
\end{align} 
where 
$(q_k)_{\tau} = \phi' \(\vert \tr( X_k^{\top} Q_{\tau} X_k ) - t_{\tau} \vert\) $,
and
$\dot{c}_k = \sum\nolimits_{\tau=1}^{m} (\phi (\vert \tr( X_k^{\top} Q_{\tau} X_k ) - t_{\tau} \vert)- (q_k)_{\tau} \tr( X_k^{\top} Q_{\tau} X_k ) - t_{\tau} \vert )+\frac{\gamma}{2} \tr( X_k^{\top} C X_k )$. 

\end{lemma}
Combining with Lemma \ref{lemma:sur_abstr}, 
we construct a new surrogate 
as follows:

\begin{prop} 
\label{pr:cvxsurr2}
$\dot{R}(\dX + X_k)\le \dot{H}_k( \dX; X_k )$, where
\begin{align*}
\dot{H}_k( \dX; X_k ) & 
=  \tr( \tilde{X}^{\top} (B \tilde{X} + \gamma C X_k)) 
\\
& + 2\sum\nolimits_{\tau=1}^{m}\vert \tr( \tilde{X}^{\top} \dot{Q}_{\tau} X_k ) + (\dot{b}_{k})_{\tau} \vert + \dot{c}_k,
\end{align*} 
$\dot{Q}_{\tau}= (q_k)_{\tau} Q_{\tau}$, 
$B$ and $C$ as defined in Proposition~\ref{pr:cvxsurr},
and
$(\dot{b}_{k})_{\tau} \! = \! \frac{1}{2} (q_k)_{\tau} \( \tr(  X_k^{\top} Q_{\tau} X_k ) \! - \! t_\tau \)$.
Equality holds iff $\tilde{X} = \mathbf{0}$.
\end{prop}
Obviously,
$\dot{H}_k(\dX; X)$ is convex w.r.t. $\dX$.
Moreover,
it can be easily seen that the three desirable properties for MM surrogates (Section~\ref{ssec:mm}) are
also satisfied by {$\dot{H}_k(\dX; X)$}.
As in Section~\ref{sec:admm},
the 
optimization subproblem  for $\dot{H}_k$ can be 
reformulated  as:
\begin{align*}
\min\nolimits_{\tilde{X}} \;&\; \tr( \tilde{X}^{\top} (Q \tilde{X} + \gamma C X_k)) 
+ 2\sum\nolimits_{\tau=1}^{m} \vert e_{\tau} \vert+ \dot{c}_k,
\\
\text{s.t.}\;&\; e_{\tau} = \tr( \tilde{X}^{\top} \dot{Q}_{\tau} X_k ) + (\dot{b}_{k})_{\tau}
\quad
\tau =1, \dots, m.
\notag
\end{align*}
This is of the same form as \eqref{eq:surradmm}, and so can be 
solved analogously with ADMM.
Let
\begin{align*}
\dot{B}_k \! & =  \! 2Q + \rho\sum\nolimits_{\tau=1}^m  \dot{Q}_{\tau} X_k X_k^{\top}
\dot{Q}_{\tau}^{\top}, 
\\
\dot{C}_k \! & =  \! \gamma C 
\! + \! \sum\nolimits_{\tau=1}^m ( \rho (\tr ( \dX_t^{\top}
\dot{Q}_{\tau} X_{k} ) \! - \! e_{\tau} \! + \! (\dot{b}_k)_\tau ) {-} \nu_{\tau} )
\dot{Q}_{\tau}, 
\\
\dot{e}^{\pm}_{\tau}
\! & =  \! \tr ( \dX_{t + 1}^{\t} \dot{Q}_{\tau} X_{k} ) + (\dot{b}_k)_\tau +
\nicefrac{\nu_{\tau} \pm 2}{2\rho}, 
\\
\!\!\!\!
\dot{\mathcal{D}}_k (\{ \nu_{\tau} \})
\! & =  \! \frac{\gamma}{2} \sum\nolimits_{\tau=1}^m \nu_\tau ( \tr ( (C X_k)^{\top} Q^{-1} \dot{Q}_\tau X_k ) 
- \frac{2}{\gamma} (\dot{b}_k)_\tau )
\\
- & \frac{1}{4} \sum\nolimits_{\tau_1 = 1}^m \!
\sum\nolimits_{\tau_2 = 1}^m \! \nu_{\tau_1} \nu_{\tau_2} ( \tr ( ( \dot{Q}_{\tau_1}
X_k)^{\top} Q^{-1} (\dot{Q}_{\tau_2}  X_k) ) \\
- & \frac{\gamma^2}{4} \tr ( (CX_k)^{\top} Q^{-1} (CX_k) ) + \dot{c}_k.
\end{align*}
The resultant procedure, which consists of
Algorithms~\ref{alg:noncvx:2} and \ref{alg:rsdpmm:2}, are
slight modifications of 
Algorithms~\ref{alg:admm} and \ref{alg:rsdpmm}, respectively.
The 
main difference is on how
$\dot{Q}_\tau$ (resp. $Q_\tau$)
and
$(\dot{b}_k)_\tau$ (resp. $(b_k)_\tau$) are computed.
Thus, the complexity results in Table~\ref{tab:compare} still apply.

\begin{algorithm}[ht]
\caption{Variant of Algorithm 1 for nonconvex loss.}
\label{alg:noncvx:2}
\begin{algorithmic}[1]
	\REQUIRE pre-defined tolerance $\epsilon_k$.
	\STATE {\bfseries Initialization:} $t = 1,\tilde{X}_{1} = 0$;
	\WHILE{$\delta_k(\dX_t, \{ (\tilde{\nu}_{\tau})_t \}) \ge \epsilon_k$}
	\STATE update $\dX_{t + 1} = \dX_{t} - \dot{B}_k^{-1} (2Q \dX_t +  \dot{C}_k X_{k} )$;
	\FOR{$\tau = 1,\dots, m$}
	\STATE update $(\dot{e}_{\tau})_{t + 1} = \max(0, \dot{e}^-_{\tau}) + \min(0, \dot{e}^+_{\tau})$;
	\STATE $(\nu_{\tau})_{t + 1} \! = \! (\nu_{\tau})_{t} 
	\! + \! \frac{1}{\rho}
	( \dot{e}_\tau \! - \! \tr( X_{k}^{\t} \dot{Q}_{\tau} \dX_{t \! + \! 1} ) \! - \! (\dot{b}_k)_\tau )$;
	\ENDFOR
	\STATE compute 
	$\delta_k(\dX_t, \{ (\tilde{\nu}_{\tau})_t \})$,
	the upper-bound 
	on inexactness;
	\STATE $t = t + 1$;
	\ENDWHILE	
	\RETURN $\dX_{t}$. 	
\end{algorithmic}	    
\end{algorithm}

\begin{algorithm}[ht]
	\caption{SDP-RL for nonconvex loss.}
	\label{alg:rsdpmm:2}
	\begin{algorithmic}[1]
		\STATE {\bfseries Initialization:}  $X_1 = 0$.
		\FOR {$k=1,\ldots, K$}	    
		\STATE obtain $\dX_t$ via Algorithm~\ref{alg:noncvx:2} with tolerance $\epsilon_k$;
		\STATE update $X_{k+1} = \dX_t + X_k$;		
		\ENDFOR	
		\RETURN $X_{K+1}$. 	
	\end{algorithmic}	    
\end{algorithm}

\subsection{Convergence Analysis}
\label{sec:conv2}

In Section~\ref{sec:conv},
the convex $\ell_1$-loss is considered, and
the critical points can be characterized
by the subgradient
of $\ell_1$.
However, the
subgradient cannot be used on a nonconvex loss $\phi$.
The following first introduces generalizations of the subgradient
and critical point.

\begin{definition}[\cite{clarke1990optimization}] \label{def:subdifferential}
The {\em Frechet subdifferential} of $f$ at $x$ is
$\hat{\partial} f(x) = \{ u : \lim\nolimits_{y \neq x}\inf_{y \rightarrow x}
\frac{f(y) - f(x) - u^{\top} (y - x)}{\NM{y - x}{2}} \ge 0\}$.
The {\em limiting subdifferential} (or simply {\em subdifferential}) of $f$ at $x$
is $\partial f(x) = \{ u : \exists x_k \rightarrow x, f(x_k) \rightarrow f(x), 
u_k \in \hat{\partial} f(x_k) \!\rightarrow\! u, \text{ as } k \rightarrow \infty \}$.
\end{definition}

\begin{figure}[ht]
	\vspace{-10px}
	\centering
	\subfigure[subgradient.]
	{\includegraphics[width=0.48\columnwidth]{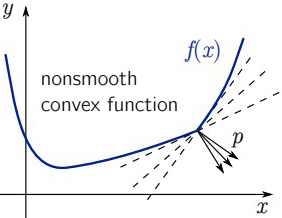}}
	\subfigure[subdifferential.]
	{\includegraphics[width=0.48\columnwidth]{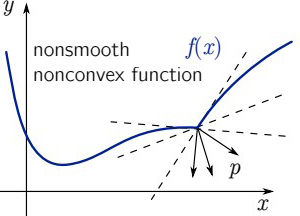}}
	\vspace{-8px}
	\caption{Plots of subgradient and  Frechet subdifferential,
		where $p$ denotes the normal direction.}
\label{fig:subdifferential}
\end{figure}

When $f$ is smooth,
$\partial f(x)$ reduces to the gradient.
When 
$f$ is nonsmooth but convex,
$\partial f(x)$ is the set of all subgradients of $f$ at $x$.
An example is shown in 
Figure~\ref{fig:subdifferential}.

\begin{definition}[\cite{clarke1990optimization}] \label{def:citical}
$x$ is a {\em critical point} of $f$ if $0 \in \partial f(x)$.
\end{definition}

We make the following Assumption
on $\dot{R}$,
which is analogous to Assumption~\ref{ass:obj} on $R$.

\begin{assumption}\label{ass:obj2}
$\underset{\|X\|_F \rightarrow \infty}{\lim} \!\!\!\! \dot{R}(X) = \infty$
and $\underset{X}{\inf} \; \dot{R}(X) > - \infty$.
\end{assumption}

Convergence to critical points is then ensured by the following Theorem.
Its proof can be easily adapted from that of Theorem~\ref{thm:conv1}.

\begin{thm} \label{thm:conv2}
With Assumptions~\ref{ass:err}, \ref{ass:loss} and \ref{ass:obj2}, 
for the sequence $\{ X_{k} \}$ generated from Algorithm~\ref{alg:rsdpmm:2},
we have
(i) $\{ X_k \}$ is bounded; 
(ii) any limit point of $\{ X_k \}$ is a critical point of $\dot{R}$.
\end{thm}

\section{Example Robust SDP Applications}
\label{sec:app}

In this section, we illustrate a number of applications that can be robustified
with the proposed formulations.
For simplicity, we focus on the $\ell_1$ loss.
These can be easily extended to the nonconvex loss
in Section~\ref{sec:nloss}.


\subsection{Positive Semi-Definite Matrix Completion}
\label{sec:matcomp}

The first example
is on
completing a partially observed PSD matrix \cite{laurent2014positive,bishop2014deterministic}.
This has 
applications  
in,
e.g.,
learning of the user-item matrix 
recommender systems~\cite{soren12} 
and 
multi-armed bandit problems~\cite{bhargava2017active}.
Let the data matrix be $O$,
and $\Omega \equiv \{ (i, j) \}$ be the set of indices 
for the 
observed
$O_{ij}$'s. 
PSD matrix completion can be formulated as finding a
$Z\in \S_+$ via
the following optimization
problem:
\begin{align}
\min\limits_{Z\in \S_+}
\sum\nolimits_{(i,j) \in \Omega} \frac{1}{2} (Z_{ij} - O_{ij})^2
+ \frac{\gamma}{2}\tr(Z),
\label{eq:recsys}
\end{align}
where the first term measures the loss,  
and the second term encourages the matrix $Z$ to be low-rank
(note that $\NM{Z}{*} = \tr(Z)$ for $Z\in \S_+$).
Let $Q^{(i,j)}$ be a matrix of zeros except that $Q^{(i,j)}_{ij} = 1$.
Problem
\eqref{eq:recsys} is then of the form in \eqref{eq:new}, 
with $Q_{\tau} = Q^{(i,j)}$,
$t_\tau= O_{ij}$, 
and $A = \mathbf{0}$.

The square loss 
in \eqref{eq:recsys}
may not be appropriate in some applications.
For example, 
the love-and-hate attack~\cite{burke2015robust}
in recommender systems
flips 
high ratings to low values, and vice versa \cite{yao2018scalable}.
The corrupted ratings then become large outliers, and
using the square loss 
can
lead to significant
performance degradation
\cite{lin2017robust,yao2018scalable}. 
To improve robustness, we replace the square loss by the $\ell_1$-loss,
leading to
\begin{align}
\min\limits_{Z\in \S_+}
\sum\nolimits_{(i,j) \in \Omega} |Z_{ij} - O_{ij}|
+ \frac{\gamma}{2}\tr(Z).
\label{eq:recsys2}
\end{align}
Let $Z=X X^{\top}$.
It is easy to see   that
$Z_{ij}=x_{i}^{\top} x_j$,
where $x_i^{\top}$ is the $i$th row of $X$,
and $\tr(Z) = \| X \|^2_F$.
Problem \eqref{eq:recsys2} can then be rewritten as:
\begin{align}
\min_{X}
\sum\nolimits_{(i,j) \in \Omega}
| \tr(X^{\top} Q^{(i,j)} X) - O_{ij} | 
+ \frac{\g}{2} \| X \|^2_F.
\label{eq:rbust:recsys}
\end{align}

\subsubsection{Utilizing Data Sparsity}
\label{ssec:spa}

Algorithms~\ref{alg:admm} and \ref{alg:rsdpmm}
can be directly used to solve \eqref{eq:rbust:recsys}. 
However,
each iteration of
Algorithm~\ref{alg:admm} 
has to
construct $Q_{\tau}$
(defined in Lemma~\ref{lemma:sur_abstr}) and invert $\tilde{B}_k$ (in \eqref{eq:admmx}).
A straightforward implementation leads to $O(n^2 r)$ time
and $O(n^2)$ space.
Recall that
the partially observed matrix $O$ is sparse.
In the following, we show how data sparsity can be used to speed up  optimization of
\eqref{eq:rbust:recsys},
as has been demonstrated in other matrix learning problems \cite{yao2018scalable,hastie2015matrix}.

\begin{prop}\label{pr:spa}
Let $\tilde{x}^{\top}_i$ (resp., $(x_k)^{\top}_i$) be the $i$th row of $\tilde{X}$ (resp., $X_k$).
The objective in \eqref{eq:surradmm} can be rewritten as
\begin{align*}
\min\nolimits_{\dX}  
\; & \; 
\frac{\g}{2}
\| \dX \|_F^2 
+  \frac{1}{2} \| \Lambda^r  \dX \|_F^2 
+  \frac{1}{2} \| \Lambda^c  \dX \|_F^2
\\ & 
+  \lambda \tr (\dX^{\top}  X_k)
+  2  \sum\nolimits_{\tau=1}^m \abs{e_\tau}
\\
\text{s.t.} 
\; & \;
e_{ij} = \tilde{x}_i^{\top} (x_k)_j + (b_k)_{ij},
\; \forall (i,j) \in \Omega,
\end{align*}
where $\Lambda^r$ and $\Lambda^c$ are as defined in Proposition~\ref{pr:rmf:surr},
and $(b_k)_{ij} = \frac{1}{2}((x_k)_i^{\top} (x_k)_j - O_{ij})$.
\end{prop}

Using Proposition~\ref{pr:spa},
the ADMM updates for $\tilde{X}$ and $e_{\tau}$ (in \eqref{eq:admmx}
and \eqref{eq:admme}, respectively)
become
\begin{align}
\dX_{t + 1} & = \dX_{t} - \hat{B}_k^{-1} ((\Lambda^r +\Lambda^c + \g I) \dX +  \hat{C}_k X_{k} ), 
\label{eq:addmm:mc}
\\
(e_{ij})_{t + 1} & = \max\left( 0, \hat{e}^-_{\tau} \right) + \min\left( 0, \hat{e}^+_{\tau} \right),
\notag
\end{align}
where
$\hat{B}_k  = \Lambda^r +\Lambda^c + \g I + {\Omega \odot (\rho X_k X_k^{\top})}$,
$\hat{C}_k  = \g I \! + \! \sum\nolimits_{(i,j) \in \Omega} ( \rho (\tilde{x}_i^{\top} (x_k)_j \! - \! e_{ij} \! + \! (b_k)_{ij} ) \! - \! \nu_{ij} ) Q^{(i,j)}$,
$\hat{e}^{\pm}_{ij} = (\tilde{x}_{t + 1})_i^{\top} (x_k)_j + (b_k)_{ij} + (\nu_{ij} \pm 2)/\rho$.
To update $\dX_t$ in
\eqref{eq:addmm:mc}
(and also $X_k$ in
Algorithm~\ref{alg:rsdpmm}), we only need to store sparse matrices ($\hat{B}_k$
and $\hat{C}_k$) and diagonal matrices ($\Lambda^r$ and $\Lambda^c$).
Moreover,
the second term on the R.H.S. of \eqref{eq:addmm:mc},
which involves inversion of $\hat{B}_k$,
can be computed in $O(n r^2 + \nnz{\Omega})$ time
using conjugate gradient descent
\cite{nocedal2006numerical}.
Finally,
each $(\nu_{\tau})_{t + 1}$ is updated as
$(\nu_{ij})_{t + 1} \! = \! (\nu_{ij})_{t} 
+ ( e_{ij} \! - \tilde{x}_i^{\top} (x_k)_j + (b_k)_{ij} )/\rho$
in $O(r)$ time.
Thus,
each ADMM iteration in Algorithm~\ref{alg:admm} only takes
$O(n r^2 +  \nnz{\Omega} r)$ time
and $O(n r + \nnz{\Omega})$ space,
which is much faster than the other SDP methods for the $\ell_1$ loss (see Table~\ref{tab:compare}).

\subsection{Robust NPKL}
\label{sec:rnpkl}

In nonparametric kernel learning (NPKL) \cite{Hoi07}, one tries to learn a kernel
matrix  from data.
In this section,
we adopt the formulation in \cite{zhenguo08,zhuang11}.
Let $\T=\M\cup\C$, where
$\M$ is the 
set containing sample pairs that should belong to the same class
({\em must-link} set),
and $\C$ is the 
set containing sample pairs that should not belong to the same class
({\em cannot-link} set).
This can be encoded by the matrix $O$, such that $O_{ij} = 1$ for a must-link $(i, j)$ pair,
and $O_{ij} = 0$ for a cannot-link pair.
The NPKL
problem is formulated as
the following SDP:
\begin{align}
\label{eq:NPKL}
\min\nolimits_{Z\in \S^+}
\sum\nolimits_{(i,j) \in \T} \( Z_{ij} - O_{ij} \)^2+\frac{\g}{2}\tr(ZL),
\end{align}
where 
$Z$ is the target kernel matrix
and $L$ is the 
Laplacian of the $k$-nearest neighbor graph of data.
The first term in \eqref{eq:NPKL} measures the difference between $Z_{ij}$  and 
$O_{ij}$, while
the second term 
encourages smoothness on the data manifold by
aligning $Z$  with $L$.

The must-links and cannot-links
are 
usually
provided by users or crowdsourcing. 
These can be noisy 
as there may be 
human errors and
spammers/attackers on crowdsourcing platforms
\cite{raykar2010learning}.
As in Section~\ref{sec:matcomp},
we let $Z=X X^{\top}$ and obtain the following robust NPKL formulation:
\begin{align*}
\min_{X} 
\!\!\!
\sum_{ (i,j)\in\T } 
\!\!\! 
| \tr(X\t  Q^{(i,j)}  X) - O_{ij} | 
+ \frac{\g}{2}\tr(X\t L X)
+ \frac{\lambda}{2}\NM{X}{F}^2,
\end{align*}
where $Q^{(i,j)}$ is the same as in Section~\ref{sec:matcomp}.
Obviously, this is again of the same form as \eqref{eq:surradmm}, 
with $Q_{\tau} = Q^{(i,j)}$,
$t_{\tau} = O_{ij}$ and $A = L$.
When $|\T|$ is small,
data sparsity can also be utilized as in Section~\ref{ssec:spa}.

\subsection{Robust CMVU}
\label{sec:rcmvu}

Maximum variance unfolding (MVU) \cite{weinberger04} 
is an effective dimensionality reduction method.
It produces a low-dimensional data representation by
simultaneously maximizing the variance of the embedding and preserving the local distances of the original data. 
Colored maximum variance unfolding (CMVU) is a ``colored'' variant of MVU
\cite{song08}, with class label information.
Let
\begin{align}
\bar{K} =  HTH,
\label{eq:lker}
\end{align}
where 
$T$ is a kernel matrix 
on labels,  and $H$ 
(with $H_{ij} = \mathbbm{1}(i=j)-{\frac{1}{n}}$) 
is a matrix 
that
centers the data and labels in the feature space.
CMVU is formulated as the following optimization problem:
\begin{align}
\label{eq:cmvu}
\min\limits_{Z\in \S^+} 
\sum\nolimits_{(i,j) \in \Omega}
\( Z_{ii} \! + \! Z_{jj} \! - \! 2 Z_{ij} \! - \! d_{ij} \)^2
\!\! - \frac{\g}{2}\tr(Z \bar{K}),
\end{align}
where $d_{ij}$ is
the squared Euclidean distance between the $i$th and $j$th samples in the original space, 
$\Omega$ is a set of neighbor pairs
whose distances are to be preserved in the embedding, 
and $\gamma$ controls the tradeoff between 
distance preservation (the first term)
and 
dependence maximization (second term).

Often, outliers and corrupted samples are introduced during data collection.
Again, by 
letting $Z=X X^{\top}$, we have the following robust CMVU formulation
which is of the same form as \eqref{eq:surradmm}:
\begin{align*}
\min_{X} \!\!
\sum\nolimits_{(i,j) \in \Omega} 
\abs{\tr(X^{\top} \! \hat{Q}^{(i,j)} X) 
\! - \! d_{ij}} 
\! - \! \frac{\g}{2}\tr(X^{\top} \! \bar{K} X)
\! + \! \frac{\lambda}{2}\NM{X}{F}^2,
\end{align*}
where $\hat{Q}^{(i,j)} = Q^{(i,i)} + Q^{(j,j)} - Q^{(i,j)} -
Q^{(i,j)}$ and
$Q^{(i,j)}$ is the same as that in Section~\ref{sec:matcomp}.
This again is the same form as 
\eqref{eq:rsdp}, 
with $Q_{\tau} = \hat{Q}^{(i,j)}$,
$t_{\tau} = d_{ij}$ and $A = -\bar{K}$.
When $|\Omega|$ is small,
data sparsity can also be utilized as in Section~\ref{ssec:spa}.

\subsection{Sparse PCA}
\label{sec:spac}

Sparse PCA 
\cite{d:spca07,zou2018selective}
is a popular method 
to extract 
sparse principal components from the data,
i.e., 
sparse vectors $x$ that maximizes $x\t \Sigma x$
for a given covariance matrix $\Sigma\in\R^{n\times n}$.
It can
be relaxed to the following SDP 
\cite{d:spca07}:
\begin{align} \label{eq:spca}
\min\nolimits_{Z\in \S_+}  
- \tr(Z\Sigma)
\quad\text{s.t.}\quad
\tr(Z)=1,\;
\NM{Z}{1} \le k,
\end{align}
where $k$ is a hyper-parameter controlling the sparsity.

As in previous sections, we let $Z=X X^{\top}$. 
Note that $\tr(Z) = \NM{X}{F}^2$ 
and 
$|Z_{ij}|
= 
|\tr(X^\t Q^{(i,j)} X)|$,
where $Q^{(i,j)}$ is in \eqref{eq:rbust:recsys}.
By moving the constraints 
in \eqref{eq:spca} 
to the objective,
it can be reformulated as
\begin{align}
\min_{X}
\sum\nolimits_{i,j} \abs{\tr(X^\t Q^{(i,j)} X)}
+ \frac{\lambda}{2}\NM{X}{F}^2
- \frac{\g}{2}\tr(X\t\Sigma X),
\label{eq:spca:trans}
\end{align}
which is the same as
\eqref{eq:rsdp}, 
with $Q_{\tau} \! = \! Q^{(i,j)}$,
$t_{\tau} \! = \! 0$ and $A \! = \! - \! \Sigma$.

\subsection{Symmetric NMF}
\label{sec:snmf}

Symmetric nonnegative matrix factorization (NMF)~\cite{he2011symmetric,shi2017inexact} 
aims to factorize a non-negative and symmetric matrix $O$ by solving
\begin{align}
\min\nolimits_{X} 
\frac{1}{2} \NM{O - X X^{\top}}{F}^2
\quad\text{s.t.}\quad
X_{ij} \ge 0.
\label{eq:snmf}
\end{align}  
SNMF is popular in clustering analysis~\cite{Kuang:2015aa} 
as it can effectively identify low-dimensional data representations.

Again, 
the square loss 
in \eqref{eq:snmf} may not be appropriate in some scenarios 
(for example, noisy observed data in clustering, which affect the observed $O_{ij}$'s), 
leading to degraded performance. 
Similar to Section~\ref{sec:matcomp}, 
we have the following robust SNMF formulation:
\begin{align}
\min_{X}
\sum\nolimits_{i,j}
| \tr(X^{\top} Q^{(i,j)} X) - O_{ij} | 
+ \frac{\lambda}{2} \| X \|^2_F,
\label{eq:snmf2}
\end{align}
which is of the form in \eqref{eq:rsdp} with 
$Q_{\tau} = Q^{(i,j)}$,
$t_{\tau} = O_{ij}$ and $A = \mathbf{0}$.

\begin{table*}[ht]
\centering
\caption{Testing RMSEs and CPU time (sec) on synthetic data with different 
matrix sizes
($m$).
The number in brackets is the percentage of observed elements.
`-' indicates the algorithm fails to converge in $10^4$ seconds.}
\vspace{-8px}
\begin{tabular}{c | c | c  c | c  c | c  c | c c}
	\hline
     \multirow{3}{*}{loss}        & \multirow{3}{*}{algorithm}  &    \multicolumn{2}{c|}{$m=500$ ($12.43\%$)}     &   \multicolumn{2}{c|}{$m=1000$ ($6.91\%$)}    &      \multicolumn{2}{c|}{$m=1500$ ($4.88\%$)}      &     \multicolumn{2}{c}{$m=2000$ ($3.80\%$)}      \\
                                  &                             & testing                  & CPU                  & testing                  & CPU                & testing                  & CPU                     & testing                  & CPU                   \\
                                  &                             & RMSE                     & time                 & RMSE                     & time               & RMSE                     & time                    & RMSE                     & time                  \\ \hline
	\multirow{3}{*}{$\!\!$square$\!\!$} & FW                          & 3.2$\pm$0.1              & \textbf{2$\pm$1}     & 3.8$\pm$0.1              & 5$\pm$1            & 4.2$\pm$0.1              & 8$\pm$1                 & 4.4$\pm$0.1              & \textbf{12$\pm$1}     \\
        \cline{2-10}           & nmAPG                       & 0.964$\pm$0.006          & \textbf{2$\pm$1}     & 0.785$\pm$0.004          & 5$\pm$1            & 0.637$\pm$0.008          & 6$\pm$1                 & 0.615$\pm$0.008          & 19$\pm$2              \\
                                  & L-BFGS                      & 0.964$\pm$0.006          & \textbf{2$\pm$1}     & 0.794$\pm$0.006          & \textbf{4$\pm$1}   & 0.638$\pm$0.006          & \textbf{6$\pm$1}        & 0.615$\pm$0.007          & 17$\pm$2              \\ \hline
                                  & ADMM($\ell_1$)              & 0.494$\pm$0.008          & $\!\!\!$54$\pm$9     & 0.394$\pm$0.008          & $\!\!\!$564$\pm$48 & 0.356$\pm$0.006          & $\!\!\!\!\!$1546$\pm$38 & 0.332$\pm$0.006          & $\!\!\!\!$2387$\pm$44 \\
                                  & SDPLR                       & 0.497$\pm$0.008          & $\!\!\!$5064$\pm$135 & 0.396$\pm$0.004          & 6784$\pm$246       & -                        & -                       & -                        & -                     \\
           $\ell_1$               & $\!\!$SDPNAL+$\!\!$         & 0.488$\pm$0.006          & $\!\!\!$397$\pm$45   & 0.388$\pm$0.006          & $\!$1562$\pm$189   & -                        & -                       & -                        & -                     \\
        \cline{2-10}           & SDP-RL-dense                & 0.246$\pm$0.004          & $\!\!\!$46$\pm$6     & 0.216$\pm$0.003          & 436$\pm$24         & 0.172$\pm$0.002          & $\!\!\!$1588$\pm$46     & 0.164$\pm$0.002          & $\!\!\!$2658$\pm$63   \\
                                  & SDP-RL($\ell_1$)            & 0.246$\pm$0.004          & 3$\pm$1              & 0.216$\pm$0.003          & 11$\pm$1           & 0.172$\pm$0.002          & 23$\pm$2                & 0.164$\pm$0.002          & 37$\pm$2              \\ \hline
     $\!\!$leaky-MCP$\!\!$        & $\!\!\!$SDP-RL(MCP)$\!\!\!$ & \textbf{0.126$\pm$0.002} & 6$\pm$1              & \textbf{0.121$\pm$0.002} & 16$\pm$2           & \textbf{0.117$\pm$0.002} & 27$\pm$3                & \textbf{0.113$\pm$0.001} & 46$\pm$2              \\ \hline
\end{tabular}
\label{tab:size}
\end{table*}

\begin{figure*}[ht]
	\centering
	
	\subfigure[$m = 500$.]{\includegraphics[width=0.243\textwidth]{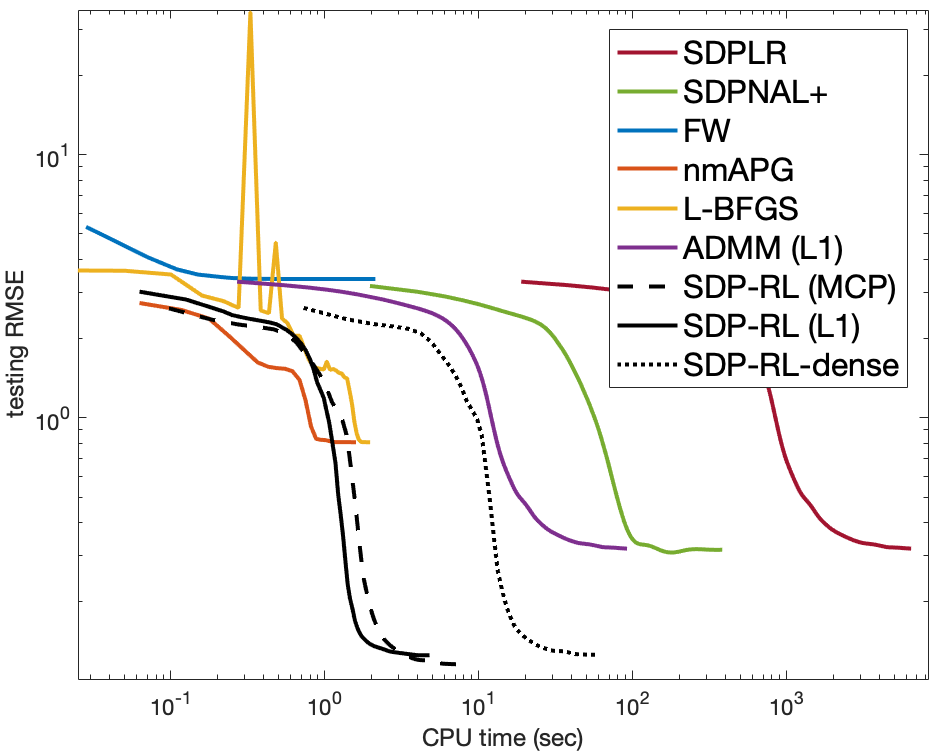}}
	\subfigure[$m = 1000$.]{\includegraphics[width=0.2425\textwidth]{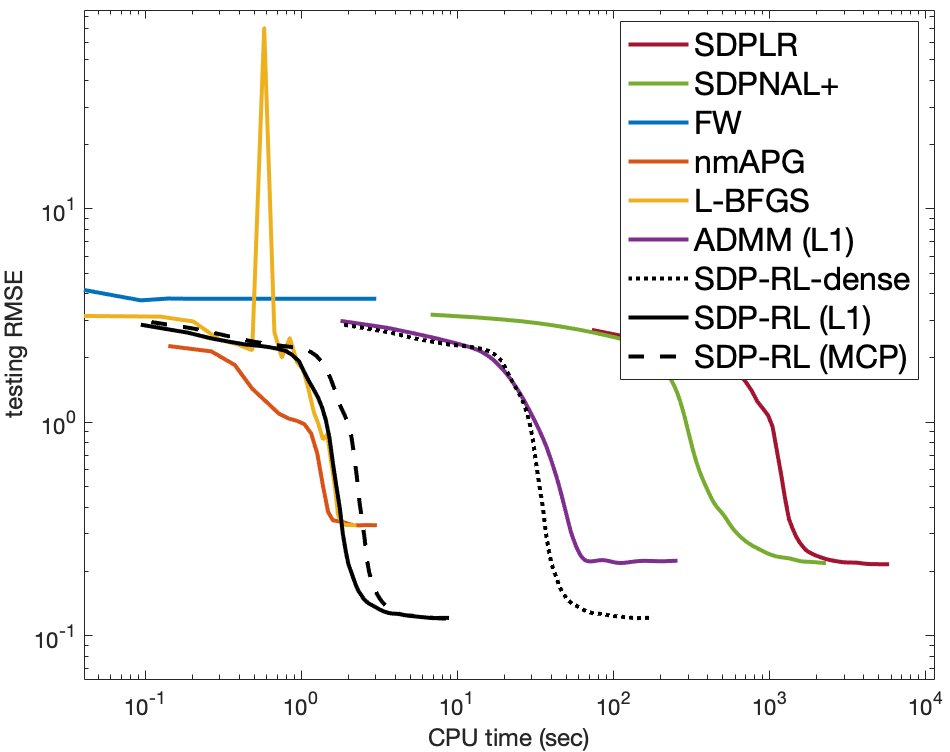}}
	\subfigure[$m = 1500$.]{\includegraphics[width=0.240\textwidth]{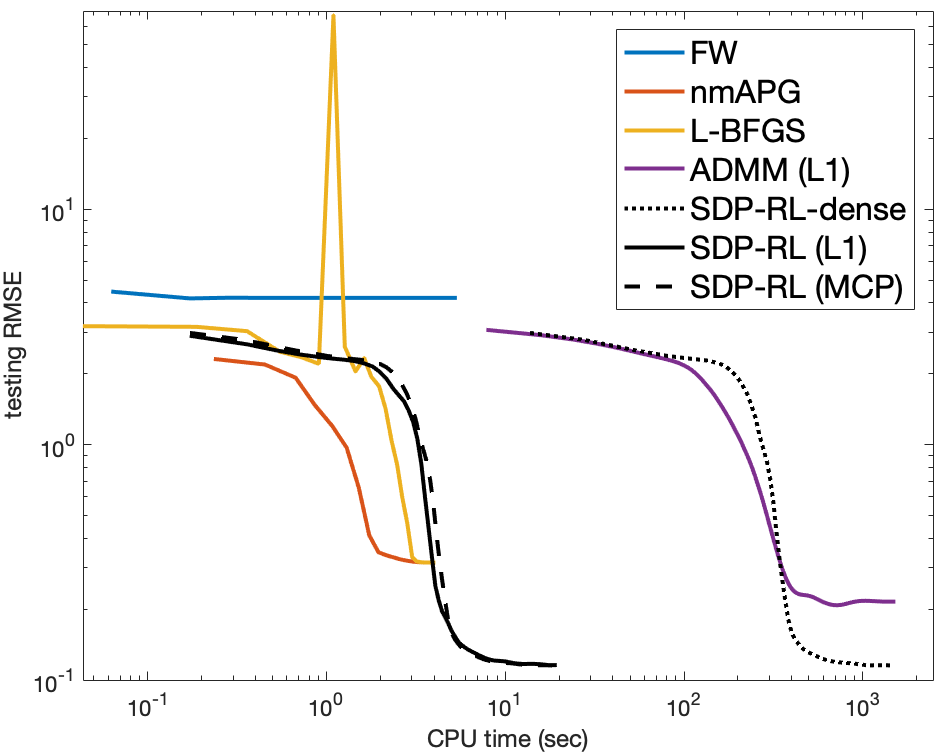}}
	\subfigure[$m = 2000$.]{\includegraphics[width=0.2425\textwidth]{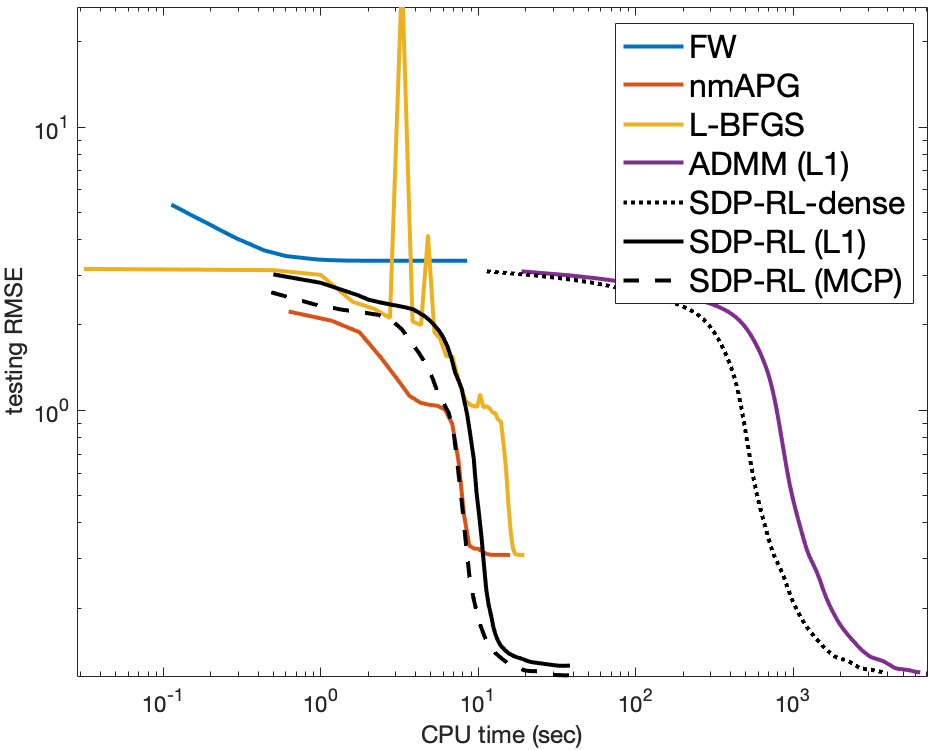}}

	\vspace{-8px}
	\caption{Convergence of the testing RMSE vs CPU time (sec) of various algorithms on synthetic data.
	SDPLR and SDPNAL+ are too slow on $m = 1500$ and $2000$, 
	thus are not shown.}
	\label{fig:syn:rmse}
\end{figure*}

\section{Experiments}
\label{sec:exp}

In this section, 
experiments 
are performed 
on five machine learning
applications, namely, PSD matrix completion
(Section~\ref{sec:exp:matcomp}),
non-parametric kernel learning 
(Section~\ref{sec:exp:rnpkl}),
maximum variance unfolding 
(Section~\ref{sec:rmvu}), 
sparse PCA
(Section~\ref{sec:spca}),
and symmetric NMF (Section~\ref{sec:exp:snmf}).
Depending on the loss function and whether the matrix variate is factored,
the following SDP solvers will be compared:
\begin{enumerate}[leftmargin=*]
\item Solver for SDP problem \eqref{eq:new} (i.e., square loss and matrix variate is not factored):
\begin{enumerate}
\item \textit{FW} 
\cite{soren12}, which 
uses the Frank-Wolfe algorithm \cite{jaggi};
\end{enumerate}
\item Solvers for problem \eqref{eq:sdpfac} (i.e., square loss and factored matrix variate):
\begin{enumerate}[leftmargin=*]
\item \textit{nmAPG} \cite{nconv_apg15},
which uses the state-of-the-art accelerated gradient descent algorithm;
and
\item \textit{L-BFGS}
\cite{nocedal2006numerical}, 
which uses the popular quasi-Newton solver for smooth minimization problem.
\end{enumerate}
\item Solvers for SDP problem with $\ell_1$-loss, i.e., \eqref{eq:rsdpeq}:
\begin{enumerate}[leftmargin=*]
\item \textit{ADMM($\ell_1$)}, which
solves the nonsmooth but convex problem with ADMM
\cite{boyd2011distributed};
\item 
\textit{SDPLR}
\cite{burer2003}, which considers  (\ref{eq:rsdpeq})
and 
solves
with the augmented Lagrangian method;

\item \textit{SDPNAL+}
\cite{toh2015sdpnal+}, which also solves
(\ref{eq:rsdpeq})
but with the Newton-CG augmented Lagrangian method.
\end{enumerate}
\item Solvers for problem~\eqref{eq:rsdp} (i.e., $\ell_1$-loss and factored matrix variate):
\begin{enumerate}[leftmargin=*]
\item \textit{SDP-RL}($\ell_1$):
the proposed Algorithm~\ref{alg:rsdpmm},
and data sparsity is utilized as discussed in Section~\ref{ssec:spa};
\item \textit{SDP-RL-dense}, which is the same as \textit{SDP-RL}($\ell_1$)
except that data sparsity is not utilized;
\end{enumerate}
\item Solver for problem~\eqref{eq:rsdpnc}
(i.e., nonconvex loss and factored matrix variate):
\begin{enumerate}[leftmargin=*]
\item
\textit{SDP-RL(MCP)}, 
the proposed Algorithm~\ref{alg:rsdpmm:2}
which 
uses the leaky-MCP
loss in Table~\ref{tab:nloss}, with $\theta=5$ and $\eta=0.05$.
As reported in \cite{yao2016efficient,yao2018large,yao2018scalable}, the nonconvex losses in Table~\ref{tab:nloss} usually have similar performance.
\end{enumerate}
\end{enumerate}

For all the SDP-RL variants above,
ADMM is used as the solver for the convex surrogate.
We set
the maximum number of ADMM iterations to $1000$, and a tolerance
$\epsilon_k$
of $\max(10^{-8}, c_0/k^{b_0})$
as in Remark~\ref{rmk:eps}, where
$b_0=1.5$.

All these algorithms are implemented in Matlab.
Each of these is stopped when the relative change of objective values in successive
iterations is smaller than $10^{-5}$ or when the number of iterations reaches $2000$.
To reduce statistical variability, results are averaged over five repetitions.
Results for the best-performing method and those that are not significantly worse (according to the 
pairwise t-test with 95\% significance level)
are highlighted.
Experiments are run on a PC with a 3.07GHz CPU and 32GB RAM.

\subsection{PSD Matrix Completion}
\label{sec:exp:matcomp}

\begin{table*}[ht]
	\centering
	\caption{Testing RMSEs and CPU time (sec) on synthetic data with different 
	observation sampling
	ratios
	($s$).
	The number in brackets is the percentage of observed elements.}
	\setlength\tabcolsep{7pt}
	\vspace{-8px}
	\begin{tabular}{c | c |c  c | c  c | c  c | c c}
		\hline
		  \multirow{3}{*}{loss}   & \multirow{3}{*}{algorithm} &     \multicolumn{2}{c|}{$s=1$ ($1.90\%$)}      &     \multicolumn{2}{c|}{$s=2$ ($3.80\%$)}      &     \multicolumn{2}{c|}{$s=4$ ($7.60\%$)}      &     \multicolumn{2}{c}{$s=8$ ($15.20\%$)}      \\
		                          &                            & testing                  & CPU                 & testing                  & CPU                 & testing                  & CPU                 & testing                  & CPU                 \\
		                          &                            & RMSE                     & time                & RMSE                     & time                & RMSE                     & time                & RMSE                     & time                \\ \hline
		 \multirow{2}{*}{square}  & nmAPG                      & 0.896$\pm$0.008          & \textbf{17$\pm$3}   & 0.615$\pm$0.008          & \textbf{19$\pm$2}   & 0.442$\pm$0.003          & \textbf{21$\pm$2}   & 0.258$\pm$0.006          & \textbf{25$\pm$3}   \\
		                          & L-BFGS                     & 0.896$\pm$0.007          & \textbf{16$\pm$1}   & 0.615$\pm$0.007          & \textbf{17$\pm$2}   & 0.443$\pm$0.003          & \textbf{21$\pm$3}   & 0.256$\pm$0.007          & \textbf{27$\pm$4}   \\ \hline
		\multirow{2}{*}{$\ell_1$} & ADMM($\ell_1$)                      & 0.436$\pm$0.008          & $\!\!\!$1638$\pm$56 & 0.332$\pm$0.006          & $\!\!\!$2387$\pm$44 & 0.264$\pm$0.005          & $\!\!\!$3765$\pm$38 & 0.189$\pm$0.003          & $\!\!\!$5582$\pm$87 \\
		                          & SDP-RL($\ell_1$)             & 0.256$\pm$0.004          & 36$\pm$3            & 0.164$\pm$0.002          & 37$\pm$2            & 0.109$\pm$0.001          & 55$\pm$5            & 0.084$\pm$0.003          & 78$\pm$6            \\ \hline
		        leaky-MCP         & SDP-RL(MCP)                  & \textbf{0.168$\pm$0.001} & 50$\pm$3            & \textbf{0.113$\pm$0.001} & 46$\pm$2            & \textbf{0.077$\pm$0.002} & 67$\pm$6            & \textbf{0.053$\pm$0.002} & $\!\!\!$121$\pm$8   \\ \hline
	\end{tabular}
	\label{tab:obsv}
\end{table*}

\begin{table*}[ht]
	\centering 
	\caption{Testing RMSEs and CPU time (sec) on synthetic data with different fractions of outlying entries ($o$).}
	\setlength\tabcolsep{2.5pt}
	\vspace{-8px}
	\begin{tabular}{c | c | c  c | c  c | c  c | c  c | c c}
		\hline
		  \multirow{3}{*}{loss}   & \multirow{3}{*}{algorithm} &          \multicolumn{2}{c|}{$o=0$}          &         \multicolumn{2}{c|}{$o=0.025$}         &         \multicolumn{2}{c|}{$o=0.05$}          &            \multicolumn{2}{c|}{$o=0.10$}            &          \multicolumn{2}{c}{$o=0.20$}          \\
		                          &                            & testing                  & CPU               & testing                  & CPU                 & testing                  & CPU                 & testing                  & CPU                      & testing                  & CPU                 \\
		                          &                            & RMSE                     & time              & RMSE                     & time                & RMSE                     & time                & RMSE                     & time                     & RMSE                     & time                \\ \hline
		 \multirow{2}{*}{square}  & nmAPG                      & \textbf{0.005$\pm$0.001} & \textbf{10$\pm$2} & 0.228$\pm$0.003          & \textbf{17$\pm$2}   & 0.309$\pm$0.002          & \textbf{16$\pm$1}   & 0.422$\pm$0.003          & \textbf{16$\pm$3}        & 0.590$\pm$0.002          & \textbf{17$\pm$3}   \\
		                          & L-BFGS                     & \textbf{0.005$\pm$0.001} & \textbf{11$\pm$1} & 0.228$\pm$0.003          & \textbf{14$\pm$2}   & 0.309$\pm$0.002          & \textbf{15$\pm$1}   & 0.422$\pm$0.003          & \textbf{14$\pm$1}        & 0.590$\pm$0.002          & \textbf{15$\pm$3}   \\ \hline
		\multirow{2}{*}{$\ell_1$} & ADMM($\ell_1$)                      & 0.009$\pm$0.002          & 2846$\pm$123      & 0.192$\pm$0.003          & $\!\!\!$2873$\pm$83 & 0.199$\pm$0.003          & $\!\!\!$2870$\pm$62 & 0.222$\pm$0.002          & $\!\!$2893$\pm$146$\!\!$ & 0.269$\pm$0.002          & $\!\!\!$2869$\pm$41 \\
		                          & SDP-RL($\ell1_1$)            & 0.007$\pm$0.001 & 39$\pm$3          & \textbf{0.110$\pm$0.002} & 44$\pm$3            & \textbf{0.113$\pm$0.001} & 40$\pm$4            & 0.142$\pm$0.002          & 37$\pm$3                 & 0.161$\pm$0.002          & 34$\pm$2            \\ \hline
		        leaky-MCP         & SDP-RL(MCP)                  & 0.007$\pm$0.001 & 47$\pm$4          & \textbf{0.109$\pm$0.001} & 51$\pm$3            & \textbf{0.111$\pm$0.001} & 47$\pm$3            & \textbf{0.119$\pm$0.001} & 44$\pm$7                 & \textbf{0.134$\pm$0.001} & 41$\pm$3            \\ \hline
	\end{tabular}
	\label{tab:rat}
\end{table*}

\begin{table*}[ht]
	\centering
	\caption{Testing RMSEs and CPU time (sec) on synthetic data with different 
	maximum outlier
	amplitudes
	($\sigma$).}
	\setlength\tabcolsep{3pt}
	\vspace{-8px}
	\begin{tabular}{c | c | c  c | c  c | c  c | c  c}
		\hline
		  \multirow{3}{*}{loss}   & \multirow{3}{*}{algorithm}  &          \multicolumn{2}{c|}{$\sigma=2.5$}          &       \multicolumn{2}{c|}{$\sigma=5.0$}        &       \multicolumn{2}{c|}{$\sigma=10.0$}       &       \multicolumn{2}{c}{$\sigma=20.0$}        \\
		                          &                                & testing                  & CPU                      & testing                  & CPU                 & testing                  & CPU                 & testing                  & CPU                 \\
		                          &                         & RMSE                     & time                     & RMSE                     & time                & RMSE                     & time                & RMSE                     & time                \\ \hline
		 \multirow{2}{*}{square}  & nmAPG                           & 0.170$\pm$0.001          & \textbf{16$\pm$1}        & 0.309$\pm$0.002          & \textbf{16$\pm$1}   & 0.615$\pm$0.008          & \textbf{19$\pm$2}   & 1.36$\pm$0.01            & \textbf{22$\pm$1}   \\
		                          & L-BFGS                              & 0.170$\pm$0.001          & \textbf{15$\pm$3}        & 0.309$\pm$0.002          & \textbf{15$\pm$1}   & 0.615$\pm$0.007          & \textbf{17$\pm$2}   & 1.36$\pm$0.02            & \textbf{17$\pm$3}   \\ \hline
		\multirow{2}{*}{$\ell_1$} & ADMM($\ell_1$)                       & 0.191$\pm$0.002          & $\!\!$2868$\pm$141$\!\!$ & 0.199$\pm$0.003          & $\!\!\!$2870$\pm$62 & 0.332$\pm$0.006          & $\!\!\!$2837$\pm$84 & 0.418$\pm$0.008          & $\!\!\!$2906$\pm$45 \\
		                          & SDP-RL($\ell_1$)               & \textbf{0.114$\pm$0.001} & 49$\pm$7                 & \textbf{0.113$\pm$0.001} & 40$\pm$4            & 0.164$\pm$0.002          & 37$\pm$2            & 0.183$\pm$0.002          & 35$\pm$3            \\ \hline
		        leaky-MCP         & SDP-RL(MCP)                          & \textbf{0.113$\pm$0.001} & 60$\pm$6                 & \textbf{0.111$\pm$0.001} & 47$\pm$3            & \textbf{0.113$\pm$0.001} & 46$\pm$2            & \textbf{0.112$\pm$0.001} & 43$\pm$2            \\ \hline
	\end{tabular}
	\label{tab:amp}
\end{table*}

In this section, experiments are performed on 
PSD matrix completion 
(Section~\ref{sec:matcomp})
in the context of recommender systems.
Following \cite{lin2017robust},
we mimic the love/hate attacks, and
some ratings 
in the synthetic data
are randomly set to the highest/lowest values.

The ground-truth matrix $M$ is generated as a low-rank matrix $VV^{\top}$,
where $V \in \mathbb{R}^{m \times r}$
with entries sampled i.i.d. from $\mathcal{N}(0,1)$.
This is then corrupted as
$M' = {M} + N + S$,
where $N$ is a noise matrix
and $S$ is a sparse matrix 
modeling  large outliers
(with 
$o$ 
being the fraction of 
nonzero entries).
The entries 
of $N$ are
sampled i.i.d. from $\mathcal{N}(0, 0.1)$, while
the nonzero entries 
of $S$ 
are sampled uniformly from $\lbrace -\sigma,\sigma \rbrace$.
We randomly draw $\frac{1}{m}s r\log(m)\%$ of the elements from 
$M'$
as (noisy) observations for training, 
where
$s$ controls the sampling ratio.
Half of the remaining uncorrupted entries in $M$ are used
for validation 
(hyper-parameter tuning)
and the rest for testing. 
We experiment with matrix size
$m\in \{ 500$, $1000$, $1500$, $2000 \}$,
and set the ranks for all factorization-based methods to the ground-truth (i.e., 5). 
The other parameters are set as $o = 0.05$, $s = 2$ and $\sigma=10$. 

Let
$X X^{\top}$ be the matrix recovered and $M$ be the clean ground-truth matrix.
For performance evaluation,
we use (i) the testing 
root mean squared error
(RMSE):
$\sqrt{\frac{1}{\NM{\Omega_{\text{test}}}{1}} \sum\nolimits_{(i,j) \in \Omega_{\text{test}}}
	(M_{ij}-(X X^{\top})_{ij})^2}$; and
(ii) CPU time.

\subsubsection{Results}
\label{sec:results}

The testing RMSEs and CPU time are in Table~\ref{tab:size}.
Convergence of 
the testing RMSE versus CPU time is  
in 
Figure~\ref{fig:syn:rmse}.
Though 
methods 
based  on
the square loss 
(\textit{FW}, \textit{nmAPG}, and \textit{L-BFGS}) are very fast, 
they have much higher testing RMSE's than 
methods based on the $\ell_1$ and nonconvex losses.
In particular, \textit{FW} 
yields a much larger testing RMSE
than 
\textit{nmAPG} and
\textit{L-BFGS}.
This is because
\textit{FW} 
does not explicitly utilize low-rank factorization but
relies only on the nuclear-norm regularizer.
Moreover, it 
uses rank-one update in each iteration, and
is only as fast as \textit{nmAPG} and \textit{L-BFGS}.
Thus, 
\textit{FW} will not be included
in the sequel. 

Among algorithms based on the $\ell_1$-loss, 
\textit{SDP-RL}($\ell_1$) is the fastest as it exploits data sparsity.
\textit{SDP-RL(MCP)} yields slightly lower RMSE, but is slightly slower than
\textit{SDP-RL}($\ell_1$).
As \textit{SDPLR} and \textit{SDPNAL+} 
have comparable accuracies with
\textit{ADMM($\ell_1$)}, 
but are much slower and even fail to converge 
on large-scale problems.
Thus, 
\textit{SDPLR} and \textit{SDPNAL+} will also be dropped in the subsequent comparisons.

\subsubsection{Varying the Number of Observed Entries}
\label{sec:nobs}

We fix the matrix dimension $m = 2000$, outlier ratio $o = 0.05$, 
outlier amplitude $\sigma=10$, 
and vary the sampling ratio $s$ in $\{ 1$, $2$, $4$, $8 \}$.
A larger $s$ means that more elements are observed.
Table~\ref{tab:obsv} shows the testing RMSEs and CPU time.
When $s$ increases,  the
testing RMSE decreases and 
CPU time increases in general,
which agrees with intuition.

\subsubsection{Varying the Number of Outlying Entries}
\label{sec:noutlier}

We vary 
the fraction
of entries 
$o$
in the sparse noise matrix $S$
in $\{ 0,0.025, 0.05, 0.1, 0.2 \}$. 
The other parameters are set as $m = 2000$, $r = 2$, $s = 2$ and $\sigma=5$. 
Results are shown in Table~\ref{tab:rat}.
When there is no outlier ($o= 0$),
nmAPG and L-BFGS
perform the best,
as they
use the square loss which matches with the Gaussian noise generated.
As $o$ increases,
the testing RMSEs of all algorithms increase 
as expected.
Moreover, using the nonconvex loss 
leads to more robust results than both the square loss and $\ell_1$ loss,
particularly when
the
noise 
is large.

\subsubsection{Varying the Magnitude of Outlying Entries}

We vary 
the magnitude 
$\sigma$
of outlying entries
in $\lbrace 2.5, 5, 10, 20 \rbrace$.
The other parameters are fixed at $m = 2000, r = 2, s = 2$ and $o=0.05$. 
Results are shown in Table~\ref{tab:amp}.
As $\sigma$ increases, the testing RMSEs of most algorithms also increase 
(as in Section~\ref{sec:noutlier}).
The only exception is SDP-RL(MCP), whose
loss remains almost unchanged. 
This again shows that SDP-RL(MCP) is more robust.

\subsubsection{Varying Tolerance for Subproblem}
\label{sec:inexact}

We  experiment with the 
termination criterion of the ADMM solver (in Algorithms~\ref{alg:admm} and \ref{alg:noncvx:2}).
We 
vary
$b_0$ in 
Remark~\ref{rmk:eps}
in
$\{ 1.25$, $1.5$, $2.0 \}$,
and $c_0 = R(X_0)$ so that the inexactness scales with the objective value.
The other parameters are fixed at $m=2000$, 
$r = 5$, $o = 0.25$, $s = 2$ and $\sigma=10$.

\begin{table*}[ht]
	\centering
	\caption{Testing RMSEs and CPU time (sec) in the robust NPKL experiment.}
	\vspace{-8px}
	\begin{tabular}{c | c | c  c | c  c }
		\hline
		\multirow{2}{*}{loss} & \multirow{2}{*}{algorithm} &   \multicolumn{2}{c|}{5\% flippd labels}    &   \multicolumn{2}{c}{10\% flipped labels}   \\
		                      &                            & testing RMSE           & CPU time           & testing RMSE           & CPU time           \\ \hline
		       square         & SimpleNPKL                 & 0.54$\pm$0.01          & $\!\!\!$407$\pm$24 & 0.60$\pm$0.01          & $\!\!\!$419$\pm$27 \\ \cline{2-6}
		                      & nmAPG                      & 0.31$\pm$0.01          & 7$\pm$2            & 0.35$\pm$0.01          & 8$\pm$1            \\
		                      & L-BFGS                     & 0.31$\pm$0.01          & \textbf{4$\pm$1}   & 0.35$\pm$0.01          & \textbf{4$\pm$1}   \\ \hline
		      $\ell_1$        & ADMM($\ell_1$)                      & 0.23$\pm$0.01          & $\!\!\!$775$\pm$24 & 0.29$\pm$0.01          & $\!\!\!$784$\pm$19 \\
		                      & SDP-RL($\ell_1$)             & 0.21$\pm$0.01          & 55$\pm$33          & 0.28$\pm$0.01          & 72$\pm$36          \\ \hline
		      leaky-MCP       & SDP-RL(MCP)                  & \textbf{0.19$\pm$0.02} & 67$\pm$27          & \textbf{0.26$\pm$0.01} & 88$\pm$27          \\ \hline
	\end{tabular}
	\label{tab:rnpkl:rmse}
\end{table*}

\begin{figure}[ht]
\centering
\subfigure[SDP-RL($\ell_1$).] {\includegraphics[width=0.24\textwidth]{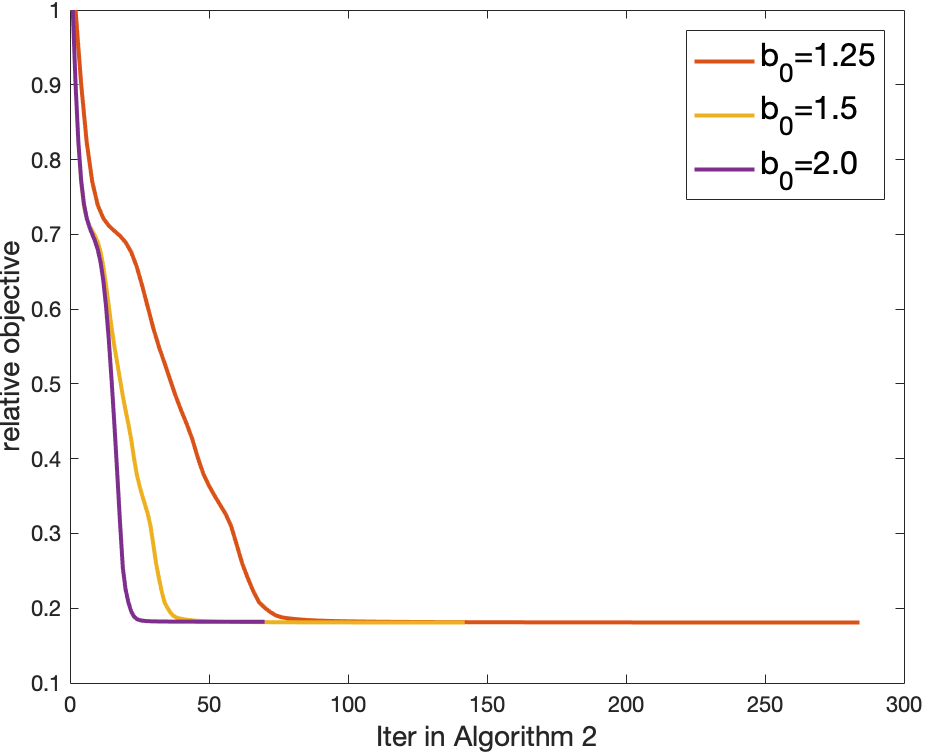}}
\subfigure[SDP-RL(MCP).]{\includegraphics[width=0.24\textwidth]{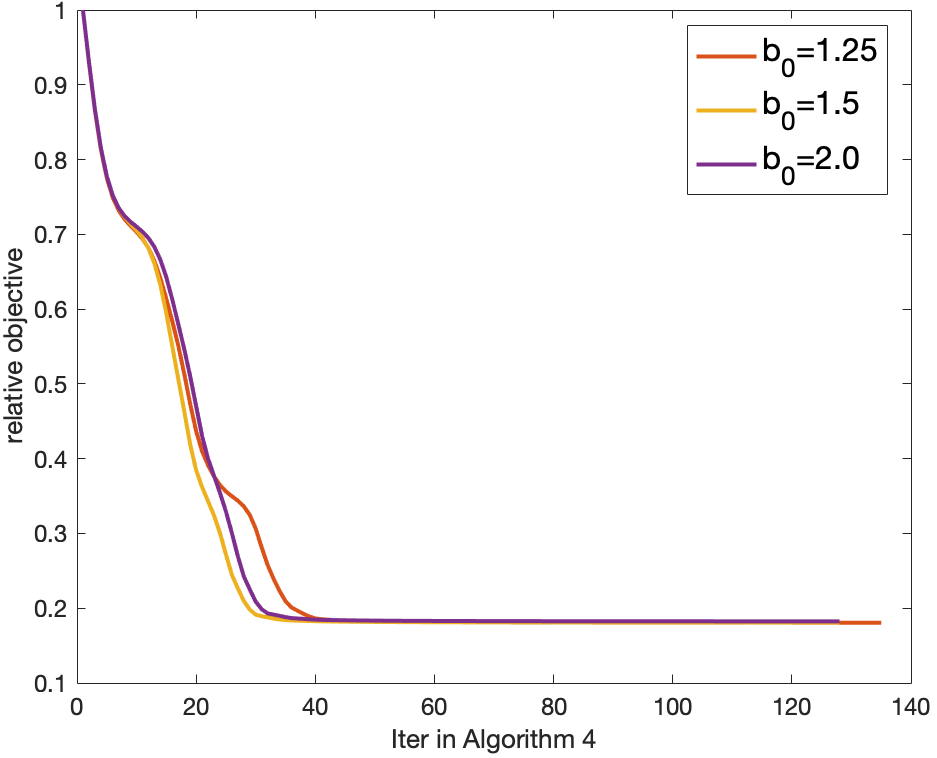}}

\vspace{-8px}
\caption{Convergence of relative objective vs number of iterations in Algorithms~\ref{alg:rsdpmm} and \ref{alg:rsdpmm:2} at different tolerance on inexactness.}
\label{fig:syn:iter}
\end{figure}

\begin{figure}[ht]
	\centering
	\subfigure[SDP-RL($\ell_1$).] {\includegraphics[width=0.24\textwidth]{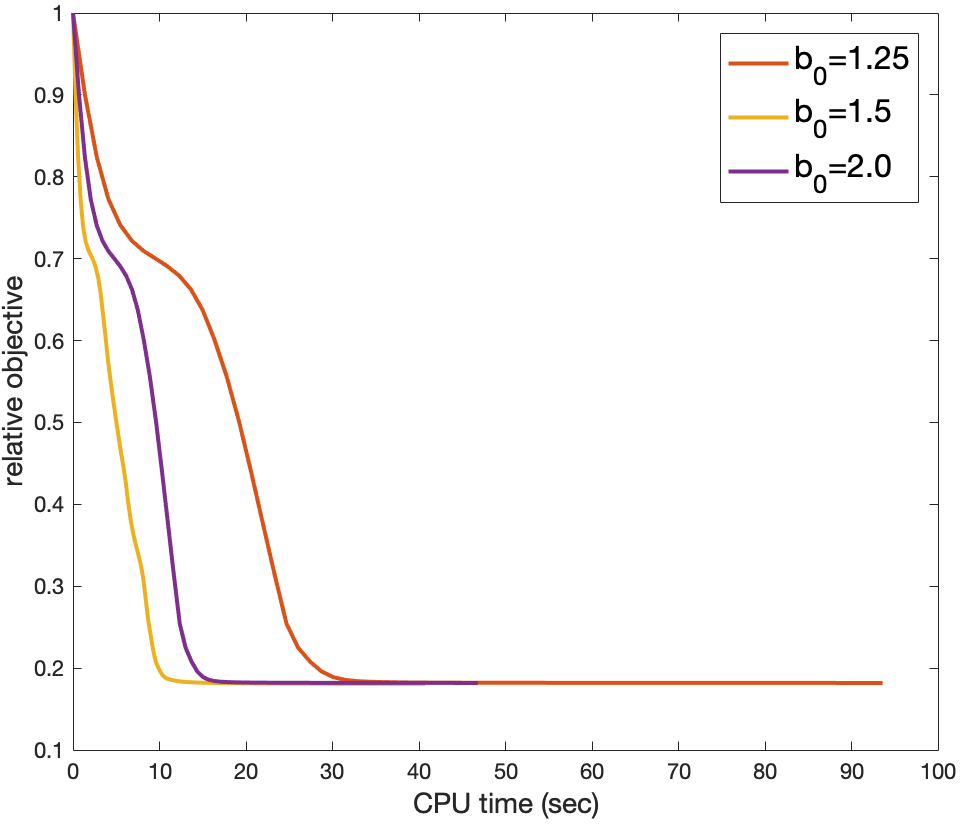}}
	\subfigure[SDP-RL(MCP).]{\includegraphics[width=0.24\textwidth]{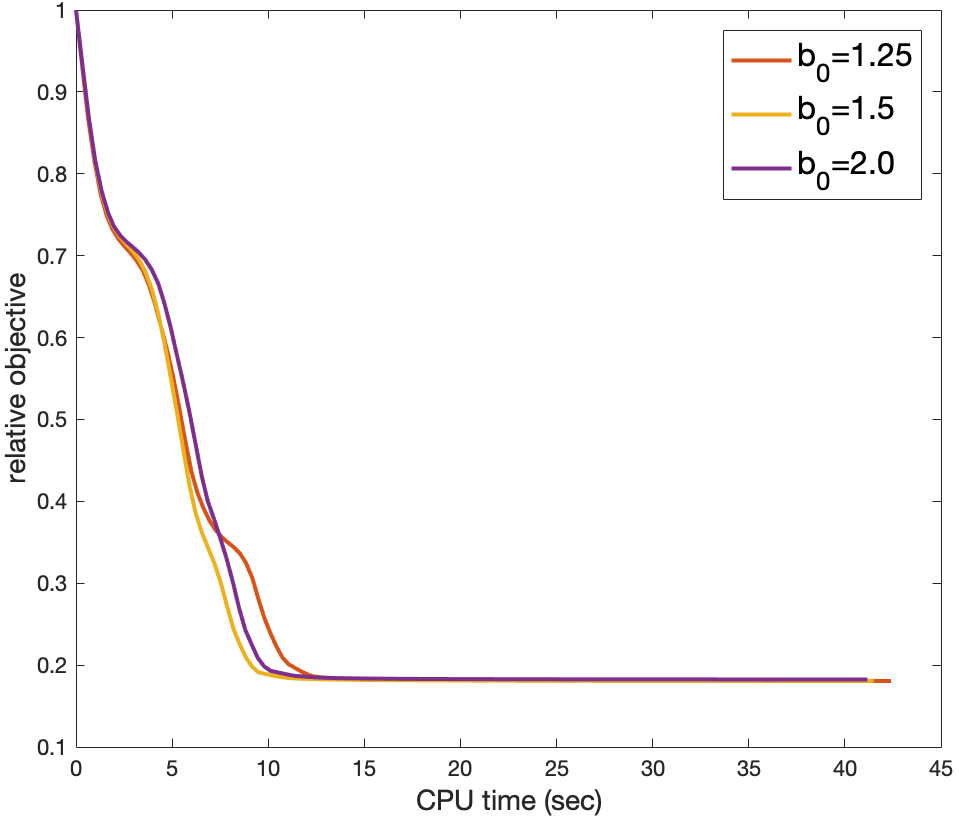}}

	\vspace{-8px}
	\caption{Convergence of relative objective vs CPU time (sec) at different tolerance on inexactness.}
	\label{fig:syn:time}
\end{figure}

Figure~\ref{fig:syn:iter} shows convergence of the 
relative objective $R(X_k)/R(X_0)$
vs the number of  iterations in
Algorithm~\ref{alg:rsdpmm} (resp.  Algorithm~\ref{alg:rsdpmm:2}) for SDP-RL($\ell_1$) (resp. SDP-RL(MCP)).
Recall that each iteration 
of Algorithm~\ref{alg:rsdpmm} (resp.  Algorithm~\ref{alg:rsdpmm:2}) makes one call
to Algorithm~\ref{alg:admm} (resp. 
Algorithm~\ref{alg:noncvx:2}).
As can be seen,
a larger $b_0$ (smaller tolerance) leads to fewer  iterations of
Algorithm~\ref{alg:rsdpmm} and Algorithm~\ref{alg:rsdpmm:2}.
However, solving the ADMM to such a higher precision 
means more 
time to solve the surrogate (Table~\ref{tab:suggtime}).
Figure~\ref{fig:syn:time} shows convergence w.r.t. the total CPU time. 
As can be seen,
$b_0 = 1.5$ is a good empirical choice,
and we will use this in the sequel.

\begin{table}[ht]
	\centering
	\caption{Average per-iteration CPU time (sec) of SDP-RL.}
	\vspace{-8px}
	\begin{tabular}{c | c | c | c}
		\hline
		               & $b_0=1.25$ & $b_0=1.5$ & $b_0=2.0$ \\ \hline
		SDP-RL($\ell_1$) & 0.15       & 0.29      & 0.47      \\ \hline
		  SDP-RL(MCP)    & 0.16       & 0.28      & 0.47      \\ \hline
	\end{tabular}
	\label{tab:suggtime}
\end{table}

\subsubsection{Effect of Different Initializations} 
\label{sec:init1}

In this experiment, we 
study the following two 
initializations
of $X$:
(i) zero initialization (i.e., $X_1=0$) as shown in 
Algorithms~\ref{alg:rsdpmm} and \ref{alg:rsdpmm:2};
and 
(ii) Gaussian initialization, in which elements of $X_1$ are independently sampled 
from the standard normal distribution. 
We randomly generate 5 Gaussian initializations. 
The other parameters are fixed at $m=2000$, 
$r = 5$, $o = 0.25$, $s = 2$ and $\sigma=10$.

\begin{figure}[ht]
	\centering
	\subfigure[Algorithm~\ref{alg:rsdpmm}. \label{fig:alg1}] 
	{\includegraphics[width=0.49\columnwidth]{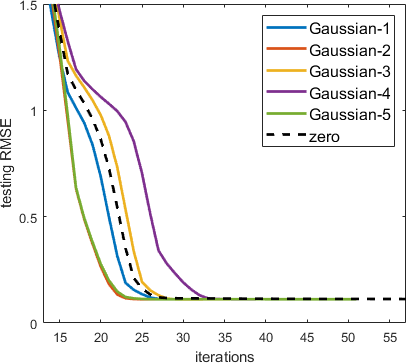}}
	\subfigure[Algorithm~\ref{alg:rsdpmm:2}. \label{fig:alg2}]
	{\includegraphics[width=0.49\columnwidth]{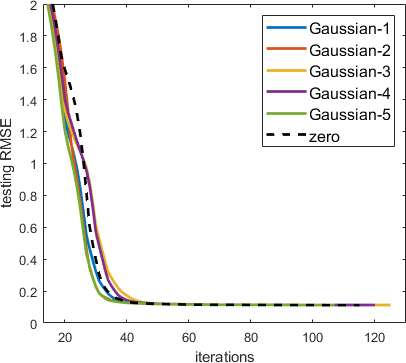}}
	\vspace{-8px}
	\caption{Convergence of testing RMSE versus different initializations for the proposed algorithms.}
	\label{fig:syn:inirm}
\end{figure}

Figure~\ref{fig:alg1}  (resp.
Figure~\ref{fig:alg2})
shows 
convergence of  
testing RMSE versus the number of 
iterations 
in Algorithm~\ref{alg:rsdpmm} 
for SDP-RL($\ell_1$) (resp.
number of 
iterations 
in Algorithm~\ref{alg:rsdpmm:2} 
for 
SDP-RL(MCP)).
As can be seen, all initializations lead to similar 
testing RMSEs.
Some initializations lead to faster convergence, but 
Gaussian initialization is not always better than
zero initialization.

\subsection{Robust NPKL}
\label{sec:exp:rnpkl}

In this section,
experiment is performed on the 
adult 
data set
``a1a",
which has been commonly used in the NPKL literature \cite{zhuang11}. 
It contains 
$\bar{n}=1605$
123-dimensional 
samples.
Following \cite{hu2019low},
we randomly sample $6 \bar{n}$ pairs and construct set 
$\T = \{ T_{ij} \}$,
where $T_{ij}=1$ if samples $i$ and $j$ have the same label, and $T_{ij}=0$
otherwise.
We 
then 
randomly sample $60\%$ of the pairs in $\T$ for training, $20\%$ for
validation and hyper-parameter tuning,
and the rest for testing. 
The numbers of must-link and cannot-link pairs in the training/validation/testing
sets are shown in Table~\ref{tab:link}.
The Laplacian $L$ in (\ref{eq:NPKL})
is constructed 
from a graph $G$.
Each node in $G$ corresponds to a training sample, and
is connected to 
its two nearest 
training samples
based on the distance in the feature space.

\begin{table}[ht]
\caption{Numbers of must-link and cannot-link pairs in the robust NPKL experiment.}
\centering
\vspace{-8px}
\begin{tabular}{c | c | c}
	\hline
	& must-link & cannot-link \\ \hline
	training & 3637 & 2141\\ \hline
	validation & 1210 & 716 \\ \hline
	testing & 1220 & 706  \\ \hline
\end{tabular}
\label{tab:link}
\end{table}

To test the robustness of NPKL algorithms,
we flip some 
must-link constraints in the training set to cannot-link constraints, and vice versa.
This mimics the 
label flipping 
attacks in real-world applications
\cite{raykar2010learning}.
The total number of constraints flipped is varied in $\{5\%, 10\%\}$.

Besides comparing with the previous methods based on the square loss, $\ell_1$ loss and leaky-MCP
loss,
we also compare with 
\textit{SimpleNPKL}~\cite{zhuang11},
which is based on the square loss
but does not use the low-rank factorization.
As for the rank $r$ of the initial solution $X$, 
we follow \cite{burer2003} and set its value to be the largest $r$ satisfying $\frac{r(r+1)}{2}\le |\T|$. 
For performance evaluation, 
we follow \cite{lin2017robust,yao2018scalable} and use the (i) testing root mean square error, 
$\text{RMSE}=(\sum\nolimits_{(i,j) \in \T_{\text{test}}} (Z_{ij}-T_{ij})^2/|\T_{\text{test}}|)^{1/2}$, 
where $\bar{X}$ is the output of the algorithm
and $\T_{\text{test}}$ is the testing set,
and (ii) CPU time.

\begin{figure}[ht]
	\centering
	\subfigure[5\% flipped labels.]
	{\includegraphics[width=0.240\textwidth]{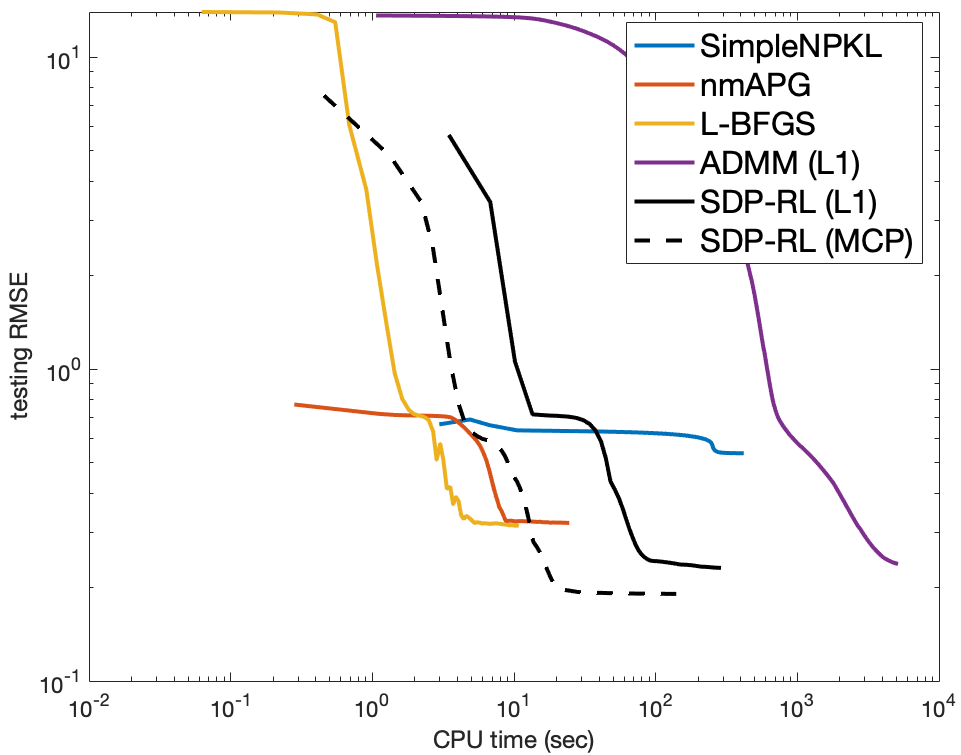}}
	\subfigure[10\% flipped labels.]
	{\includegraphics[width=0.240\textwidth]{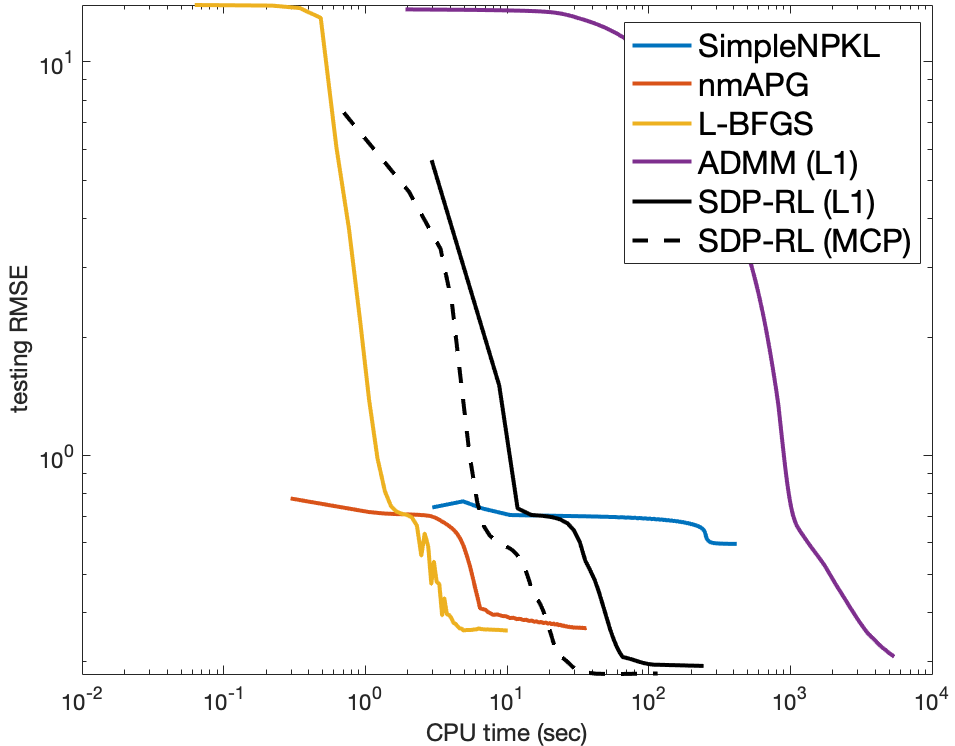}}
	
	\vspace{-8px}
	\caption{Convergence of testing RMSE vs CPU time (sec) in the robust NPKL experiment.}
	\label{fig:npkl:obj}
\end{figure}

\begin{table*}[ht]
	\centering
	\caption{Testing RMSEs and CPU time (sec) in the robust CMVU experiment.}
	\vspace{-8px}
	\begin{tabular}{c | c | c  c | c  c | c  c}
		\hline
		\multirow{2}{*}{loss} & \multirow{2}{*}{algorithm} &    \multicolumn{2}{c|}{small deviations}     &   \multicolumn{2}{c|}{5\% large outliers}    &   \multicolumn{2}{c}{10\% large outliers}    \\
		                      &                            & testing RMSE           & CPU time            & testing RMSE           & CPU time            & testing RMSE           & CPU time            \\ \hline
		       square         & SimpleCMVU                 & 0.48$\pm$0.02          & $\!$837$\pm$19      & 0.77$\pm$0.03          & $\!\!\!$1675$\pm$49 & 0.97$\pm$0.03          & $\!\!\!$1263$\pm$33 \\
		                      & nmAPG                      & 0.34$\pm$0.01          & $\!\!\!$342$\pm$5   & 0.65$\pm$0.04          & 691$\pm$15          & 0.76$\pm$0.01          & $\!\!\!$280$\pm$2   \\
& L-BFGS                     & 0.34$\pm$0.01          & $\!\!\!$424$\pm$7   & 0.46$\pm$0.01          & 645$\pm$15          & 0.58$\pm$0.01          & $\!\!\!$574$\pm$2   \\ \hline
$\ell_1$       & ADMM($\ell_1$)                      & 0.30$\pm$0.03          & $\!\!\!$3090$\pm$27 & 0.34$\pm$0.03          & $\!\!\!$2944$\pm$23 & 0.35$\pm$0.03          & $\!\!\!$3124$\pm$29 \\
		                      & SDP-RL($\ell_1$)             & 0.29$\pm$0.02          & \textbf{165$\pm$23} & 0.32$\pm$0.03          & \textbf{113$\pm$50} & 0.33$\pm$0.02          & \textbf{113$\pm$33} \\ \hline
		      leaky-MCP       & SDP-RL(MCP)                  & \textbf{0.25$\pm$0.02} & 206$\pm$38          & \textbf{0.29$\pm$0.02} & 156$\pm$53          & \textbf{0.30$\pm$0.03} & 162$\pm$40          \\ \hline
	\end{tabular}
	\label{tab:rmvu:rmse}
\end{table*}

\begin{figure*}[ht]
	\centering
	\subfigure[small deviations.]
	{\includegraphics[width=0.245\textwidth]{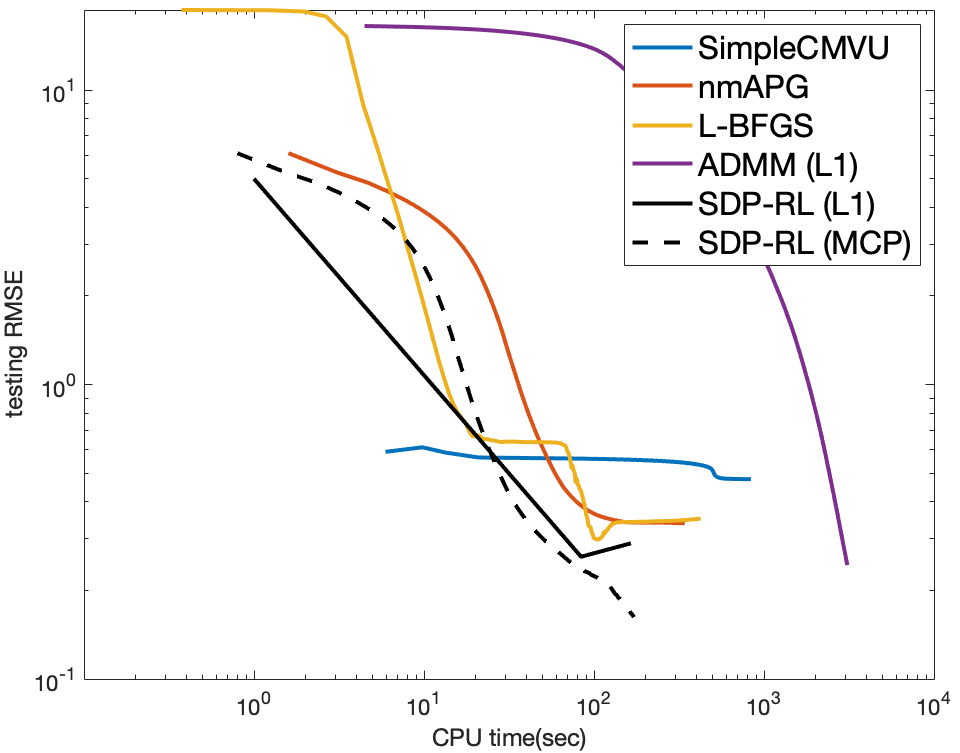}}
	\quad
	\subfigure[5\%  large outliers.]
	{\includegraphics[width=0.245\textwidth]{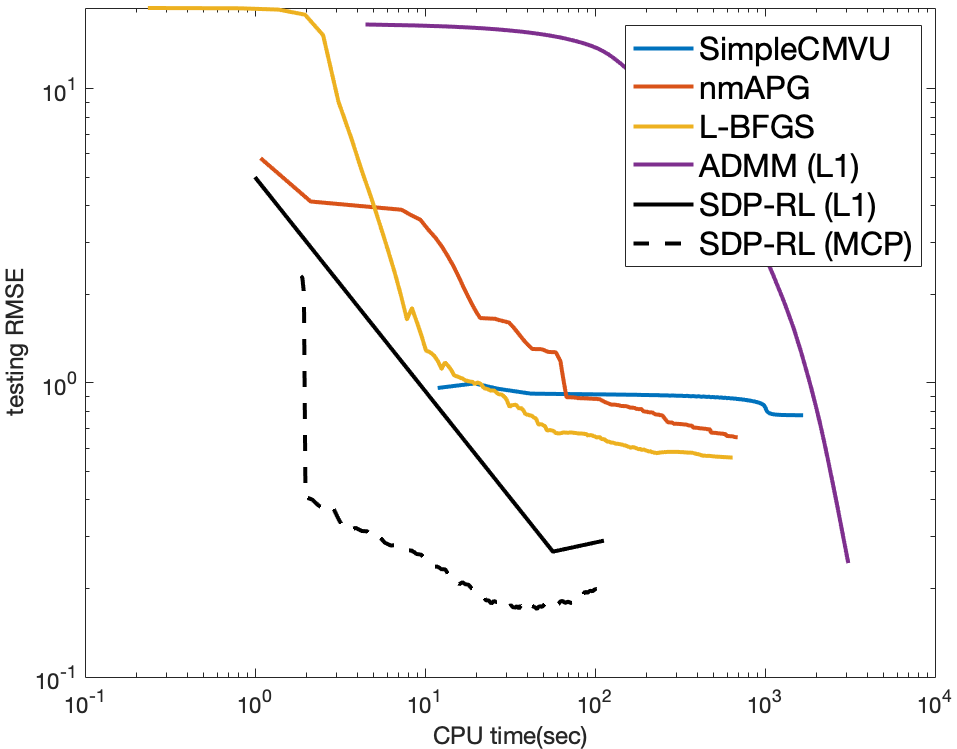}}
	\quad
	\subfigure[10\%  large outliers.]
	{\includegraphics[width=0.245\textwidth]{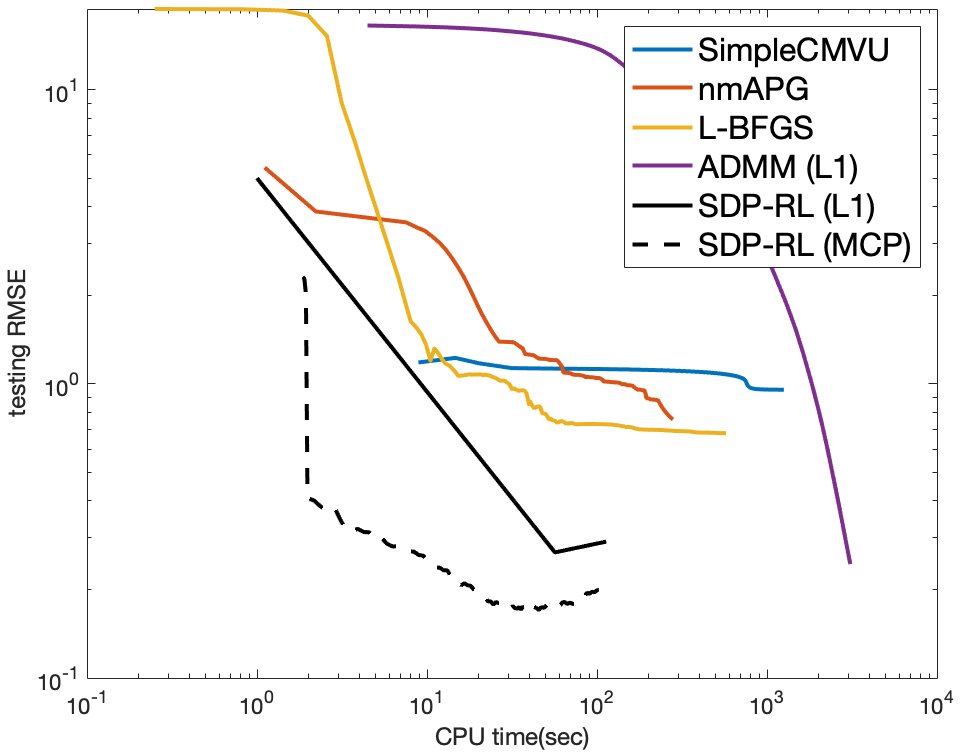}}
	
	\vspace{-8px}
	\caption{Convergence of testing RMSE vs CPU time (sec) 
		in the robust CMVU experiment.}
	\label{fig:cmvu:obj}
\end{figure*}

Table~\ref{tab:rnpkl:rmse} and Figure~\ref{fig:npkl:obj}
show performance of the compared algorithms. 
As can be seen, 
algorithms based on the $\ell_1$ and non-convex losses have lower
testing RMSE's than those 
based on
the square loss, with
\textit{SDP-RL(MCP)} being the best. Moreover,
\textit{SDP-RL}($\ell_1$) and \textit{SDP-RL(MCP)} are faster than
\textit{ADMM($\ell_1$)}.

\subsection{Robust CMVU}
\label{sec:rmvu}

In this section, we perform
experiment
on robust CMVU
using the commonly-used {USPS} data set,
which contains 2007 256-dimensional samples. 
As in \cite{song08}, 
the set 
$\Omega$
in (\ref{eq:cmvu})
is constructed 
by using the nearest $1\%$ neighbors of each sample, leading to $|\Omega|=401,400$. 
The squared Euclidean distance $t_{ij}$
for each $(i,j) \in \Omega$
is computed 
from the clean data set (before adding noise).
We randomly sample $60\%$ of the pairs from $\Omega$ for training, $20\%$ for
validation and hyper-parameter tuning, and the
rest for testing. 
As for the rank $r$ of the initial solution $X$, we follow \cite{burer2003} and set its value to the largest $r$ satisfying $r(r+1)/2 \le |\Omega|$. 
The tradeoff parameter $\g$ 
in (\ref{eq:cmvu})
is fixed at $0.01$.

For performance evaluation, 
we use (i) the testing 
$\text{RMSE}$: $(\sum\nolimits_{(i,j) \in \Omega_{\text{test}}} \big( Z_{ii} \! + \! Z_{jj} \! - \! 2 Z_{ij} \! - \! t_{ij} \big)^2 / |\Omega_{\text{test}}|)^{\frac{1}{2}}$, where
$\Omega_{\text{test}}$ is the testing portion of 
$\Omega$,
and (ii) CPU time.
Since NPKL and CMVU can be solved using the same algorithm, 
we use the same baselines as in Section~\ref{sec:exp:rnpkl}, i.e. SimpleCMVU \cite{zhuang11}.

\begin{figure*}[ht]
	\centering
	\subfigure[$r = 50$.] {\includegraphics[width=0.245\textwidth]{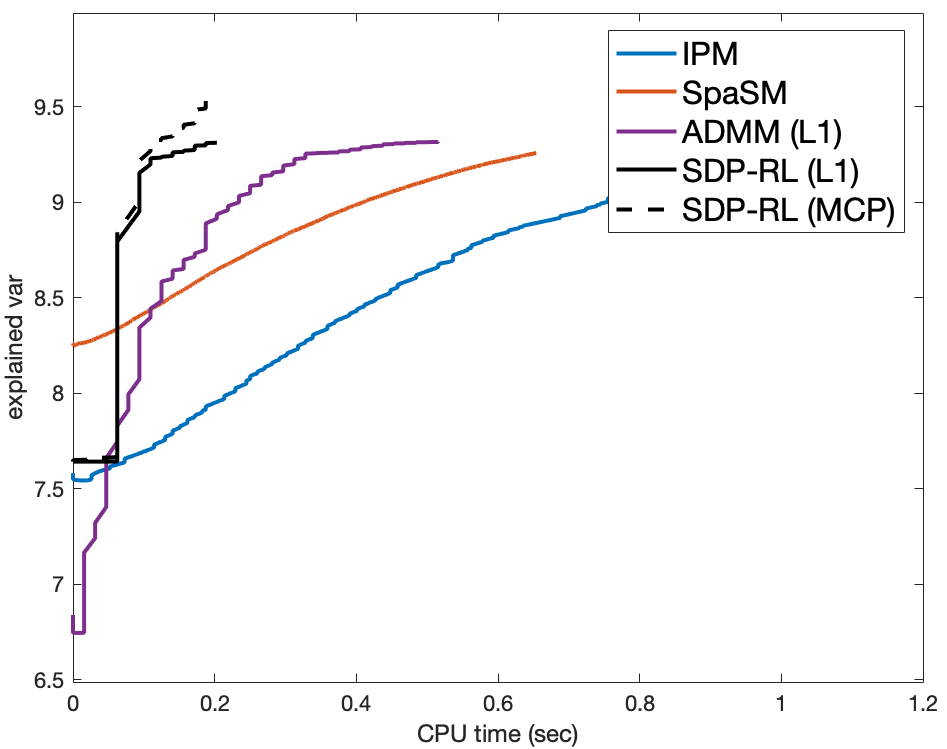}}
	\quad
	\subfigure[$r = 100$.]{\includegraphics[width=0.241\textwidth]{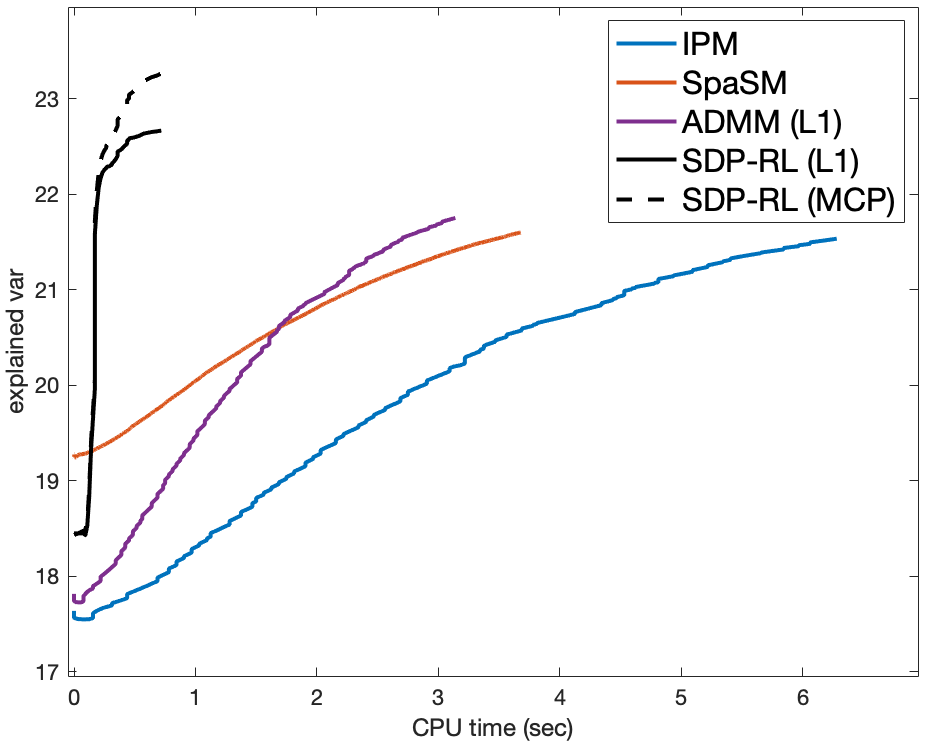}}
	\quad
	\subfigure[$r = 200$.] {\includegraphics[width=0.242\textwidth]{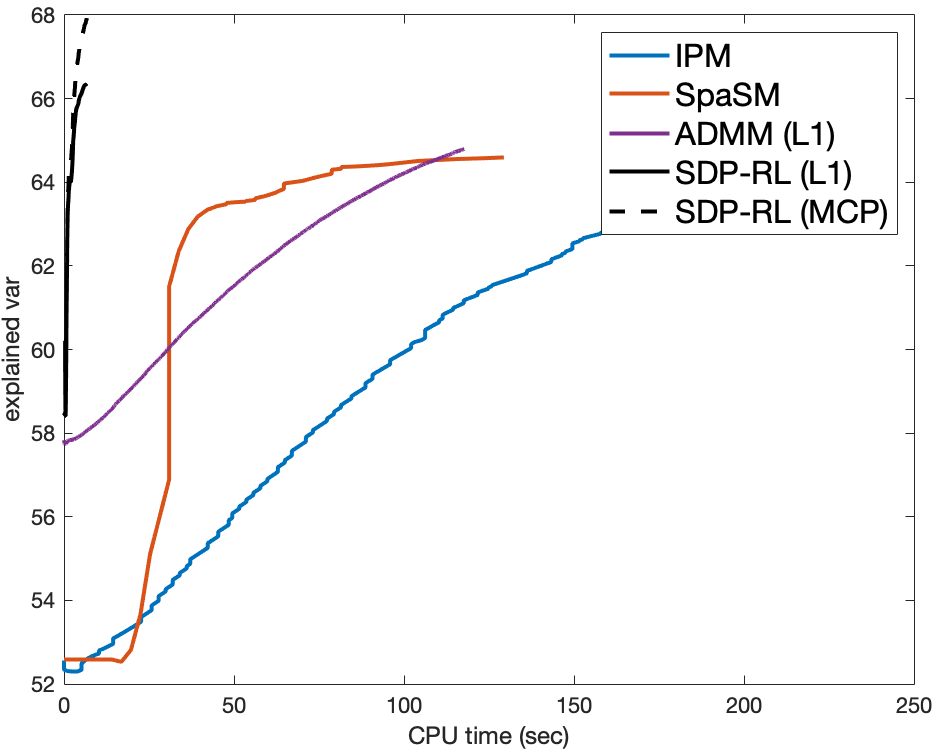}}
	
	\vspace{-8px}
	\caption{Percentage of explained variance vs CPU time (sec) for the various algorithms 
		on the sparse PCA problem.}
	\label{fig:spca:var}
	\vspace{-10px}
\end{figure*}

\subsubsection{Small Gaussian Noise }

Here,
we add
Gaussian noise from $\mathcal{N}(0, 0.01 \cdot \text{Var}(x))$
to each feature 
in the training set,
where $\text{Var}(x)$ is a vector containing the variance of each feature.
Table~\ref{tab:rmvu:rmse} and Figure~\ref{fig:cmvu:obj}
show the
results.
The observations
are almost the same as that in Section~\ref{sec:exp:rnpkl}.
\textit{SDP-RL(MCP)}
has the lowest testing RMSE,
while \textit{ADMM($\ell_1$)} and \textit{SDP-RL}($\ell_1$) are better than 
\textit{nmAPG} and \textit{L-BFGS}. 
\textit{SDP-RL}($\ell_1$) is also much more efficient than \textit{ADMM($\ell_1$)}.

\subsubsection{Large Outliers}

In this experiment,
we add large outliers which may appear in the data~\cite{lin2017robust,yao2018scalable}.
First, we randomly sample 
some samples
($5\%$ and $10\%$) from the 
training set.
For each selected sample $x_i$,
we 
add 
random noise
from $\mathcal{N}(0, 5 \tilde{x})$, where $\tilde{x}$ is a vector containing the largest 
absolute feature value for that dimension.
Table~\ref{tab:rmvu:rmse} and Figure~\ref{fig:cmvu:obj} show the performance against outliers. 
Again,
\textit{SDP-RL(MCP)}
has the lowest testing RMSE among the algorithms. Moroever,
\textit{SDP-RL}($\ell_1$) is much faster than \textit{ADMM($\ell_1$)}.

\subsection{Sparse PCA}
\label{sec:spca}

In this section, we perform 
sparse PCA
experiment on
the 
colon cancer data set
\cite{soren12}, which contains $2000$ micro-array
readings from $62$ subjects. 
We 
set $\l=0,
\g=10$
in  \eqref{eq:spca:trans}, and
try different embedding dimensions $r = \{ 50$, $100$, $200\}$.
As there are no missing data in sparse PCA, data sparsity is not utilized for SDP-RL.
We also compare with 
two state-of-the-art sparse PCA methods:
\begin{enumerate}[leftmargin = *]
\item 
nonlinear 
IPM~\cite{hein2010inverse}, which obtains 
the sparse principal components
from the following inverse
eigenvalue problem:
$\min_{x \in \mathbb{R}^n} 
\frac{(1-\alpha) \| x \|_2 + \alpha \| x \|_1}{ x^{\top} \Sigma x }$,
where 
$\alpha$ is a hyper-parameter controlling the sparsity of $x$. When $\alpha=0$,
it reduces to original PCA. 

\item
SpaSM~\cite{sjostrand2018spasm}, 
which solves the sparse PCA problem in \eqref{eq:spca:trans} with the SPCA algorithm in \cite{zou2018selective}.
\end{enumerate}
For performance evaluation,
as in \cite{soren12,d:spca07},
we use the (i) CPU time, (ii)
sparsity of $X X^{\top}$
(i.e., ratio of zero
elements), 
and (iii) explained variance
(i.e., $\tr(Z\Sigma)$ in \eqref{eq:spca}).
Experiments are repeated five times.

\subsubsection{Results}

Results are shown in 
Table~\ref{tab:spca}.
As can been seen, 
due to the use of the non-convex loss,
\textit{SDP-RL(MCP)} produces the best solution compared with the other approaches.
Besides,
both \textit{SDP-RL}($\ell_1$) and \textit{SDP-RL(MCP)} are much faster than \textit{ADMM($\ell_1$)}, \textit{SpaSM} and \textit{nonlinear IPM}.
Figure~\ref{fig:spca:var} shows
convergence of the explained variance with CPU time.
As can be seen, 
\textit{SDP-RL}($\ell_1$) and \textit{SDP-RL(MCP)} also converge much faster than the other approaches.

\begin{table}[ht]
	\centering
	\caption{Performance of various sparse PCA algorithms on the colon cancer data set.}
	\vspace{-8px}
	\begin{tabular}{c | c | c | c | C{40px}}
		\hline
		$r$ & algorithm        & CPU time (sec)     & sparsity  & explained variance \\ \hline
		50  & nonlinear IPM    & 1.06$\pm$0.12      & 0.73      & 8.98               \\
		    & SpaSM            & 0.64$\pm$0.03      & 0.63      & 8.92               \\
		    & ADMM($\ell_1$)   & 0.55$\pm$0.06      & \hl{0.76} & 9.23               \\
		    & SDP-RL($\ell_1$) & \hl{0.21$\pm$0.02} & \hl{0.76} & 9.23               \\
		    & SDP-RL(MCP)      & \hl{0.23$\pm$0.02} & \hl{0.76} & \hl{9.58}          \\ \hline
		100 & nonlinear IPM    & 6.18$\pm$0.36      & 0.75      & 21.83              \\
		    & SpaSM            & 3.49$\pm$0.23      & 0.67      & 21.87              \\
		    & ADMM($\ell_1$)   & 3.12$\pm$0.25      & \hl{0.79} & 21.86              \\
		    & SDP-RL($\ell_1$) & \hl{0.75$\pm$0.07} & \hl{0.79} & 22.67              \\
		    & SDP-RL(MCP)      & 0.86$\pm$0.12      & \hl{0.79} & \hl{23.22}         \\ \hline
		200 & nonlinear IPM    & 244.36$\pm$20.68   & 0.79      & 60.18              \\
		    & SpaSM            & 120.94$\pm$8.26    & 0.75      & 62.74              \\
		    & ADMM($\ell_1$)   & 118.28$\pm$12.25   & \hl{0.82} & 64.24              \\
		    & SDP-RL($\ell_1$) & \hl{7.42$\pm$0.23} & \hl{0.82} & 66.44              \\
		    & SDP-RL(MCP)      & 7.68$\pm$0.35      & \hl{0.82} & \hl{67.92}         \\ \hline
	\end{tabular}
	\label{tab:spca}
\end{table}

\subsubsection{Effect of Different Initializations} 

In this experiment, we 
study the following two 
initializations
of $X$:
(i)
zero initialization as in Algorithm~\ref{alg:rsdpmm} and \ref{alg:rsdpmm:2};
and (ii) standard PCA.
Results are shown in Table~\ref{tbl:new1}.
As can be seen,
different initializations have little impact on the final performance,
but initialization by PCA can have faster convergence.
This also agrees with 
the common practice 
of using PCA as initialization 
for sparse PCA~\cite{zou2018selective}.

\begin{table}[ht]
	\centering
	\caption{Effect of different ways to initialize SDP-RL in the sparse PCA experiment. }
	\label{tbl:new1}
	\vspace{-8px}
	\setlength\tabcolsep{3pt}
	\begin{tabular}{c|C{40px}|c|C{40px}|c|C{40px}}
		\hline
		        $r$          &      loss in SDP-RL       & initialization &     CPU time (sec)     &   sparsity    & explained variance \\ \hline
		\multirow{4}{*}{50}  & \multirow{2}{*}{$\ell_1$} &      zero      &     0.21$\pm$0.02      &    {0.76}     &       {9.23}       \\
		                     &                           &      PCA       & \textbf{0.11$\pm$0.01} & \textbf{0.75} &   \textbf{9.26}    \\
		   \cline{2-6}    &   \multirow{2}{*}{MCP}    &      zero      &     0.23$\pm$0.02      & \textbf{0.76} &   \textbf{9.58}    \\
		                     &                           &      PCA       & \textbf{0.11$\pm$0.02} &    {0.78}     &       {9.57}       \\ \hline
		\multirow{4}{*}{100} & \multirow{2}{*}{$\ell_1$} &      zero      &     0.75$\pm$0.07      & \textbf{0.79} &      {22.67}       \\
		                     &                           &      PCA       & \textbf{0.29$\pm$0.04} &    {0.80}     &   \textbf{22.78}   \\
		   \cline{2-6}    &   \multirow{2}{*}{MCP}    &      zero      &     0.86$\pm$0.12      & \textbf{0.79} &      {23.22}       \\
		                     &                           &      PCA       & \textbf{0.35$\pm$0.07} & \textbf{0.79} &   \textbf{23.24}   \\ \hline
		\multirow{4}{*}{200} & \multirow{2}{*}{$\ell_1$} &      zero      &     7.42$\pm$0.23      &    {0.82}     &      {66.44}       \\
		                     &                           &      PCA       & \textbf{4.14$\pm$0.19} & \textbf{0.81} &   \textbf{66.48}   \\
		   \cline{2-6}    &   \multirow{2}{*}{MCP}    &      zero      &     7.68$\pm$0.35      & \textbf{0.82} &      {67.92}       \\
		                     &                           &      PCA       & \textbf{4.35$\pm$0.23} & \textbf{0.82} &   \textbf{68.03}   \\ \hline
	\end{tabular}
\end{table}

As can be seen from 
experiments in both Section~\ref{sec:init1} and here,
the choice of initialization is application-specific.
Empirically,
different initializations have little impact on the final performance of the
proposed algorithm,
but a better initialization can lead to faster convergence.

\subsection{Symmetric NMF}
\label{sec:exp:snmf}

In this section, we perform experiments on symmetric nonnegative matrix factorization (SNMF). 
Data generation is similar to that in Section~\ref{sec:exp:matcomp},
with the ground-truth matrix $M$ generated as $VV^{\top}$.
In the first experiment,
$V \in \mathbb{R}^{m \times 5}$
is synthetic,
with $m\in \{1000, 2000 \}$.
Each element of $V$ is sampled independently 
from the standard exponential distribution. 
We 
then
corrupt $M$ by adding a sparse matrix $S$, 
which models a fraction of $o$ 
large outliers sampled uniformly from $\lbrace 0, \sigma \rbrace$, 
to obtain $M' = M + S$. 
The 
training/validation/test set split
follows that in Section~\ref{sec:exp:matcomp}.
The second experiment is similar, except that
$V$ is constructed from real-world data set.
Specifically,
following~\cite{Kuang:2015aa},
we construct
$V \in \mathbb{R}^{2007 \times 10}$
as the one-hot label matrix for 
the USPS dataset
in Section~\ref{sec:rmvu}.

The rank $r$ of the initial $X$ solution is 
set to the ground-truth, i.e. $5$ for the esynthetic data and $10$ for USPS dataset.  
The other parameters are set as $o = 0.05$, $s = 2$ and $\sigma=10$. 

We compare \textit{SDP-RL} with three 
commonly-used SNMF
methods 
\cite{Kuang:2015aa, shi2017inexact}:
Newton's method (\textit{Newton})~\cite{he2011symmetric}, regularized eigenvalue decomposition (\textit{rEVD})~\cite{huang2014nonnegative}, 
and block-coordinate descent (\textit{BCD})~\cite{shi2017inexact}.
All three solve \eqref{eq:snmf}
(with the square loss)
while the proposed \textit{SDP-RL} solves problem~\eqref{eq:snmf2} (with the
$\ell_1$-loss).
For performance evaluation,
we follow Section~\ref{sec:exp:matcomp} 
and use the 
testing RMSE and CPU time. 

\begin{figure*}[ht]
	\centering
	\subfigure[synthetic ($m=1000$).]
	{\includegraphics[width=0.250\textwidth]{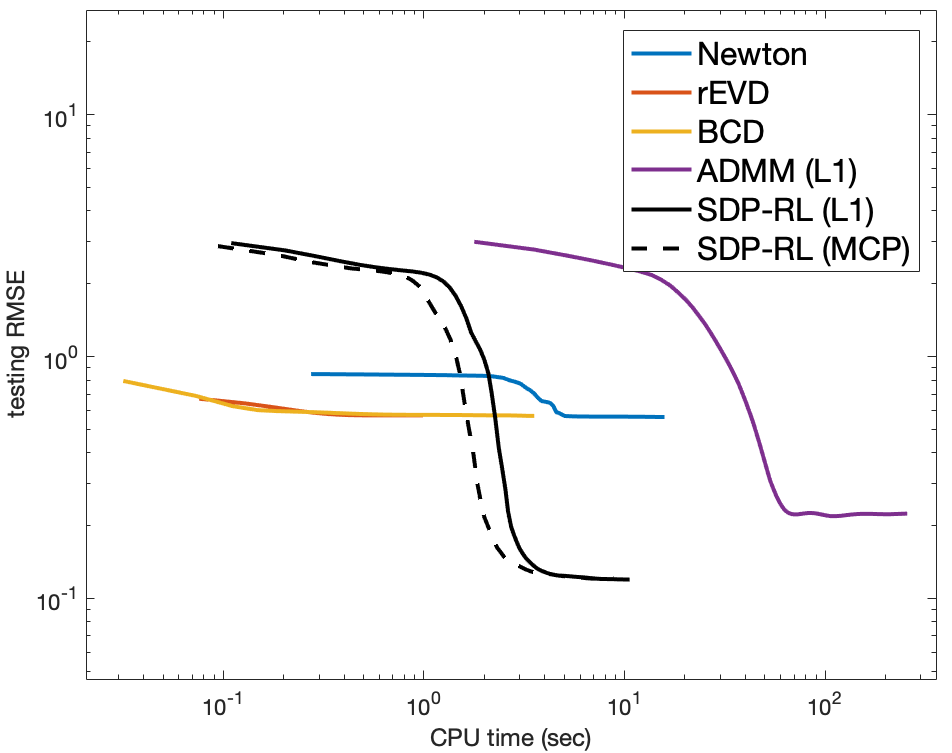}}
	\quad
	\subfigure[synthetic ($m=2000$).]
	{\includegraphics[width=0.245\textwidth]{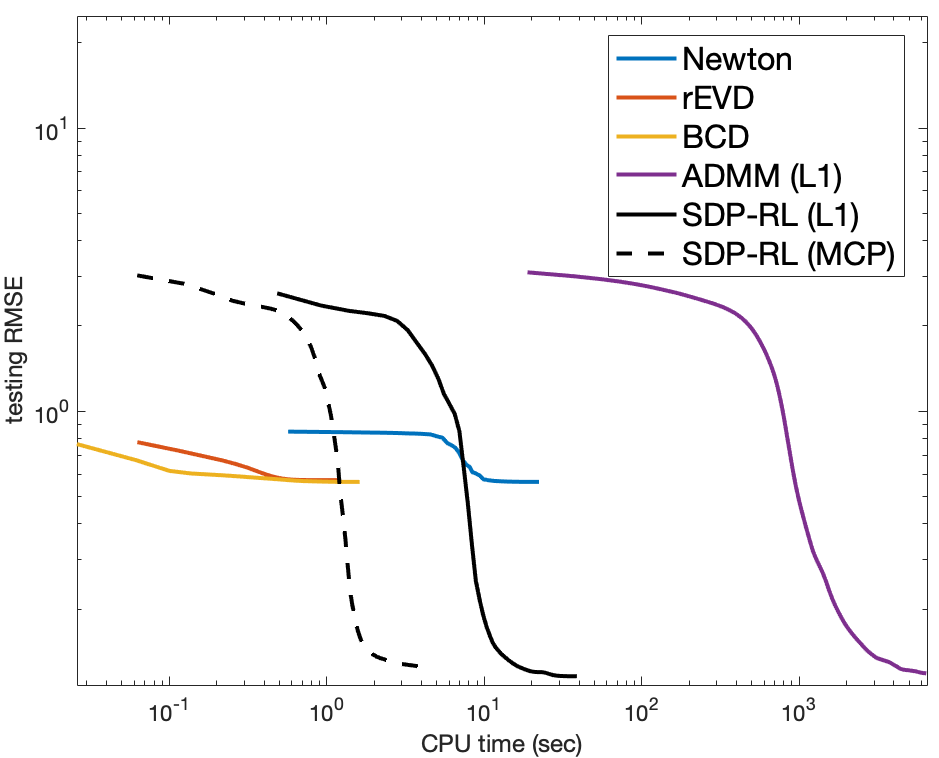}}
	\quad
	\subfigure[USPS.]
	{\includegraphics[width=0.245\textwidth]{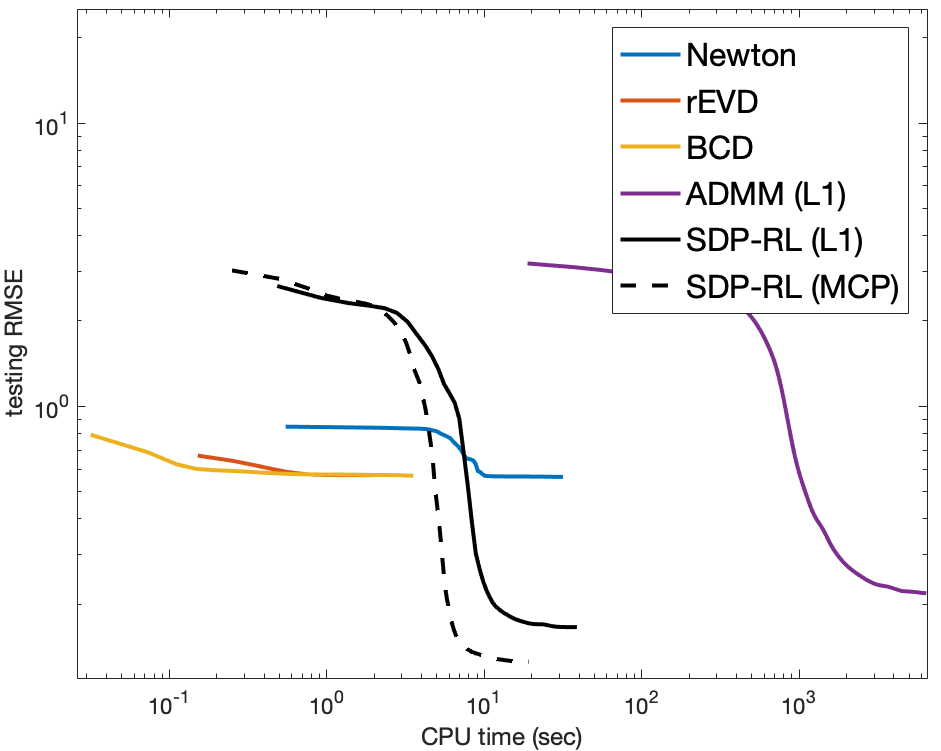}}

	\vspace{-8px}
	\caption{Convergence of testing RMSE vs CPU time (sec) 
		in the SNMF experiment.}
	\label{fig:snmf:obj}
\end{figure*}

Results are shown in Table~\ref{tab:nmf},
and the convergence of testing RMSE w.r.t. CPU time 
is shown in Figure~\ref{fig:snmf:obj}. 
Again, they demonstrate that SDP-RL (using either the $\ell_1$ or MCP loss) 
is significantly more robust (lower testing RMSE)
on noisy data as compared to methods based on the square loss.  Moreover,
\textit{SDP-RL($\ell_1$)} is more efficient than \textit{ADMM($\ell_1$)}.

\begin{table*}[ht]
	\centering
	\caption{Testing RMSEs and CPU time (sec) in the SNMF experiment.}
	\vspace{-8px}
	\begin{tabular}{c | c | c  c | c  c | c  c }
		\hline
\multirow{2}{*}{loss} & \multirow{2}{*}{algorithm} &
\multicolumn{2}{c|}{synthetic ($m=1000$)}   &   \multicolumn{2}{c|}{synthetic ($m=2000$)}   &   \multicolumn{2}{c}{USPS}    \\
		                      &                            & testing RMSE             & CPU time    & testing  RMSE            & CPU time    & testing RMSE    & CPU time    \\ \hline
		       square         & Newton                     & 0.783$\pm$0.007          & 9$\pm$1     & 0.564$\pm$0.003          & 22$\pm$6    & 0.821$\pm$0.006 & 25$\pm$7    \\
		                      & rEVD                       & 0.799$\pm$0.005          & 0.5$\pm$0.1 & 0.571$\pm$0.002          & 1.2$\pm$0.2 & 0.832$\pm$0.007 & 2.0$\pm$0.4 \\
		                      & BCD                        & 0.781$\pm$0.008          & 1.6$\pm$0.5 & 0.565$\pm$0.003          & 2.9$\pm$0.8 & 0.823$\pm$0.009 & 3.2$\pm$0.6 \\ \hline
		      $\ell_1$        & ADMM($\ell_1$)                      & 0.433$\pm$0.006          & 515$\pm$27  & 0.330$\pm$0.005          & 2763$\pm$47 & 0.486$\pm$0.007 & 3018$\pm$65 \\
		                      & SDP-RL($\ell_1$)             & 0.212$\pm$0.004          & 12$\pm$1    & 0.158$\pm$0.002          & 35$\pm$2    & 0.267$\pm$0.009 & 39$\pm$4    \\ \hline
		      leaky-MCP       & $\!\!\!$SDP-RL(MCP)$\!\!\!$  &
				\textbf{0.119$\pm$0.002} & 15$\pm$3    & \textbf{0.112$\pm$0.001} &
				48$\pm$4    & \textbf{0.194$\pm$0.006} & 57$\pm$8    \\ \hline
	\end{tabular}
	\label{tab:nmf}
\end{table*}

\section{Conclusion}

In this paper,
we propose a robust and factorized formulation of SDP
by replacing the commonly used square loss with more robust losses
($\ell_1$-loss and non-convex losses).
As the resulting optimization problem 
is neither convex nor smooth,
existing SDP solvers cannot be applied.
We propose a new solver based on majorization-minimization.
By allowing inexactness in the underlying ADMM  subproblem, 
the algorithm is much more efficient 
while 
still guaranteed
to converge to a critical point.
Experiments are performed on
five applications: matrix completion, kernel learning, matrix variance unfolding,
sparse PCA, and symmetric non-negative matrix factorization.
Empirical results demonstrate the efficiency and robustness
over state-of-the-arts SDP solvers.

\section{Acknowledgment}
This research was supported in part by  the National Natural Science Foundation of China (No.61663049).


\ifCLASSOPTIONcaptionsoff
  \newpage
\fi



%

\bibliographystyle{IEEEtran}
\bibliography{IEEEabrv}
%
%

%


\begin{IEEEbiography}[{\includegraphics[width = 1\textwidth]{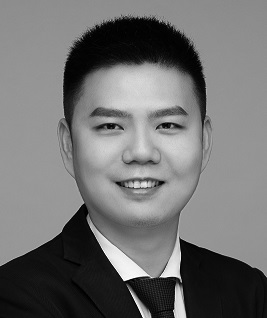}}]{Quanming Yao}
(member, IEEE)
is currently a senior scientist in 4Paradigm (Hong Kong)
and an incoming assistant professor (tenure-track) of 
Department of Electrical Engineering
Tsinghua University. 
His research interests are in
machine learning,
nonconvex optimization,
and automated machine learning.
He obtained his Ph.D. degree in the Department of Computer Science and Engineering at Hong Kong University of Science
and Technology (HKUST) in 2018. 
He has received 
Wunwen Jun Prize for Excellence Youth of Artificial Intelligence (issued by CAAI, 2019),
the 1st runner up of Ph.D. Research Excellence Award (School of Engineering, HKUST, 2018-2019) 
and Google Fellowship (machine learning, 2016).
\end{IEEEbiography}


\begin{IEEEbiography}[{\includegraphics[width = 1\textwidth]{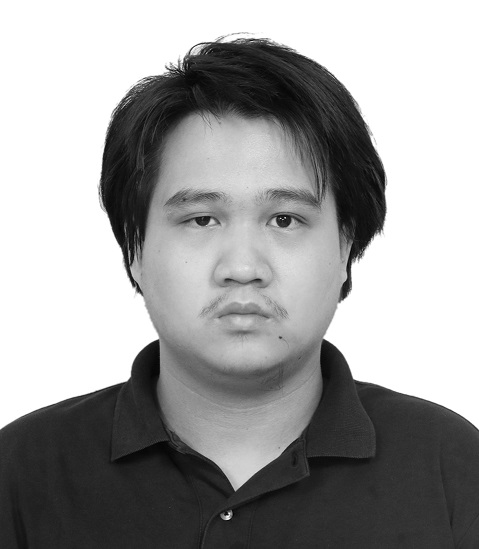}}]{Hansi Yang}
joins Tsinghua University in 2017,
and is an undergraduate student with
Department of Electronic Engineering.
He is currently an intern in machine learning research group of 4Paradigm Inc supervised by Dr. Yao.
His research interests are in machine learning and automated machine learning.
\end{IEEEbiography}


\begin{IEEEbiography}[{\includegraphics[width = 1\textwidth]{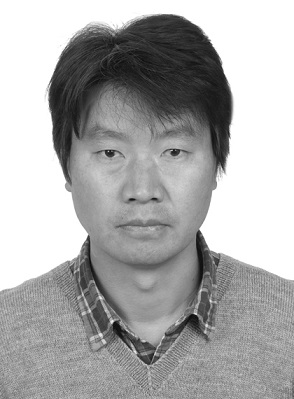}}]{En-Liang Hu} (member, IEEE)
En-Liang Hu received his Ph.D. degree in computer
science from the Nanjing University of Aeronautics
and Astronautics, Nanjing, China, in 2010.
He is currently a Professor with the
Department of Mathematics, Yunnan Normal University, Cheng Gong Campus, Kunming. His current research interests include machine learning, data mining, and optimization.
\end{IEEEbiography}


\begin{IEEEbiography}[{\includegraphics[width = 1\textwidth]{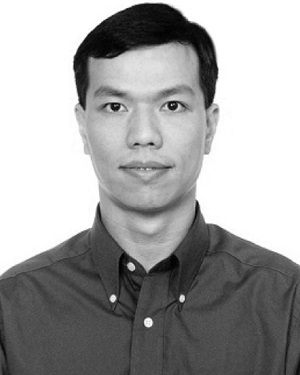}}]{James T. Kwok} (Fellow, IEEE)
	received the Ph.D. degree in computer science from The Hong Kong University of Science and Technology in 1996. 
	He is a Professor with the Department of Computer Science and Engineering, Hong Kong
	University of Science and Technology. His research interests include machine
	learning, deep learning, and artificial intelligence. He received the
	IEEE Outstanding 2004 Paper Award and the Second Class Award in Natural Sciences by
	the Ministry of Education, China, in 2008. 
	He is serving as an Associate Editor for the IEEE Transactions on Neural Networks and Learning Systems, Neural Networks, Neurocomputing, Artificial Intelligence Journal, International Journal of Data Science and Analytics, Editorial Board Member of Machine Learning,
	Board Member, and Vice President for Publications of the Asia Pacific Neural Network
	Society. He also served/is serving as Senior Area Chairs / Area Chairs of major machine learning /
	AI conferences including NIPS, ICML, ICLR, IJCAI, AAAI and ECML.
\end{IEEEbiography}


\cleardoublepage
\appendices

\section{Proof}

\subsection{Lemma~\ref{pr:upper1}}

\begin{proof}
First, we have 
\begin{align}
& \abs{\tr\( (X_k+\dX)\t Q_{\tau}(X_k + \dX)\) - t_{\tau}} 
\notag
\\
& 
= \abs{\tr(2\dX\t Q_{\tau}X_k + X_k^{\top} Q_{\tau}X_k) - t_{\tau} + \tr(\dX\t Q_\tau \dX)},
\notag
\\
& \le \abs{\tr(2\dX\t Q_{\tau}X_k  +  X_k^{\top} Q_{\tau}X_k)  -  t_{\tau}} 
+ \abs{\tr(\dX\t Q_\tau \dX)}.
\!\!\!
\label{eq:app1}
\end{align}
Then,
let $C = A+\frac{\l}{\g}I$, we have, 
\begin{align}
& \frac{\g}{2} \tr\( (X_k  +  \dX)\t A (X_k  +  \dX)\)  +  \frac{\l}{2} \tr\( (X_k  +  \dX)\t (X_k  +  \dX)\) 
\notag
\\ 
& = \frac{\gamma}{2} \tr\(\dX^\t C\dX + (X_k + 2\dX)^{\top} C X_k\).
\label{eq:app2}
\end{align}
Recall that
\begin{align*}
R(\dX & +X_k)  
= \sum\nolimits_{\tau=1}^m \abs{\tr( (\dX+X_k)^{\top} Q_{\tau} (\dX+X_k )) - t_{\tau}}
\\
+ & \frac{\gamma}{2} \tr( (\dX+X_k)^{\top} A (\dX+X_k) )
+ \frac{\lambda}{2}\| \dX+X_k \|_F^2.
\end{align*}
Combining \eqref{eq:app1}, \eqref{eq:app2}
and the definition of $R(\dX +X_k)$ above,
thus for any $\tilde{X} \in \R^{n \times r}$ we have 
\begin{align*}
& 
R(\dX+X_k) 
\le \sum\nolimits_{\tau=1}^m
\abs{\tr(2\dX\t Q_{\tau}X_k + X_k^{\top} Q_{\tau}X_k) - t_{\tau}} 
\\
&  + 
\sum\nolimits_{\tau=1}^m \abs{\tr(\dX\t Q_\tau \dX)} 
 +  \frac{\gamma}{2} \tr\(\dX^\t C\dX  +  (X_k + 2\dX)^{\top} C X_k\).
\end{align*}
Thus,
we obtain the Lemma.
\end{proof}

\subsection{Lemma~\ref{lemma:sur_abstr}}

\begin{proof}
Since we have $\tr(\dX\t Q_\tau\dX)=
\tr(\dX\t \frac{1}{2} (Q_\tau+Q_\tau^{\top}) \dX)$ 
and $\frac{1}{2} (Q_\tau+Q_\tau^{\top})$ is always symmetric for any $Q_{\tau} \in \mathbb{S}^+$, 
we only need to prove that: let $\bar{S} = S_+ - S_{-}$ 
where $S \in \mathbb{S}^{n}$ is any symmetric matrix,
then
$| \tr(\dX\t S \dX) | \le \tr( \dX\t \bar{S} \dX )$
holds for every $\tilde{X} \in \R^{n \times r}$. 
Let $\lambda_{\max} = \sum\nolimits_{i=1}^n \max(\l_i,0) v_i v_i^{\top}$
and $\lambda_{\min} = \sum\nolimits_{i=1}^n \min(\l_i,0) v_i v_i^{\top}$.
Thus, we have 
\begin{align*}
| \tr(\dX\t S \dX) | 
& = | \tr(\dX\t (\sum\nolimits_{i=1}^n \l_i v_i v_i^{\top}) \dX ) |,
\\ 
& \le | \tr(\dX\t \lambda_{\max} \dX ) | 
+ | \tr(\dX\t \lambda_{\min} \dX ) |, 
\\
& = \tr(\dX\t \lambda_{\max} \dX ) 
- \tr(\dX\t \lambda_{\min} \dX ), 
\\
& = \tr(\dX\t S_+ \dX ) - \tr(\dX\t S_{-} \dX ) =  \tr( \dX\t \bar{S} \dX ).
\end{align*}
Thus,
we obtain the Lemma.
\end{proof}

\subsection{Proposition~\ref{pr:cvxsurr}}

\begin{proof}
Combining Lemma~\ref{pr:upper1} and \ref{lemma:sur_abstr} we will have, 
\begin{align*}
& 
R(\dX+X_k) 
\le \sum\nolimits_{\tau=1}^m
\abs{\tr(2\dX\t Q_{\tau}X_k + X_k^{\top} Q_{\tau}X_k) - t_{\tau}} 
\\
&  + 
\sum\nolimits_{\tau=1}^m \abs{\tr(\dX\t Q_\tau \dX)} 
 +  \frac{\gamma}{2} \tr\(\dX^\t 
 C\dX  +  (X_k  +  2\dX)^{\top} C X_k\), 
\\ 
& \le 2 \sum\nolimits_{\tau=1}^m
\abs{\tr(\dX\t Q_{\tau}X_k) + \frac{1}{2} ( \tr(X_k^{\top} Q_{\tau}X_k) - t_{\tau})} 
\\
&  + 
\sum\nolimits_{\tau=1}^m \tr(\dX\t \bar{Q}_\tau \dX)
 +  \frac{\gamma}{2} \tr\(\dX^\t C\dX  +  (X_k  +  2\dX)^{\top} C X_k\),
\end{align*}
Then, let
$B = Q+\frac{1}{2}(\l I +\g A_+)$, $Q=\sum\nolimits_{\tau=1}^m \bar{Q}_\tau$,
$(b_k)_\tau = \frac{1}{2}(\tr( X_k \t Q_{\tau} X_k ) - t_{\tau})$, 
and $c_k=\frac{\gamma}{2}\tr(X_k\t (A+\frac{\l}{\g}I)X_k)$,
\begin{align*}
R(\dX+X_k) 
\le & 2 \sum\nolimits_{\tau=1}^m
\abs{\tr(\dX\t Q_{\tau}X_k) + (b_k)_\tau} 
\\
&  +  \tr( \dX\t ( B \dX + \g C X_k ) ) + c_k,
\end{align*}
then $R(\dX + X_k)\le H_k( \dX; X_k )$ where
\begin{align}
H_k(\dX; X_k) 
=& \tr( \dX\t ( B \dX + \g C X_k ) )
\notag
\\
&+ 2\sum\nolimits_{\tau=1}^m \abs{\tr( \dX \t Q_{\tau} X_k ) + (b_k)_\tau} + c_k,
\notag
\end{align}
and the equality holds iff $\tilde{X} = \mathbf{0}$.
\end{proof}

\subsection{Proposition~\ref{pr:dual}}

\begin{proof}
Following its definition, denote the dual problem to be
$\max_{ \{ \tilde{\nu}_\tau \} } \mathcal{D}_k(\{ \tilde{\nu}_{\tau} \})$, 
which is defined as: 
\begin{align*}
\mathcal{D}_k(\{ \tilde{\nu}_{\tau} \}) 
= & \min_{\dX, e_\tau} \tr( \dX\t ( B \dX + \g C X_k ) ) 
+ 2\sum\nolimits_{\tau=1}^m \abs{e_\tau} + c_k,
\\
& + \sum\nolimits_{\tau=1}^m \tilde{\nu}_\tau( \tr( \dX \t Q_{\tau} X_k ) + (b_k)_\tau - e_\tau ).
\end{align*}
The problem above can be solved by minimizing w.r.t.
$\dX$ and $e_\tau$ separately. 
We first consider minimizing w.r.t. $e_\tau$, which is given by 
$\min_{e_\tau} 2\sum\nolimits_{\tau=1}^m \abs{e_\tau} - \sum\nolimits_{\tau=1}^m \tilde{\nu}_\tau e_\tau$
and it follows easily that when $|\tilde{\nu}_{\tau}| > 2$, 
the problem above achieves $-\inf$. 
So it requires $|\tilde{\nu}_{\tau}| \le 2$ and the minimized point is 0. 
We then turned on to consider minimizing w.r.t. $\dX$, which is 
(we omit the $c_k$ term for simplicity, as it has no influence to the optimization problem)
\begin{align}
& \min_{\dX}  \tr( \dX\t  
 ( B \dX 
\! + \! \g C X_k ) ) 
\! + \! \sum_{\tau=1}^m \tilde{\nu}_\tau( \tr( \dX \t Q_{\tau} X_k ) 
\! + \! (b_k)_\tau),
\notag
\\
&  
\!\!\! = \! \min_{\dX} \tr( \dX\t  ( B \dX 
\! + \! (\g C 
\! + \!\! \sum_{\tau=1}^m \tilde{\nu}_\tau Q_{\tau})X_k ) ) 
\! + \!\! \sum_{\tau=1}^m \tilde{\nu}_\tau (b_k)_\tau.
\!\!\!
\label{eq:app12}
\end{align}
Let $D=\g C + \sum\nolimits_{\tau=1}^m \tilde{\nu}_\tau Q_{\tau}$, 
\eqref{eq:app12} is equivalent to
\begin{align*}
\min_{\dX} \sum\nolimits_{\tau=1}^m \tilde{\nu}_\tau (b_k)_\tau-\frac{1}{4}\tr(D^{\top} B^{-1} D).
\end{align*}
Thus,
the optimal of \eqref{eq:app12} is  
\begin{align*}
\dX^* = - \frac{1}{2}B^{-1} D.
\end{align*}
Then, the dual problem is
$\max_{ |\tilde{\nu}_{\tau}| \le 2 }
\mathcal{D}_k(\{ \tilde{\nu}_{\tau} \})$ where $\mathcal{D}_k$
is defined in Proposition~\ref{pr:dual}.
\end{proof}

\subsection{Lemma~\ref{lem:dual}}
\label{app:str:dual}

\begin{proof}
Specifically, since our problem in \eqref{eq:surradmm} is convex 
(convex objective and linear constraints), 
we only need to check the Slater's condition, 
i.e. there exists a strictly feasible point for this problem, 
in order to prove its strong duality. 
And the proof is trivial as we have 
$\tilde{X} \in \mathbb{R}^{n \times r}, \{ e_{\tau} \in \mathbb{R} \}$, 
therefore the constraints $e_\tau =\tr( \tilde{X}^{\top} Q_{\tau} X_{k} ) + (b_k)_\tau$ 
can always be satisfied by choosing an appropriate $e_{\tau}$. 
\end{proof}

\subsection{Theorem~\ref{thm:conv1}}
\label{app:conv1}

We first introduce the following Lemma~\ref{lem:app2} and \ref{lem:app1}.
When a function $f$ has multiple input parameters,
$\partial_k f$ means taking subdifferential to its $k$th parameter.

\begin{lemma}
\label{lem:app2}
There exists a positive constant $\alpha>0$,
such that
\begin{enumerate}
\item $H_k(\dX_1; X_k) \ge H_k(\dX_2; X_k) + \frac{\alpha}{2} \NM{\dX_1 - \dX_2}{F}^2$
	holds for any $\dX_1, \dX_2$; and

\item $R(X_{k}) - R(X_{k + 1}) \ge \frac{\alpha}{2}\| X_{k + 1} - X_k \|^2 - \epsilon_k$.
\end{enumerate}
\end{lemma}

\begin{proof}
\textbf{Part 1).}
Recall from \eqref{eq:surrh} in Proposition~\ref{pr:cvxsurr} that $H_k$ is defined as
\begin{align}
H_k(\dX; X_k) 
& \equiv \tr( \dX\t ( B \dX + \g C X_k ) )
\notag
\\
& + 2\sum\nolimits_{\tau=1}^m \abs{\tr( \dX \t Q_{\tau} X_k ) + (b_k)_\tau} + c_k.
\label{eq:app13}
\end{align}
Thus,
to show Part 1) holds,
we only need to show  that
the smallest eigenvalue of $B$ in the quadratic term,
i.e., the first term in $H_k$,
is positive.
This can be easily seen from the definition of $B$,
i.e.,
$B = \sum\nolimits_{\tau=1}^m \bar{Q}_\tau + \frac{1}{2}(\l I +\g A_+)$,
since $A_+$, $\bar{Q}_\tau$ are PSD 
from its definition and Lemma~\ref{lemma:sur_abstr}, 
$I$ is the identity matrix,
and $\lambda$ is required to be positive.

\vspace{5px}

\noindent\textbf{Part 2).}
From Proposition~\ref{pr:cvxsurr},
we know 
\begin{align}
R(X_k)
& =H_k( \bm{0}, X_k ),
\label{eq:app14}
\\
R(X_k + \dX^*) 
& \le H_k(\dX^*, X_k ).
\label{eq:app15}
\end{align}
Recall that
$\dX^*$ is approximated by $\dX_t$
in step~3 of Algorithm~\ref{alg:rsdpmm}.
Using \eqref{eq:app14} and \eqref{eq:app15},
we have
\begin{align}
& R(X_{k}) 
-  R(X_k + \dX^*) 
\ge H_k( \bm{0}, X_k ) - H_k(\dX^*, X_k ),
\notag
\\ 
& =  
\big[ H_k( \bm{0}, X_k ) - H_k( \dX_t, X_k ) \big]  
\notag
\\
& \qquad\qquad +  
\big[
H_k( \dX_t, X_k )
-
H_k(\dX^*, X_k )  
\big].
\label{eq:app3}
\end{align}
Using $\delta_k$'s definition in (\ref{eq:gap}),
we have
\begin{align*}
\epsilon_k 
\ge 
\delta_k(\dX_t, \{ (\tilde{\nu}_{\tau})_t \}) 
& = H_k(\dX_{t}; X_k) - \mathcal{D}_k(\{ \nu_{\tau} \}_{t}),
\\
& \ge H_k(\dX_t; X_k) - H_k(\dX^*; X_k).
\end{align*}
Thus,
\begin{align}
- \epsilon_k 
\le 
H_k( \dX_t, X_k ) 
-
H_k(\dX^*, X_k )
\le 0.
\label{eq:app17}
\end{align} 
From part 1),
we also have
\begin{align}
H_k( \bm{0}, X_k ) & - H_k( \dX^*, X_k)
\notag
\\
& \ge \frac{\alpha}{2}\| X_{k + 1} - X_k \|^2 
= \frac{\alpha}{2}\| \dX^* \|^2.
\label{eq:app16}
\end{align}
Finally,
combining \eqref{eq:app3}-\eqref{eq:app16},
we then obtain 
\begin{align*}
R(X_{k}) - R(X_{k + 1}) \ge \frac{\alpha}{2}\| X_{k + 1} - X_k \|^2 - \epsilon_k.
\end{align*}
Thus,
part 2) holds.
\end{proof}

\begin{lemma}\label{lem:app1}
\begin{enumerate}
\item $\partial_{2} H_k(\bm{0}, X_k)  = \partial R(X_k)$; and

\item $\bm{0} \in \lim\nolimits_{t \rightarrow \infty} \partial_{2} H_k(\dX_t, X_k)$.
\end{enumerate}
\end{lemma}

\begin{proof}
\textbf{Part 1).}
Recall the definition of $R(X)$ in \eqref{eq:rsdp}, i.e., 
\begin{align*}
	R(\dX & +X_k)  
	= \sum\nolimits_{\tau=1}^m \abs{\tr( (\dX+X_k)^{\top} Q_{\tau} (\dX+X_k )) - t_{\tau}}
	\\
	+ & \frac{\gamma}{2} \tr( (\dX+X_k)^{\top} A (\dX+X_k) )
	+ \frac{\lambda}{2}\| \dX+X_k \|_F^2.
\end{align*}
Thus,
\begin{align} 
& \partial R(X_k)   
= \g AX_k + \l X_k +
\notag
\\
& + \sum\nolimits_{\tau=1}^m \text{sign}(\tr(X_k^{\top} Q_{\tau} X_k) -t_{\tau})
\frac{1}{2}(Q_{\tau}+Q^{\top}_{\tau})
X_k.
\label{eq:app10}
\end{align} 
Then,
recall the definition of $H_k$ in \eqref{eq:app13},
we have
\begin{align} 
\partial_{2} H_k( \bm{0}, X_k) 
& = \g CX_k 
\notag
\\
+ & 2 \sum\nolimits_{\tau=1}^m \text{sign}
\big( (b_k)_\tau \big)
\frac{1}{2}(Q_{\tau}+Q^{\top}_{\tau})
X_k.
\label{eq:app11}
\end{align} 
Since $C=A + \frac{\l}{\g} I$ and $(b_k)_\tau = \frac{1}{2}(\tr( X_k \t Q_{\tau} X_k ) - t_{\tau})$
(defined in Proposition~\ref{pr:cvxsurr}), 
we can deduce that the Lemma holds
by comparing \eqref{eq:app10} and \eqref{eq:app11}.

\vspace{5px}
\noindent
\textbf{Part 2).}
Recall that $\dX_t = X_{k+1} - X_k$ is defined 
at step~3 of Algorithm~\ref{alg:rsdpmm}.
Using $\delta_k$'s definition in (\ref{eq:gap}),
we have
\begin{align*}
\delta_k(\dX_t, \{ (\tilde{\nu}_{\tau})_t \}) 
& = H_k(\dX_{t}; X_k) - \mathcal{D}_k(\{ \nu_{\tau} \}_{t}),
\\
& \ge H_k(\dX_t; X_k) - H_k(\dX^*; X_k) \ge 0.
\end{align*}
Since
$H_k$ is a continuous function
and
\begin{align*}
\lim\nolimits_{t \rightarrow \infty} \delta_k(\dX_t, \{ (\tilde{\nu}_{\tau})_t \}) = 0,
\end{align*}
we have
\begin{align*}
\lim\nolimits_{t \rightarrow \infty}
H_k(\dX_t; X_k) - H_k(\dX^*; X_k) = 0,
\end{align*}
which means $\bm{0} \in \lim\nolimits_{t \rightarrow \infty} \partial_{2} H_k(\dX_t, X_k)$.
\end{proof}

\begin{proof}
(of Theorem~\ref{thm:conv1})
\textbf{Conclusion (i).}
From part 2) in Lemma~\ref{lem:app2}, 
we have
\begin{align*}
\frac{\alpha}{2}\| X_{k + 1} - X_k \|^2 \le R(X_{k}) - R(X_{k + 1}) + \epsilon_k.
\end{align*}
Thus,
\begin{align}
\sum\nolimits_{k = 1}^K & \frac{\alpha}{2}  \| X_{k + 1}  -  X_k \|_F^2
\notag
\\
& \le 
\sum\nolimits_{k = 1}^K R(X_{k}) - R(X_{k + 1}) + \epsilon_k
\notag
\\
& \le R(X_1) - R(X_{K + 1}) + \sum\nolimits_{k = 1}^K \epsilon_k
\notag
\\
& \le R(X_1) - \inf R + \sum\nolimits_{k = 1}^{\infty} \epsilon_k.
\label{eq:app9}
\end{align}
From Assumption~\ref{ass:err},
we know that the last term in \eqref{eq:app9},
i.e., $\sum\nolimits_{k = 1}^{\infty} \epsilon_{k}$, 
is finite. 
Together with Assumption~\ref{ass:obj},
we have 
\begin{align}
\lim\nolimits_{k\rightarrow \infty} 
&
\| X_{k + 1} - X_k \|_F^2 
= 
0,
\label{eq:app51}
\\
\text{\; and \;}
&
0  
\le 
\lim\nolimits_{k\rightarrow \infty} \| X_k \|_F^2
<
\infty,
\notag
\end{align}
which means that the sequence $\{ X_k \}$ is bounded
and has at least one limit points.

\vspace{5px}
\noindent
\textbf{Conclusion (ii).}
From Part 1) in Lemma~\ref{lem:app1},
we have
$\partial_{2} H_k(\bm{0}, X_k)  = \partial R(X_k)$. 
Then, denote the limit point of sequence $\{ X_k \}$ as $X_*$, 
and let $\{ X_{k_j} \}$ be a sub-sequence of $\{ X_k \}$
such that
\begin{align}
X_* = \lim\nolimits_{k_j \rightarrow \infty} X_{k_j}.
\label{eq:app61}
\end{align}
Thus, to prove the limit point $X_*$ is also a critical point for $R(X)$, 
we only need to show 
\begin{align}
\bm{0} \in \lim\nolimits_{k_j \rightarrow \infty} \partial_{2} H_{k_j}(\bm{0}, X_*).
\label{eq:app81}
\end{align}
Using Part 2) in Lemma~\ref{lem:app1},
we should have 
\begin{align}
\bm{0} 
& \in \lim\nolimits_{t \rightarrow \infty} \partial_{2} H_{k_j}(\dX^{k_j}_{t}, X_{k_j}),
\notag
\\
& = \partial_{2} H_{k_j}( \lim\nolimits_{t \rightarrow \infty} \dX^{k_j}_{t}, X_{k_j}).
\label{eq:app18}
\end{align}
Thus, denote $\lim\nolimits_{t \rightarrow \infty} \dX^{k_j}_{t} = \dX^{k_j}_*$, 
we only need to prove 
\begin{align}
\lim\nolimits_{k_j \to \infty} \dX^{k_j}_* = \bm{0}.
\label{eq:app19}
\end{align}
Since $\sum\nolimits_{k = 1}^{\infty} \epsilon_{k}$ is finite, we must have $\lim\nolimits_{k_j \to \infty} \epsilon_{k_j} = 0$, 
which implies that 
\begin{align}
\lim\nolimits_{k_j \rightarrow \infty}(X_{k_j + 1} - X_{k_j} - \dX^{k_j}_* )= \mathbf{0}.
\label{eq:app36}
\end{align} 
Then from \eqref{eq:app51}, we have 
\begin{align}
\lim\nolimits_{k_j \rightarrow \infty} X_{k_j + 1} - X_{k_j} = \bm{0}, 
\label{eq:app31}
\end{align}
Then, 
\eqref{eq:app19} follows easily from \eqref{eq:app36} and \eqref{eq:app31}. 
Finally,
\eqref{eq:app81} can be obtained
by combining \eqref{eq:app18} and \eqref{eq:app19}.
Thus,
any limit point of $\{ X_k \}$ is a critical point of $R$.
\end{proof}

\subsection{Lemma~\ref{pr:upper2}}

\begin{proof}
Since $\phi(\cdot)$ is concave on $(0,\infty)$,  
we have $\phi(y) \le \phi(x) + \phi '(x) (y-x)$  for any $x,y \in (0,\infty)$. 
That means for any $\alpha,\beta \in \mathbb{R}$, 
$\phi(\abs{\beta}) \le \phi(\abs{\alpha}) + \phi '(\abs{\alpha}) (\abs{\beta}-\abs{\alpha})$. 
Then, we consider our objective:
\begin{align*}
R(X_k +\widetilde{X}) 
& 
\! = \! 
\sum\nolimits_{\tau=1}^{m} \phi(\vert \tr( (X_k+\widetilde{X})^{\top} Q_{\tau} (X_k+\widetilde{X}) ) - t_{\tau}\vert ), 
\\
&
\! + \! 
\frac{\gamma}{2} \tr( (X_k+\widetilde{X})^{\top} A (X_k+\widetilde{X}) ) + \frac{\l}{2}\NM{X_k+\widetilde{X}}{F}^2, 
\\
& \! = \! 
\sum\nolimits_{\tau=1}^{m} \phi(\vert \tr( (X_k+\widetilde{X})^{\top} Q_{\tau} (X_k+\widetilde{X}) ) - t_{\tau}\vert ), 
\\
&
\! + \! 
\frac{\gamma}{2} \tr( X_k^{\top} C X_k ) + \frac{\gamma}{2} \tr( (2 X_k+\tilde{X})^{\top} C \tilde{X} ),
\end{align*}
where $C=A+\frac{\lambda}{\gamma} I$. 
Denote $(q_k)_{\tau} = \phi' \(\vert \tr( X_k^{\top} Q_{\tau} X_k ) - t_{\tau} \vert\) $, 
we will have
\begin{align*}
& \phi(\vert \tr( (X_k+\widetilde{X})^{\top} Q_{\tau} (X_k+\widetilde{X}) ) - t_{\tau}\vert ) \\
& \le (q_k)_{\tau} \vert \tr( (X_k+\widetilde{X})^{\top} Q_{\tau} (X_k+\widetilde{X}) ) - t_{\tau}  \vert \\
& + \(\phi (\vert \tr( X_k^{\top} Q_{\tau} X_k )- t_{\tau} \vert)-(q_k)_{\tau} \vert \tr( X_k^{\top} Q_{\tau} X_k ) - t_{\tau} \vert \)
\end{align*}
And it follows easily that:
\begin{align*}
& \vert \tr( (X_k+\widetilde{X})^{\top} Q_{\tau} (X_k+\widetilde{X}) ) - t_{\tau}  \vert \\
& \le \vert \tr( (2 \tilde{X} + X_k)^{\top} Q_{\tau} X_k )  -  t_{\tau} \vert + \vert \tr( \tilde{X}^{\top} Q_{\tau} \tilde{X} ) \vert
\end{align*}
Denote 
\begin{align*}
\dot{c}_k & = \sum\nolimits_{\tau=1}^{m} \(\phi (\vert \tr( X_k^{\top} Q_{\tau} X_k ) 
\\ 
& - t_{\tau} \vert)- (q_k)_{\tau} \vert \tr( X_k^{\top} Q_{\tau} X_k ) - t_{\tau} \vert \) + \frac{\gamma}{2} \tr( X_k^{\top} C X_k ),
\\
C & = A+\frac{\l}{\g}I,
\\
(q_k)_{\tau} 
& = \phi' \(\vert \tr( X_k^{\top} Q_{\tau} X_k ) - t_{\tau} \vert\).
\end{align*}
Thus,
we have 
\begin{align*}
\dot{R}(X_k + \tilde{X}) 
\le &
\frac{\gamma}{2} \tr( (2 X_k+\tilde{X})^{\top} C \tilde{X} ) + \dot{c}_k
\notag
\\ 
+ & 
\sum\nolimits_{\tau=1}^{m} (q_k)_{\tau}
|  \tr( \tilde{X}^{\top} Q_{\tau} \tilde{X} ) | 
\notag
\\
+ &  
\sum\nolimits_{\tau=1}^{m} 
(q_k)_{\tau}
|  \tr( (2 \tilde{X} + X_k)^{\top} Q_{\tau} X_k )  -  t_{\tau} |.
\end{align*} 
for any $\tilde{X} \in \R^{n \times r}$. 
\end{proof}

\subsection{Proposition~\ref{pr:cvxsurr2}}

\begin{proof}
Since $\phi(\cdot)$ is an increasing function on $(0,\infty)$, 
we have $(q_k)_{\tau}>0$.
Let $\dot{Q}_{\tau}= (q_k)_{\tau} Q_{\tau}$ and $(\dot{b}_{k})_{\tau} = \frac{1}{2} (q_k)_{\tau} \( \tr(  X_k^{\top} Q_{\tau} X_k )-t_\tau \)$, we will have:
\begin{align*}
& \dot{R}(X_k+
\tilde{X}) 
\le
\frac{\gamma}{2} \tr( (2 X_k+\tilde{X})^{\top} C \tilde{X} ) + \dot{c}_k
\notag
\\ 
& + 
\sum\nolimits_{\tau=1}^{m}
|  \tr( \tilde{X}^{\top} \dot{Q}_{\tau} \tilde{X} ) | 
+ 
2 \sum\nolimits_{\tau=1}^{m} 
|  \tr( \tilde{X}^{\top} \dot{Q}_{\tau} X_k)  +  (\dot{b}_{k})_{\tau} |.
\end{align*}
Then, similar with Proposition~\ref{pr:cvxsurr}, denote $\tilde{Q}_{\tau} = (q_k)_{\tau} \bar{Q}_{\tau}$. Following Lemma~\ref{lemma:sur_abstr}, we will have $| \tr( \tilde{X}^{\top} \dot{Q}_{\tau} \tilde{X} ) | \le \tr( \tilde{X}^{\top} \tilde{Q}_{\tau} \tilde{X} )$. 
Let $Q=\sum\nolimits_{\tau=1}^m \tilde{Q}_\tau+\frac{1}{2}(\l I +\g A_+)$ and we will have: 
\begin{align*}
\dot{R}(X_k + \tilde{X}) 
\le & \tr( \tilde{X}^{\top} (Q \tilde{X} + \gamma C X_k)) 
\\
& + 2\sum\nolimits_{\tau=1}^{m}\vert \tr( \tilde{X}^{\top} \dot{Q}_{\tau} X_k ) + (\dot{b}_{k})_{\tau} \vert + \dot{c}_k.
\end{align*}

Thus $\dot{R}(\dX + X_k)\le \dot{H}_k( \dX; X_k )$ where
\begin{align*}
\dot{H}_k( \dX; X_k ) =  & \tr( \tilde{X}^{\top} (Q \tilde{X} + \gamma C X_k)) 
\\
& + 2\sum\nolimits_{\tau=1}^{m}\vert \tr( \tilde{X}^{\top} \dot{Q}_{\tau} X_k ) + (\dot{b}_{k})_{\tau} \vert + \dot{c}_k.
\end{align*} 
and the equality holds iff $\tilde{X} = \mathbf{0}$.
\end{proof}

\subsection{Theorem~\ref{thm:conv2}}
\label{app:conv2}

The proof here is similar to that of Theorem~\ref{thm:conv1}. 
We first introduce the following Lemma~\ref{lem:app3} and \ref{lem:app4}.

\begin{lemma}
\label{lem:app3}
There exists a positive constant $\alpha>0$,
such that
\begin{enumerate}[leftmargin=*]
\item 
$\dot{H}_k( \dX_1, X_k ) - \dot{H}_k(\dX_2, X_k) \ge \frac{\alpha}{2}\| \dX_1-\dX_2 \|^2$ holds for any $\dX_1, \dX_2$; and

\item $\dot{R}(X_{k}) - \dot{R}(X_{k + 1}) \ge \frac{\alpha}{2}\| X_{k + 1} - X_k \|^2 - \epsilon_k$.
\end{enumerate}
\end{lemma}

\begin{proof}
\textbf{Part 1).}
Recall from \eqref{eq:surrh} in Proposition~\ref{pr:cvxsurr2} that $\dot{H}_k$ is defined as
\begin{eqnarray*}
\dot{H}_k( \dX; X_k ) & \equiv & \tr( \tilde{X}^{\top} (B \tilde{X} + \gamma C X_k)) 
\\
& & + 2\sum\nolimits_{\tau=1}^{m}\vert \tr( \tilde{X}^{\top} \dot{Q}_{\tau} X_k ) + (\dot{b}_{k})_{\tau} \vert + \dot{c}_k,
\end{eqnarray*}
Thus,
to show the Lemma holds,
we only need to show  that
the smallest eigenvalue of $B$ in the quadratic term,
i.e., the first term in $\dot{H}_k$,
is positive.
This can be easily seen from the definition of $B$,
i.e.,
$B = \sum\nolimits_{\tau=1}^m \dot{Q}_\tau + \frac{1}{2}(\l I +\g A_+)$,
since $A_+$, $\dot{Q}_\tau$ are PSD 
from its definition and Lemma~\ref{lemma:sur_abstr}, 
$I$ is the identity matrix,
and $\lambda$ is required to be positive.

\vspace{5px}

\noindent\textbf{Part 2).}
From Proposition~\ref{pr:cvxsurr2},
we have
\begin{align}
\dot{R}(X_k)
& =\dot{H}_k( \bm{0}, X_k ),
\label{eq:app142}
\\
\dot{R}(X_k + \dX^*) 
& \le \dot{H}_k(\dX^*, X_k ).
\label{eq:app152}
\end{align}
Recall that
$\dX^*$ is approximated by $\dX_t$ in step~3 of Algorithm~\ref{alg:rsdpmm:2}. 
Using \eqref{eq:app14} and \eqref{eq:app15},
we have
\begin{align}
& \dot{R}(X_{k}) 
-  \dot{R}(X_{k + 1}) 
\ge \dot{H}_k( \bm{0}, X_k ) - \dot{H}_k(\dX^*, X_k ) ,
\notag
\\ 
& =  
\big[ \dot{H}_k( \bm{0}, X_k ) - \dot{H}_k( \dX_t, X_k ) \big]  
\notag
\\
& \qquad +  
\big[
\dot{H}_k( \dX_t, X_k )
-
\dot{H}_k(\dX^*, X_k )  
\big].
\label{eq:app32}
\end{align}
Using $\delta_k$'s definition in (\ref{eq:gap}),
we have
\begin{align*}
\epsilon_k 
\ge 
\delta_k(\dX_t, \{ (\tilde{\nu}_{\tau})_t \}) 
& = H_k(\dX_{t}; X_k) - \mathcal{D}_k(\{ \nu_{\tau} \}_{t}),
\\
& \ge H_k(\dX_t; X_k) - H_k(\dX^*; X_k).
\end{align*}
Thus, 
\begin{align}
- \epsilon_k 
\le 
\dot{H}_k( \dX_t, X_k ) 
-
\dot{H}_k(\dX^*, X_k )
\le 0.
\label{eq:app172}
\end{align} 
From part 1), we also have: 
\begin{align}
\dot{H}_k( \bm{0}, X_k )
& - \dot{H}_k( \dX^*, X_k) \notag \\
& \ge \frac{\alpha}{2}\| X_{k + 1} - X_k \|^2 = \frac{\alpha}{2}\| \dX^* \|^2.
\label{eq:app162}
\end{align}
Finally,
combining \eqref{eq:app32}-\eqref{eq:app162},
we then obtain 
\begin{align*}
\dot{R}(X_{k}) - \dot{R}(X_{k + 1}) \ge \frac{\alpha}{2}\| X_{k + 1} - X_k \|^2 - \epsilon_k.
\end{align*}
Thus,
part 2) holds.
\end{proof}

\begin{lemma}\label{lem:app4}
\begin{enumerate}
\item $\partial_{2} \dot{H}_k(\bm{0}, X_k)  = \partial \dot{R}(X_k)$; and

\item $\bm{0} \in \lim\nolimits_{k \rightarrow \infty} \partial_{2} \dot{H}_k(\dX_k, X_k)$.
\end{enumerate}
\end{lemma}

\begin{proof}
\textbf{Part 1).}
First, 
recall from the definition of $\dot{R}(X)$ in \eqref{eq:rsdpnc} that 
\begin{align*}
\dot{R}(\dX 
 +  X_k)  
 =  & \sum\nolimits_{\tau=1}^{m} \phi(\vert \tr( (X_k+\widetilde{X})^{\top} Q_{\tau} (X_k
 +  \widetilde{X}) ) 
 -  t_{\tau}\vert ) \\
& + \frac{\gamma}{2} \tr( X_k^{\top} C X_k ) + \frac{\gamma}{2} \tr( (2 X_k+\tilde{X})^{\top} C \tilde{X} ),
\end{align*}
and 
\begin{align*}
\dot{H}_k( \dX; X_k ) =  & \tr( \tilde{X}^{\top} (Q \tilde{X} + \gamma C X_k)) 
\\
& + 2\sum\nolimits_{\tau=1}^{m}\vert \tr( \tilde{X}^{\top} \dot{Q}_{\tau} X_k ) 
+ (\dot{b}_{k})_{\tau} \vert + \dot{c}_k.
\end{align*} 
We have
\begin{align} 
& 
\partial \dot{R}(X_k) = \g AX_k + \l X_k +
\notag \\
& \! + \! 
\sum\nolimits_{\tau=1}^m (q_k)_{\tau} 
\text{sign}(\tr(X_k^{\top} Q_{\tau} X_k) \! - \! t_{\tau})
\frac{1}{2}(\dot{Q}_{\tau} \! + \! \dot{Q}^{\top}_{\tau})
X_k
\label{eq:app102}
\end{align} 
and
\begin{align} 
\partial_{2} 
& \dot{H}_k(\bm{0}, X_k) = \g CX_k 
\notag
\\
& + 2\sum\nolimits_{\tau=1}^m \text{sign}((\dot{b}_k)_\tau)
\frac{1}{2}(\dot{Q}_{\tau}+\dot{Q}^{\top}_{\tau})
X_k
\label{eq:app112}
\end{align} 
where $(q_k)_{\tau} = \phi' \(\vert \tr( X_k^{\top} Q_{\tau} X_k ) - t_{\tau} \vert\)$. 
Since 
\begin{align*}
C=A+\frac{\l}{\g} I
\;\text{and}\;
(\dot{b}_{k})_{\tau} = \frac{1}{2} (q_k)_{\tau} 
\big(
\tr(  X_k^{\top} Q_{\tau} X_k )-t_\tau
\big), 
\end{align*}
we can deduce that the Lemma holds from \eqref{eq:app102} and \eqref{eq:app112}.

\vspace{5px}
\noindent
\textbf{Part 2).}
Recall that $\dX_k$ is defined 
at step~3 of Algorithm~\ref{alg:rsdpmm:2} 
and $\delta_k$ is in (\ref{eq:gap}),
we have
\begin{align*}
\delta_k(\dX_t, \{ (\tilde{\nu}_{\tau})_t \}) 
& = \dot{H}_k(\dX_{t}; X_k) - \dot{\mathcal{D}}_k(\{ \nu_{\tau} \}_{t}),
\\
& \ge \dot{H}_k(\dX_t; X_k) - \dot{H}_k(\dX^*; X_k) \ge 0.
\end{align*}
Since
$\dot{H}_k$ is a continuous function
and
$\lim\nolimits_{t \rightarrow \infty} \delta_k(\dX_t, \{ (\tilde{\nu}_{\tau})_t \}) = 0$,
we have
\begin{align*}
\lim\nolimits_{t \rightarrow \infty}
\dot{H}_k(\dX_t; X_k) - \dot{H}_k(\dX^*; X_k) = 0,
\end{align*}
which means $\bm{0} \in \lim\nolimits_{t \rightarrow \infty} \partial_{2} \dot{H}_k(\dX_t, X_k)$.
\end{proof}

\begin{proof}
(of Theorem~\ref{thm:conv2})
\textbf{Conclusion (i).}
From part 2) in Lemma~\ref{lem:app3}, 
we have
$\frac{\alpha}{2}\| X_{k + 1} - X_k \|^2 \le \dot{R}(X_{k}) - \dot{R}(X_{k + 1}) + \epsilon_k$.
Thus,
\begin{align}
\sum\nolimits_{k = 1}^K \frac{\alpha}{2}\| X_{k + 1} -  X_k \|_F^2
& \le 
\sum\nolimits_{k = 1}^K \dot{R}(X_{k}) - \dot{R}(X_{k + 1}) + \epsilon_k,
\notag
\\
\le & \dot{R}(X_1) - \dot{R}(X_{K + 1}) + \sum\nolimits_{k = 1}^K \epsilon_k,
\notag
\\
\le & \dot{R}(X_1) 
\! - \! \inf \dot{R} 
\! + \! \sum\nolimits_{k = 1}^{\infty} \epsilon_k.
\label{eq:app92}
\end{align}
From Assumption~\ref{ass:err},
we know that the last term in \eqref{eq:app92},
i.e., $\sum\nolimits_{k = 1}^{\infty} \epsilon_{k}$, 
is finite. 
Together with Assumption~\ref{ass:obj},
we have 
\begin{align}
\lim\nolimits_{k\rightarrow \infty} 
& \| X_{k + 1} - X_k \|_F^2  
= 
0,
\label{eq:app52}
\\
\text{\; and \;}
&
0  
\le 
\lim\nolimits_{k\rightarrow \infty} \| X_k \|_F^2 
< 
\infty,
\notag
\end{align}
which means that the sequence $\{ X_k \}$ is bounded
and has at least one limit points.

\vspace{5px}
\noindent
\textbf{Conclusion (ii).}
From Part 1) in Lemma~\ref{lem:app4},
we have
$\partial_{2} \dot{H}_k(\bm{0}, X_k)  = \partial \dot{R}(X_k)$.
{Then, denote the limit point of sequence $\{ X_k \}$ as $X_*$, 
and let $\{ X_{k_j} \}$ be a sub-sequence of $\{ X_k \}$
such that }
\begin{align}
X_* = \lim\nolimits_{k_j \rightarrow \infty} X_{k_j}.
\label{eq:app20}
\end{align}
Thus, to prove a limit point $X^*$ is also a critical point for $\dot{R}(X)$, 
we only need to show 
\begin{align}
\bm{0} \in \lim\nolimits_{k_j \rightarrow \infty} \partial_{2} \dot{H}_{k_j}(\bm{0}, X^*).
\label{eq:app34}
\end{align}
Using Part 2) in Lemma~\ref{lem:app4},
we have 
{\begin{align}
\bm{0} 
& \in \lim\nolimits_{t \rightarrow \infty} \partial_{2} \dot{H}_{k_j}(\dX^{k_j}_{t}, X_{k_j}),
\notag
\\
& = \partial_{2} \dot{H}_{k_j}( \lim\nolimits_{t \rightarrow \infty} \dX^{k_j}_{t}, X_{k_j}).
\label{eq:app21}
\end{align}
Thus, denote $\lim\nolimits_{t \rightarrow \infty} \dX^{k_j}_{t} = \dX^{k_j}_*$, 
we only need to prove 
\begin{align}
\lim\nolimits_{k_j \to \infty} \dX^{k_j}_* = \bm{0}
\label{eq:app23}
\end{align}
Since $\sum\nolimits_{k = 1}^{\infty} \epsilon_{k}$ is finite, we must have $\lim\nolimits_{k_j \to \infty} \epsilon_{k_j} = 0$, 
which implies that 
\begin{align}
\lim\nolimits_{k_j \rightarrow \infty}(X_{k_j + 1} - X_{k_j} - \dX^{k_j}_* )= \mathbf{0}.
\label{eq:app22}
\end{align} 
Then from \eqref{eq:app52}, we have 
\begin{align}
\lim\nolimits_{k_j \rightarrow \infty} X_{k_j + 1} - X_{k_j} = \bm{0}, 
\label{eq:app35}
\end{align}
And \eqref{eq:app23} follows easily from \eqref{eq:app22} and \eqref{eq:app35}. 
Finally,
\eqref{eq:app20} can be obtained
by combining \eqref{eq:app34} and \eqref{eq:app23}.
Thus,
any limit point of $\{ X_k \}$ is a critical point of $\dot{R}$.}
\end{proof}

\subsection{Proposition~\ref{pr:spa}}

We first introduce the following Lemma.

\begin{lemma} \label{lem:matcomp}
	$\tr( \dX \t \bar{Q}_\tau \dX)=\frac{1}{2}(\tilde{x}_i^{\top} \tilde{x}_i+\tilde{x}_j^{\top} \tilde{x}_j)$.
\end{lemma}

\begin{proof}
	We first denote $\tilde{Q}_{\tau} = (Q_\tau+Q_\tau^{\top})/2$, 
	thus $\tilde{Q}_{\tau}$ is a zero matrix with only $(\tilde{Q}_{\tau})_{ij}=(\tilde{Q}_{\tau})_{ji}=1/2$. 
	It can be easily seen that $\tilde{Q}_{\tau}$ has only three different 
	eigenvalues: 0 and $\pm 1/2$. 
	Thus, for $\bar{Q}_\tau = (Q_\tau+Q_\tau^{\top})_+/2 - (Q_\tau+Q_\tau^{\top})_{-}/2$ 
	we can see that it is also a zero matrix 
	with only $(\bar{Q}_{\tau})_{ii}=(\bar{Q}_{\tau})_{jj}=1/2$. 
	{Therefore it follows easily that $\tr( \dX \t \bar{Q}_\tau \dX)= (\tilde{x}_i^{\top} \tilde{x}_i+\tilde{x}_j^{\top} \tilde{x}_j) /2$.}
\end{proof}

\noindent
Next,
we start to prove Proposition~\ref{pr:spa}.

\begin{proof}
Let $\tilde{x}^{\top}_i$ (resp., $(x_k)^{\top}_i$) be the $i$th row of $\tilde{X}$ (resp., $X_k$). 
{Denote $Q^{(i,j)}$ as a zero matrix with only $Q^{(i,j)}_{ij} = 1$. 
Obviously we should have $\tr(\dX \t Q^{(i,j)} X_k)=\tilde{x}_i^{\top} (x_k)_j$ and
$\tr(\dX \t Q^{(i,j)} \dX)=\tilde{x}_i^{\top} \tilde{x}_j$.}
Recall that the objective in \eqref{eq:surradmm} 
for general SDP problem is 
\begin{align*}
\min_\dX \;&\; \tr(\dX\t (B\dX+\g CX_{k})) +
2\sum\nolimits_{\tau=1}^m \abs{e_\tau}
\\
\text{s.t.}\;&\;e_\tau=\tr( \dX \t Q_{\tau} X_{k} ) + (b_k)_\tau
\quad
\tau =1, \dots, m.
\end{align*}
For our matrix completion problem, we have $Q_{\tau} = Q^{(i,j)}$,
$t_{\tau} = O_{ij}$ and $A = 0$.
This gives us $B = Q+\frac{\g}{2} I$, $\g C= \g I$ and $(b_k)_{ij} = \frac{1}{2}((x_k)_i^{\top} (x_k)_j - O_{ij})$.

From Lemma~\ref{lem:matcomp}, 
we need to sum 
the row $\tilde{x}_i^{\top} \tilde{x}_i$ 
and column $\tilde{x}_j^{\top} \tilde{x}_j$ once
when $\Omega_{ij}$ is not zero. 
Thus,
for a specific $\tilde{x}_i^{\top} \tilde{x}_i$, 
we will sum it $\nnz{ \Omega_{(i,:)} } +\nnz{ \Omega_{(:,i)} }$ times,
i.e.,
\begin{align*}
\sum\nolimits_{\tau=1}^m \tr( \dX \t \bar{Q}_\tau \dX)
 =  \sum\nolimits_{i=1}^n \frac{1}{2} (\nnz{ \Omega_{(i,:)} } +\nnz{ \Omega_{(:,i)} } ) \tilde{x}_i^{\top} \tilde{x}_i.
\end{align*}
Let 
$\Lambda^r
= \text{Diag}
(\sqrt{\smash[b]{ \nnz{ \Omega_{(1,:)} } }}, \dots, \sqrt{\smash[b]{ \nnz{\Omega_{(n,:)} }} })$
and $\Lambda^c  = \text{Diag}(  \sqrt{\smash[b]{  \nnz{ \Omega_{(:,1)}} } }, \dots, \sqrt{\smash[b]{
		\nnz{ \Omega_{(:, n)} }}})$, 
we have 
\begin{align*}
\| \Lambda^r  \dX \|_F^2 
& = \sum\nolimits_{i=1}^n \nnz{ \Omega_{(i,:)} } \tilde{x}_i^{\top} \tilde{x}_i,
\\
\| \Lambda^c  \dX \|_F^2 
& = \sum\nolimits_{j=1}^n \nnz{ \Omega_{(:,j)} } \tilde{x}_j^{\top} \tilde{x}_j.
\end{align*}
Combining them together,
we will have
\begin{align*}
\| \Lambda^r  \dX \|_F^2 
 +  \| \Lambda^c  \dX \|_F^2 
 =  \sum\nolimits_{i=1}^n (\nnz{ \Omega_{(i,:)} } 
 +  \nnz{ \Omega_{(:,i)} } ) \tilde{x}_i^{\top} \tilde{x}_i.
\end{align*}
Thus, 
\begin{align*}
\tr( \dX \t Q \dX)
& = \sum\nolimits_{\tau=1}^m \tr( \dX \t \bar{Q}_\tau \dX),
\\
& = \frac{1}{2} \| \Lambda^r  \dX \|_F^2 +  \frac{1}{2} \| \Lambda^c  \dX \|_F^2.
\end{align*}
and
\begin{align*}
\tr( \dX \t B \dX)& = \tr( \dX \t Q \dX) + \frac{\g}{2}\tr(\dX \t \dX), 
\\
& = \frac{1}{2} \| \Lambda^r \dX \|_F^2 + \frac{1}{2} \| \Lambda^c \dX \|_F^2+\frac{\g}{2}\| \dX \|_F^2.
\end{align*}
And it follows easily that 
\begin{align*}
\tr( \dX \t \g C X_k)
= \tr( \dX \lambda I X_k)
= \lambda \tr( \dX \t X_k).
\end{align*}
\noindent
Combining it all together, the objective then becomes
\begin{align*}
 \underset{\dX}{\min}  
& 
\frac{\gamma}{2}
\| \dX \|_F^2  +  \frac{1}{2} \| \Lambda^r  \dX \|_F^2  + \frac{1}{2} \| \Lambda^c  \dX \|_F^2 
\\
&  + \lambda \tr (\dX^{\top}  X_k) +  2 \sum\nolimits_{(i,j)\in \Omega} \abs{e_{ij}},
\\
\text{s.t.\;} & 
e_{ij} = \tilde{x}_i^{\top} (x_k)_j + (b_k)_{ij},
\; \forall (i,j) \in \Omega,
\end{align*}
where $(b_k)_{ij} = \frac{1}{2}((x_k)_i^{\top} (x_k)_j - O_{ij})$.
\end{proof}

\end{document}